\documentclass[11pt]{article}
\usepackage{fix-cm}
\usepackage[letterpaper,margin=1in]{geometry}
\usepackage[T1]{fontenc}
\usepackage{newpxtext}
\usepackage{amsmath,amssymb,amsfonts,amsthm,mathtools,bm}
\usepackage{newpxmath}
\usepackage{aliascnt}
\usepackage{graphicx,booktabs,placeins,array,float}
\usepackage{enumitem}
\usepackage{microtype}
\usepackage[numbers,sort&compress]{natbib}
\usepackage[hyphens]{url}
\usepackage{xcolor}
\usepackage{umoline}
\usepackage{tikz}
\usetikzlibrary{arrows.meta,decorations.pathreplacing,shapes.geometric,positioning}
\definecolor{linkblue}{RGB}{0,92,153}
\usepackage[colorlinks=true,linkcolor=linkblue,citecolor=linkblue,urlcolor=linkblue]{hyperref}
\usepackage[nameinlink,capitalize,noabbrev]{cleveref}
\allowdisplaybreaks
\numberwithin{equation}{section}
\newtheorem{theorem}{Theorem}[section]
\newaliascnt{proposition}{theorem}
\newtheorem{proposition}[proposition]{Proposition}
\aliascntresetthe{proposition}
\newaliascnt{lemma}{theorem}
\newtheorem{lemma}[lemma]{Lemma}
\aliascntresetthe{lemma}
\newaliascnt{corollary}{theorem}

\aliascntresetthe{corollary}
\theoremstyle{definition}
\newaliascnt{definition}{theorem}
\newtheorem{definition}[definition]{Definition}
\aliascntresetthe{definition}
\newaliascnt{assumption}{theorem}
\newtheorem{assumption}[assumption]{Assumption}
\aliascntresetthe{assumption}
\newaliascnt{remark}{theorem}

\aliascntresetthe{remark}
\newaliascnt{example}{theorem}
\newtheorem{example}[example]{Example}
\aliascntresetthe{example}
\crefname{theorem}{Theorem}{Theorems}
\crefname{proposition}{Proposition}{Propositions}
\crefname{lemma}{Lemma}{Lemmas}
\crefname{corollary}{Corollary}{Corollaries}
\crefname{definition}{Definition}{Definitions}
\crefname{assumption}{Assumption}{Assumptions}
\crefname{remark}{Remark}{Remarks}
\crefname{example}{Example}{Examples}

\floatstyle{ruled}
\newfloat{algorithm}{tbp}{loa}
\floatname{algorithm}{Algorithm}
\crefname{algorithm}{Algorithm}{Algorithms}
\graphicspath{{figures/}}

\newcommand{\PP}{\mathbb P}
\newcommand{\EE}{\mathbb E}
\newcommand{\F}{\mathcal F}
\newcommand{\ind}{\mathbf 1}
\DeclareMathOperator{\kl}{kl}
\DeclareMathOperator{\KL}{KL}
\DeclareMathOperator{\Var}{Var}
\renewcommand{\hat}{\widehat}
\renewcommand{\tilde}{\widetilde}

\title{Sequential Selection against an Adaptive Opponent}
\author{Zhongyuan Lyu\\
{\small The University of Sydney Business School}\\
{\small \href{zhongyuan.lyu@sydney.edu.au}{\nolinkurl{zhongyuan.lyu@sydney.edu.au}} }
\and Xiaocong Xu \\
{\small Department of Statistics, University of Pittsburgh}\\
{\small \href{xix169@pitt.edu}{\nolinkurl{xix169@pitt.edu}} }
}
\date{}
\hypersetup{pdftitle={Sequential Selection against an Adaptive Opponent}}

\begin{document}
\maketitle

\begin{abstract}
Selecting a strategy by testing it against an opponent is common in game-playing AI, red-teaming of language models, and safety evaluations of AI agents. In many of these settings, the opponent adapts to what it observes, so a strategy that keeps winning in practice may already have stopped working by the time it is used. Motivated by these settings, we study a player who tries strategies in practice games against an adaptive opponent, decides when to stop, and then either selects one strategy for the next game or abstains. The goal is to select a strategy that meets a target win probability while rarely selecting one that has already fallen below it. For a single strategy, we propose the \emph{Sequential Private Accumulated-excess Rule} (SPAR), which accumulates evidence through a capped likelihood ratio and selects the strategy once this evidence reaches a privately drawn uniform threshold. We show that the stopping time must involve such private randomness. The number of games SPAR needs is inversely proportional to the error tolerance, and this rate is optimal even when the opponent happens not to adapt. To select among multiple strategies, we introduce \emph{Screen-then-SPAR}, which screens candidates in stages before applying SPAR. The procedure adapts to the best strategy's unknown initial margin above the target and achieves the minimax rate up to logarithmic factors. The performance of Screen-then-SPAR is demonstrated through  simulations and two real-data studies, one using recorded coding-agent attacks from MonitoringBench and the other using rock-paper-scissors matches between human players and an adaptive bot.
\end{abstract}

\section{Introduction}\label{sec:intro}

Game-playing programs have essentially solved heads-up limit hold'em \citep{bowling2015} and now beat professional players at no-limit hold'em \citep{moravcik2017,brown2018,brown2019}. Some of these programs keep learning while they play. Libratus, for instance, computed new responses between sessions of its 120{,}000-hand match to the bet sizes that its opponents had used most heavily \citep{brown2018}. A player who wants to exploit such an opponent \citep{ganzfried2015} can find out whether a strategy works only by using it, and each use also shows the opponent what the strategy does. Related concerns arise in red-teaming of language models \citep{perez2022,ganguli2022,wei2023,zou2023} and in adversarial classification and fraud detection \citep{dalvi2004,dalpozzolo2018}, where discovered failures can inform subsequent updates to a model or detector. Similar feedback appears when agents respond to changes in economic policy \citep{lucas1976} or to the criteria used to evaluate them \citep{strathern1997}.

\begin{figure}[!t]
\centering
\begingroup
\newcommand{\inkstone}[2]{%
  \draw[fill=white,line width=.48pt]
    (#1-.14,#2-.045) .. controls (#1-.18,#2+.08) and (#1-.08,#2+.14)
    .. (#1+.04,#2+.12) .. controls (#1+.18,#2+.10) and (#1+.18,#2-.04)
    .. (#1+.10,#2-.09) .. controls (#1+.01,#2-.12) and (#1-.10,#2-.10)
    .. cycle;
  \draw[black!55,line width=.3pt] (#1-.08,#2-.05) -- (#1+.07,#2-.04);}
\newcommand{\inkpile}[3]{%
  \foreach \i in {1,...,#1}{%
    \pgfmathsetmacro{\px}{#2+.34*(mod(\i-1,4)-1.5)}
    \pgfmathsetmacro{\py}{#3-.25*floor((\i-1)/4)}
    \inkstone{\px}{\py}}}
\newcommand{\inkplayer}[3]{%
 \begin{scope}[shift={(#1,#2)},scale=#3,line width=.72pt]
  \draw[fill=white] (0,.90) ellipse[x radius=.28,y radius=.34];
  \draw[line width=1.15pt] (-.27,.98) .. controls (-.36,1.30) and (.15,1.36)
   .. (.29,1.02);
  \draw (-.26,1.12) -- (-.13,1.21) -- (-.03,1.14) -- (.10,1.24);
  \fill (.10,.97) circle[radius=.025];
  \draw (.24,.96) -- (.34,.89) -- (.25,.86);
  \draw (.07,.78) .. controls (.15,.74) and (.21,.76) .. (.24,.80);
  \draw[fill=white] (-.10,.56) -- (-.23,-.12) -- (.24,-.12) -- (.10,.56) -- cycle;
  \draw (-.12,.43) -- (-.45,.10) -- (-.22,-.05);
  \draw (.13,.44) -- (.39,.18) -- (.75,.34);
  \draw (-.11,-.12) -- (-.22,-.65) -- (-.43,-.68);
  \draw (.13,-.12) -- (.30,-.63) -- (.51,-.63);
  \draw[black!45,line width=.3pt] (-.10,.25) -- (.13,.04)
   (-.14,.13) -- (.10,-.07);
 \end{scope}}
\newcommand{\inkopponent}[3]{%
 \begin{scope}[shift={(#1,#2)},scale=#3,line width=.7pt]
  \draw[fill=white] (-.33,.58) -- (-.36,1.16) -- (.31,1.19)
    -- (.35,.59) -- cycle;
  \draw (0,1.18) -- (.03,1.39);
  \draw[fill=white] (.03,1.45) circle[radius=.065];
  \draw[fill=white] (-.16,.96) circle[radius=.07];
  \draw[fill=white] (.16,.96) circle[radius=.07];
  \draw (-.18,.74) -- (.18,.74);
  \draw (-.08,.70) -- (-.08,.78) (.06,.70) -- (.06,.78);
  \draw (-.35,.94) -- (-.46,.94) -- (-.46,.79) -- (-.35,.79);
  \draw (.35,.94) -- (.45,.94) -- (.45,.79) -- (.35,.79);
  \draw[fill=white] (-.20,.58) -- (-.26,-.11) -- (.27,-.11) -- (.20,.58) -- cycle;
  \draw[fill=white] (-.13,.34) rectangle (.15,.07);
  \draw (-.23,.43) -- (-.48,.14) -- (-.79,.35);
  \draw (.23,.43) -- (.47,.20) -- (.62,.37);
  \draw (-.14,-.11) -- (-.19,-.63) -- (-.42,-.63);
  \draw (.14,-.11) -- (.23,-.63) -- (.45,-.63);
  \foreach \yy in {.46,.51,.56}{\draw[black!55,line width=.28pt]
    (.19,\yy) -- (.26,\yy-.05);}
 \end{scope}}
\resizebox{\linewidth}{!}{%
\begin{tikzpicture}[x=1cm,y=1cm,line cap=round,line join=round,
 every node/.style={font=\fontsize{8.6}{10.2}\selectfont,inner sep=1.5pt},
 flow/.style={-{Stealth[length=1.55mm,width=1.05mm]},line width=.65pt},
 smallnote/.style={font=\fontsize{8.3}{10.0}\selectfont},
 headline/.style={font=\bfseries\fontsize{10.2}{12}\selectfont,anchor=west}]
\path[use as bounding box] (-.05,5.05) rectangle (16.55,10.5);
\draw[line width=.95pt] (.02,5.14) -- (0,10.43) -- (16.48,10.47)
 -- (16.50,5.12) -- cycle;

\node[headline] at (.25,10.10) {Find a weakness before it is repaired};
\node[smallnote,anchor=west] at (.28,9.67)
 {Take 1 or 2 stones. Taking the last stone wins.};
\draw[line width=.6pt] (1.50,7.38) -- (3.20,7.38);
\draw[line width=.4pt] (1.70,7.38) -- (1.70,6.55)
 (3.00,7.38) -- (3.00,6.55);
\inkplayer{.70}{7.05}{.92}
\inkopponent{4.05}{7.05}{.92}
\inkpile{8}{2.30}{7.75}
\node[smallnote] at (2.30,6.18) {8 stones};
\node[smallnote] at (.70,6.18) {Player};
\node[smallnote] at (4.05,6.18) {Opponent};
\draw[line width=.6pt,fill=white]
 (.30,8.95) .. controls (.25,9.59) and (2.84,9.59) .. (2.88,8.95)
 .. controls (2.88,8.53) and (1.64,8.46) .. (1.35,8.55)
 -- (1.03,8.50) -- (1.10,8.60)
 .. controls (.68,8.60) and (.48,8.65) .. cycle;
\node[align=center,font=\itshape\fontsize{8.8}{10.4}\selectfont]
 at (1.54,8.98) {Which move\\works?};
\draw[flow] (4.70,7.81) -- (5.10,7.81);
\node[smallnote,align=center] at (4.95,8.43) {Opponent\\takes 2};
\inkpile{6}{6.00}{7.92}
\node[smallnote] at (6.00,7.26) {6 stones};
\node[smallnote] at (6.00,6.92) {Player's turn};
\draw[flow] (6.99,7.78) -- (7.25,7.78) -- (7.25,8.79) -- (7.55,8.79);
\draw[flow] (7.25,7.78) -- (7.25,6.50) -- (7.55,6.50);
\node[smallnote,align=center,anchor=east] at (7.04,8.97) {Strategy 1:\\take 2};
\node[smallnote,align=center,anchor=east] at (7.04,6.15) {Strategy 2:\\take 1};
\inkpile{4}{8.50}{8.83}
\inkpile{5}{8.50}{6.62}
\node[smallnote] at (8.50,8.35) {4 stones};
\node[smallnote] at (8.50,8.01) {Opponent's turn};
\node[smallnote] at (8.50,6.01) {5 stones};
\node[smallnote] at (8.50,5.67) {Opponent's turn};
\draw[flow,line width=.9pt] (9.50,8.79) -- (9.80,8.79) -- (9.80,9.28) -- (12.70,9.28);
\draw[flow,densely dashed] (9.80,8.79) -- (9.80,8.13) -- (12.70,8.13);
\node[smallnote] at (11.25,9.53) {Opponent takes 2 (error)};
\node[smallnote] at (11.25,8.39) {Opponent takes 1};
\node[font=\bfseries\fontsize{9.1}{10.7}\selectfont] at (14.38,9.28) {Player wins};
\node[smallnote] at (14.38,8.88) {takes the last 2};
\node at (14.38,8.13) {Opponent wins};
\node[smallnote] at (14.38,7.77) {leaves 3 for the player};
\draw[flow,line width=.9pt] (9.50,6.49) -- (9.80,6.49) -- (9.80,6.99) -- (12.70,6.99);
\draw[flow,densely dashed] (9.80,6.49) -- (9.80,5.85) -- (12.70,5.85);
\node[smallnote] at (11.25,7.25) {Opponent takes 1 (error)};
\node[smallnote] at (11.25,6.10) {Opponent takes 2};
\node[font=\bfseries\fontsize{9.1}{10.7}\selectfont] at (14.38,6.99) {Player wins};
\node[smallnote] at (14.38,6.60) {leaves 3; wins next turn};
\node at (14.38,5.85) {Opponent wins};
\node[smallnote] at (14.38,5.49) {leaves 3 for the player};

\end{tikzpicture}%
}
\endgroup
\caption{The stone game of the example. After the opponent's opening move, the player chooses between two strategies. The labels on the right give the winner when both sides play perfectly after the opponent's reply. Solid branches are the opponent's errors, which the player can exploit, and dashed branches are its winning replies.}
\label{fig:stones}
\end{figure}
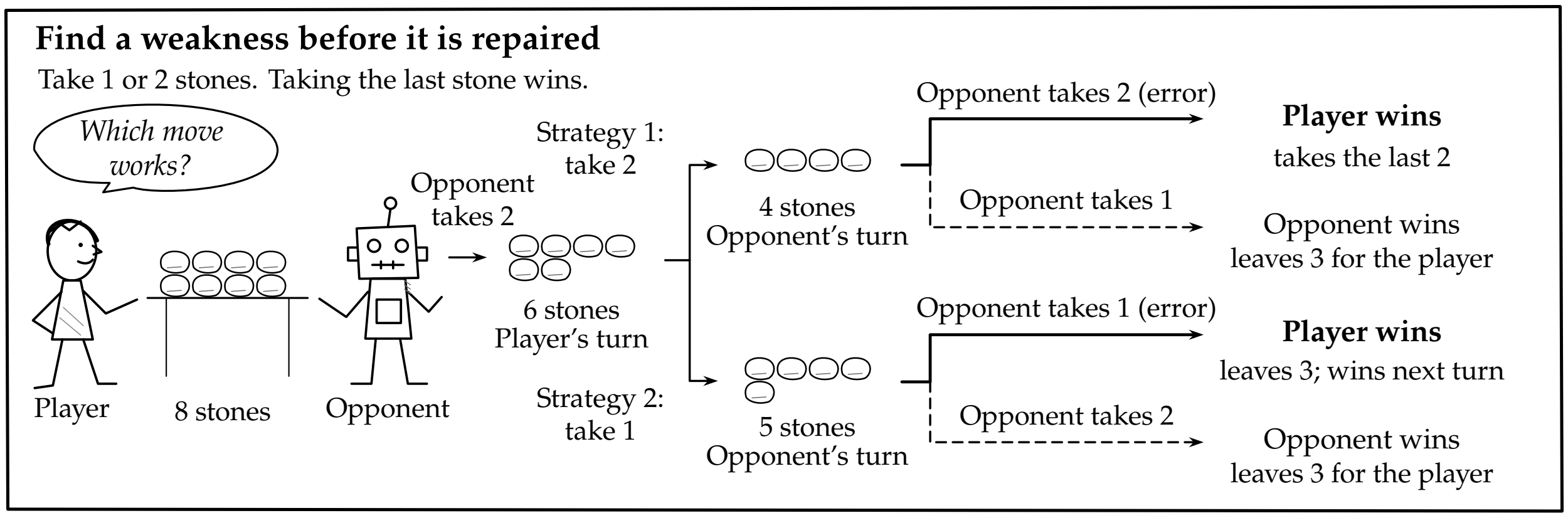

A much simpler game, shown in \cref{fig:stones}, has the same feature. The player and the opponent take turns removing one or two stones from a pile, and whoever takes the last stone wins. The opponent moves first and takes two of eight stones, after which the player can take two stones (strategy~1) or one stone (strategy~2). Both sides play perfectly after the opponent's reply, so the player wins exactly when that reply is an error. The stones are reset after each game, but the opponent keeps what it has learned and becomes less likely to err at a position each time it reaches it. As the opponent learns, past wins can overstate strategy~1's chance of winning the next game.

\begin{figure}[!htb]
\centering
\tikzset{
 fig2act/.style={draw,rounded corners=3pt,align=center,font=\small,inner sep=4pt,minimum height=1.05cm},
 fig2eval/.style={draw,line width=1.1pt,align=center,font=\small,inner sep=4pt,minimum height=.8cm},
 fig2val/.style={draw,rounded corners=3pt,fill=black!6,font=\small,inner sep=2pt,minimum width=1.3cm,minimum height=.75cm},
 fig2dec/.style={draw,diamond,aspect=1.6,inner sep=1.5pt,font=\small},
 fig2arr/.style={-{Stealth[length=2.4mm,width=1.7mm]},line width=.55pt},
 fig2note/.style={font=\footnotesize\itshape,text=black!70,align=center},
 fig2panel/.style={font=\small\bfseries,anchor=west}}
\resizebox{\linewidth}{!}{%
\begin{tikzpicture}[x=1cm,y=1cm]
\path[use as bounding box] (-1.72,-2.15) rectangle (15.74,2.6);
\node[fig2panel] at (-1.68,2.35) {(a) The player's choices};
\node[fig2act,minimum width=3.3cm] (try) at (0,0) {practice game:\\try one of the $K$ strategies};
\node[fig2act,minimum width=2.3cm] (see) at (3.9,0) {see whether\\it wins};
\node[fig2dec] (stop) at (6.75,0) {stop?};
\node[fig2act,minimum width=3.0cm] (sel) at (10.2,.95) {select one strategy\\for the next game};
\node[fig2act,minimum width=3.0cm] (abs) at (10.2,-.95) {abstain:\\select nothing};
\node[fig2eval,minimum width=3.2cm] (good) at (14.1,1.5) {value $\geq q$: correct};
\node[fig2eval,minimum width=3.2cm] (bad) at (14.1,.35) {value $<q$: incorrect};
\draw[fig2arr] (try) -- (see);
\draw[fig2arr] (see) -- (stop);
\draw[fig2arr] (stop.north) -- ++(0,.75) -| node[fig2note,above,pos=.25] {no: play another game} (try.north);
\draw[fig2arr] (stop.east) -- node[fig2note,above,pos=.45] {yes} ++(.55,0) |- (sel.west);
\draw[fig2arr] (stop.east) ++(.55,0) |- (abs.west);
\draw[fig2arr] (sel.east) -- ++(.35,0) |- (good.west);
\draw[fig2arr] (sel.east) ++(.35,0) |- (bad.west);
\node[fig2note,below=3pt of try] {(1) which strategy to try};
\node[fig2note,below=4pt of stop] {(2) when to stop};
\node[fig2note] at (10.2,-1.85) {(3) what to select};
\end{tikzpicture}}

\bigskip
\resizebox{\linewidth}{!}{%
\begin{tikzpicture}[x=1cm,y=1cm]
\path[use as bounding box] (-1.72,-.5) rectangle (15.74,6.0);
\node[fig2panel] at (-1.68,5.75) {(b) What the player observes and what counts};
\node[fig2val] (v0) at (-1.0,4) {$V_0$};
\node[fig2val] (v1) at (1.7,4) {$V_1$};
\node at (3.1,4) {$\cdots$};
\node[fig2val] (vs) at (4.5,4) {$V_{\tau-1}$};
\node[fig2val,line width=1.1pt,fill=black!16] (vt) at (8.2,4) {$V_\tau$};
\node[fig2act,minimum width=3.2cm] (g1) at (0,1) {game $1$, strategy $A_1$\\win prob.\ $V_0(A_1)$};
\node at (3.1,1) {$\cdots$};
\node[fig2act,minimum width=3.3cm] (gt) at (6.3,1) {game $\tau$, strategy $A_\tau$\\win prob.\ $V_{\tau-1}(A_\tau)$};
\node[fig2act,minimum width=3.0cm] (sel) at (10.6,1) {stop after game $\tau$:\\select $\hat J$ or abstain};
\node[fig2eval,minimum width=2.7cm] (ev) at (14.3,2.75) {next game:\\is $V_\tau(\hat J)\geq q$?};
\draw[fig2arr] (v0) -- (g1);
\draw[fig2arr] (g1) -- node[fig2note,sloped,above] {adapts} (v1);
\draw[fig2arr] (vs) -- (gt);
\draw[fig2arr,line width=1.1pt] (gt) -- node[fig2note,sloped,above] {adapts} (vt);
\draw[fig2arr] (gt) -- (sel);
\draw[fig2arr] (vt) -- (ev);
\draw[fig2arr] (sel) -- (ev);
\draw[decorate,decoration={brace,amplitude=5pt}] (-1.65,4.55) -- (5.15,4.55) node[midway,above=6pt,fig2note] {values that generate the practice outcomes};
\node[fig2note,above=3pt] at (vt.north) {value used to judge\\the selection};
\draw[decorate,decoration={brace,amplitude=5pt,mirror}] (-1.6,.38) -- (7.95,.38) node[midway,below=6pt,fig2note] {practice observations, seen by the player; success requires $\tau\leq N$};
\end{tikzpicture}}
\caption{The problem. (a) Before each practice game, the player chooses which strategy to try, and after each game it decides whether to stop. When it stops, it either selects one strategy, which is then used in the next game, or abstains. A selection is correct if the selected strategy meets the target, that is, if its value is at least $q$. Abstaining is neither correct nor incorrect. (b) The player does not observe the values $V_n=(V_n(1),\ldots,V_n(K))$. In practice game $n$, strategy $A_n$ wins with probability $V_{n-1}(A_n)$, and the values then change as the opponent responds to the observation. The selection is judged by $V_\tau(\hat J)$, a value that has generated no practice outcome. In both panels, rounded boxes are the player's actions and heavy boxes evaluate a selection. The shaded boxes in (b) are the unobserved values.}
\label{fig:problem}
\end{figure}
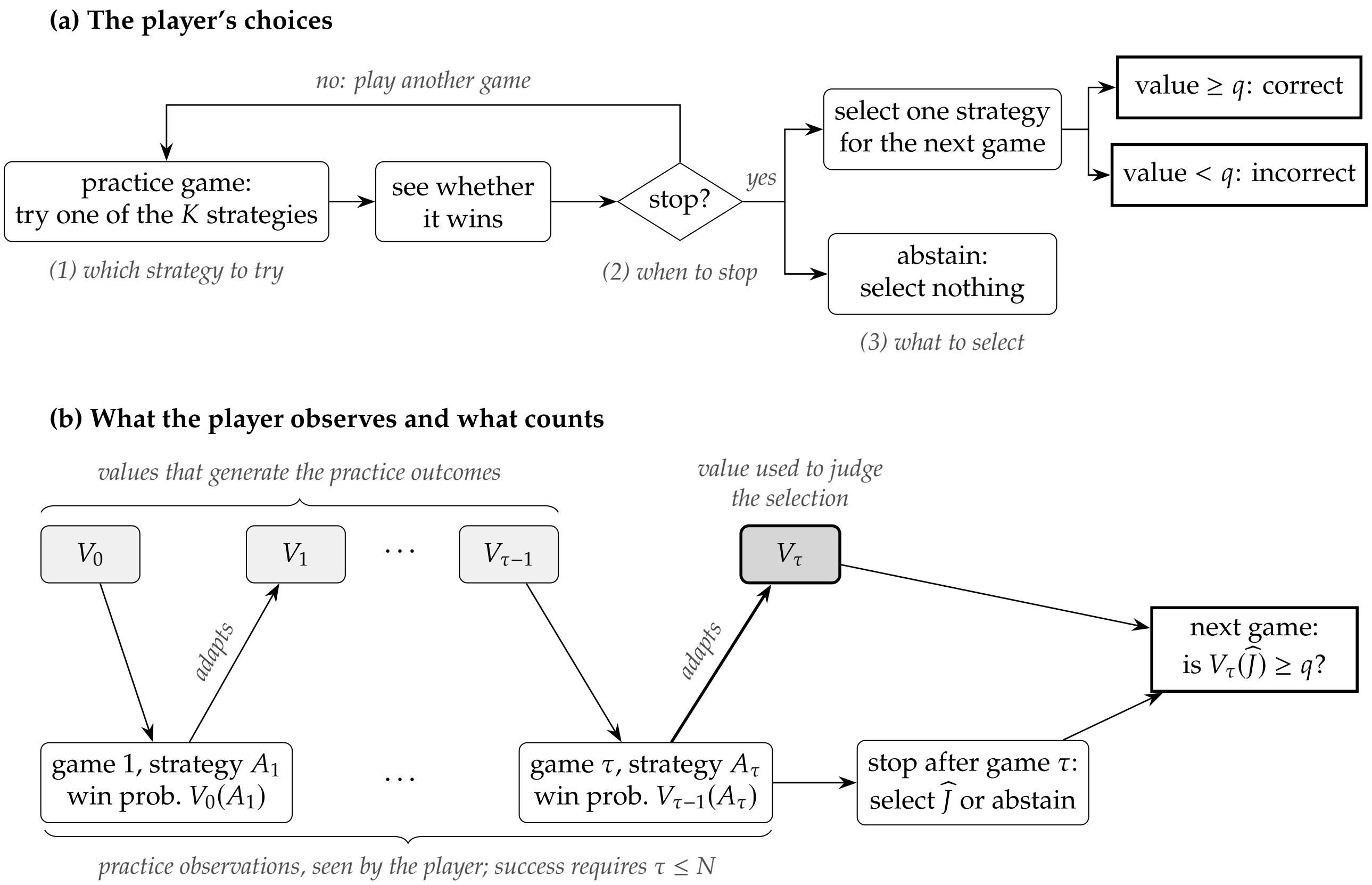

Motivated by such interactions, we study a sequential selection problem in which the player tries strategies in practice games, decides when to stop, and then either selects one strategy for the next game or abstains. This setup is summarized in \cref{fig:problem}(a). We call a strategy's probability of winning the next game its \emph{value} (defined formally in \eqref{eq:value}), and we say that the strategy \emph{meets the target} if its value is at least a given threshold $q$. Practice involves a tradeoff:
\begin{align}\label{tradeoff}\tag{T}
\text{more practice games}\quad\Longrightarrow\quad \begin{cases} \text{the player learns more about the strategies it tries},\\[2pt]
\text{the opponent learns more and may repair its weaknesses}. \end{cases}
\end{align}
This tradeoff raises the following question:
\begin{equation}\label{eq:question}\tag{Q}
\parbox{0.8\linewidth}{\centering\itshape When can the player reliably select a strategy whose chance of winning the next game meets a prescribed target?}
\end{equation}

To make this question precise, consider $K$ strategies, and let $V_n(j)$ be the value of strategy $j$ after $n$ practice games, given the observed history. A procedure chooses which strategy to use before each practice game, stops after $\tau$ games, and either selects a strategy $\hat J$ or abstains, which we write as $\hat J=\varnothing$. The selected strategy is then used in game $\tau+1$ and evaluated by its post-update value $V_\tau(\hat J)$, as shown in \cref{fig:problem}(b).

We call the mechanism by which the opponent reacts to the observed history an \emph{environment} $\nu$. It determines all the values $V_n(j)$, and we write $\PP_\nu$ (with expectation $\EE_\nu$) for the law of the experiment under $\nu$. A desirable procedure should rarely select a strategy whose value has fallen below the target, and should select one that meets the target within a budget of $N$ practice games whenever a viable strategy exists. We formalize these two criteria as follows.

\begin{definition}[Validity and success]\label{def:requirements}
Let $q\in(1/2,1)$ be a target,\footnote{The restriction $q>1/2$ is for simplicity. Analogous results hold for $q\leq1/2$, with different constants in the bounds.} $\varepsilon\in(0,1/2)$ a tolerance, and $\mathcal C$ a class of environments. A procedure is \emph{$\varepsilon$-valid} on $\mathcal C$ if
\begin{align}\label{eq:validity}
\sup_{\nu\in\mathcal C} \PP_\nu\{\tau<\infty,\ \hat J\neq\varnothing, \ V_\tau(\hat J)<q\}\leq\varepsilon.
\end{align}
For an integer budget $N\geq1$, it is \emph{$(N,\varepsilon)$-successful} on $\mathcal C$ if
\begin{align}\label{eq:success}
\inf_{\nu\in\mathcal C} \PP_\nu\{\tau\leq N,\ \hat J\neq\varnothing, \ V_\tau(\hat J)\geq q\}\geq1-\varepsilon.
\end{align}
\end{definition}

As the underlying environment $\nu$ is unknown to the player, \cref{def:requirements} requires both guarantees to hold uniformly over a specified class of environments. It worth noting that we will impose validity and success on different classes. In particular, let $\mathcal E_q$ be the class of environments in which repairs last, in the sense that once the value of a strategy falls below $q$, it stays at or below $q$ for as long as the player keeps using that strategy. For $\Delta>0$ and $\lambda\geq0$, let $\mathcal E_q(\Delta,\lambda)\subseteq\mathcal E_q$ be the subclass in which, in addition, some strategy starts at least $\Delta$ above the target and no value falls by more than $\lambda$ in one game:
\begin{align}\label{eq:intro-class}
\max_{j}V_0(j)\geq q+\Delta \qquad\text{and}\qquad V_n(j)\geq V_{n-1}(j)-\lambda\ \ \text{for all $n\geq1$ and $j$}.
\end{align}
\Cref{sec:model} defines these classes precisely. We call a procedure \emph{valid} if it is $\varepsilon$-valid on $\mathcal E_q$ and \emph{successful} if it is $(N,\varepsilon)$-successful on $\mathcal E_q(\Delta,\lambda)$. The two requirements play the roles of the level and the power of a hypothesis test. Validity alone is easy to achieve, since a procedure that always abstains is valid, just as a test that never rejects has level zero. The success requirement rules out this trivial solution by requiring a correct selection with high probability within the budget. Obviously, such a guarantee is impossible over all of $\mathcal E_q$, which includes environments in which no strategy ever meets the target or the opponent can eliminate all initial advantages arbitrarily quickly. We therefore require success only on the smaller class $\mathcal E_q(\Delta,\lambda)$. The challenge is to remain valid on $\mathcal E_q$ while being successful on $\mathcal E_q(\Delta,\lambda)$, without knowing whether the environment belongs to the smaller class.

Our formulation is related to adversarial online learning and to classical sequential pure exploration, but it differs from each in a basic way.

\paragraph{Comparison with adversarial online learning.}
Adversarial online learning also allows the opponent to respond to earlier play, but it judges the player by the total reward collected over the games played. Policy regret \citep{arora2012}, for example, compares this total with what the player would have collected by playing one fixed strategy throughout, against an opponent that responds to that strategy. Adversarial best-arm identification with oblivious reward sequences \citep{abbasiyadkori2018} does end with a recommendation, but it still judges the recommended arm by the cumulative reward assigned to it over the sampling horizon. In our problem, by contrast, the practice games carry no reward. The player chooses when to stop and whether to select anything at all, and the requirements of \cref{def:requirements} involve only the value $V_\tau(\hat J)$ of the selected strategy. With a single strategy, regret is zero over any horizon and every recommendation is trivially correct, yet the player still has to decide whether the strategy meets the target for one more use. The essential difference is that these formulations judge the player by the games it has already played, whereas we judge the selected strategy by its value in the next game.

\paragraph{Comparison with sequential testing and best-arm identification.}
If the value of a single strategy never changes, validity and success reduce to the level and power of a one-sided sequential test \citep{wald1945,chernoff1959}. With several strategies, the problem resembles ranking and selection \citep{bechhofer1954,kim2001,hong2021} and best-arm and good-arm identification \citep{evendar2006,audibert2010,kano2019}. All of these formulations assume that each strategy has a fixed value, so every observation is evidence about the same quantity. In our problem, however, the value changes as the opponent learns, and the observations lag behind it. For a single strategy with value $p_t$ after $t$ games, the outcome of game $t$ is a win with probability $p_{t-1}$, whereas a selection made after game $t$ is judged by $p_t$.

To see the consequence, let $n\geq1$ and $0<r<q<p<1$, and compare an opponent that keeps the value at $p$ with one that keeps it at $p$ for $n$ games and then lowers it to $r$. The first $n$ outcomes have the same distribution under both opponents, yet after game $n$ the value is above the target under the first and below it under the second. A valid procedure selects right after game $n$ with probability at most $\varepsilon$ against the second opponent, and hence also against the first, since nothing in the first $n$ outcomes tells them apart. Playing one more game would provide evidence of the drop, but it would also give the opponent another chance to change the value. The essential difference, then, is that in these formulations enough evidence justifies selecting at once, but in ours no amount of evidence does, because the selection is judged by a value set after the last observation.

\subsection{Main results and ideas}\label{sec:overview}

We begin with the case of a single strategy, where the player only decides when to stop and whether to select the strategy. Suppose that the player's stopping time depends only on the observed history. Since the opponent observes the same history, it knows when the player stops and can lower the strategy's value below $q$ right before the next game.  As we show in \cref{prop:history}, no such rule can be both valid and successful.

In light of the above result, the stopping time must therefore depend on private randomization that the opponent does not observe, as in selective prediction \citep{drucker2013,qiao2019}. Since the opponent can still lower the value at a time of its choosing, the procedure must also select at any given time with probability at most $\varepsilon$. Our verification rule, the \emph{Sequential Private Accumulated-excess Rule} (SPAR, \cref{alg:verify}), meets both requirements. It accumulates a statistic $S_t$ that increases with the evidence that the value exceeds $q$, and it selects the strategy once $S_t$ reaches a private threshold drawn uniformly from $[0,1]$. Conditionally on the outcomes, the probability that SPAR has selected by game $t$ equals $\min\{S_t,1\}$. The statistic is constructed from a likelihood ratio that is capped after every game, so $S_t$ increases by at most $\varepsilon$ in each game, and only when the likelihood ratio reaches a new maximum. Once the value falls below $q$, the sum of $S_t$ and the capped likelihood ratio becomes a supermartingale, so the expected remaining increase of $S_t$ is at most $\varepsilon$. Hence the probability of an incorrect selection is at most $\varepsilon$.

When $\Delta$ is known, we show in \cref{thm:single} that SPAR is valid and that under \eqref{eq:intro-class} it is successful whenever
\begin{align}\label{eq:intro-single}
N\gtrsim\frac{q(1-q)}{\varepsilon\Delta^2}\qquad\text{and}\qquad\frac{\Delta}{\lambda}\gtrsim\frac{q(1-q)}{\varepsilon\Delta^2}.
\end{align}
Both conditions are also necessary, and the first remains necessary even against an opponent that never changes the value. If validity were required only against such opponents, a budget of order $q(1-q)\log(1/\varepsilon)/\Delta^2$ would suffice.

The second condition in \eqref{eq:intro-single} can be written as
\begin{align}\label{eq:tradeoff}
\underbrace{\frac{\Delta^2}{q(1-q)}}_{\mathclap{\substack{\text{information}\\
\text{per game}}}}\qquad\times\qquad\underbrace{\frac{\Delta\vphantom{\Delta^2}}{\lambda\vphantom{q(1-q)}}}_{\mathclap{\substack{\text{games for which}\\
\text{the margin lasts}}}}\qquad\quad\gtrsim\qquad\quad\underbrace{\frac{1\vphantom{\Delta^2}}{\varepsilon\vphantom{q(1-q)}}}_{\mathclap{\substack{\text{information needed}\\
\text{for a valid selection}}}}.
\end{align}
This makes precise the tradeoff stated in \eqref{tradeoff}. A margin $\Delta$ that falls by at most $\lambda$ per game lasts for at least $\Delta/\lambda$ games, each of which provides Kullback--Leibler information of order $\Delta^2/[q(1-q)]$ for distinguishing $q+\Delta$ from $q$. A procedure that is both valid and successful must collect information of order $1/\varepsilon$ before the margin disappears.

When $\Delta$ is unknown or there are multiple strategies, we propose \emph{Screen-then-SPAR} (\cref{alg:search}). Because verifying every strategy would be costly, it screens the strategies in blocks of doubling length and runs SPAR only on the first strategy that passes its screen. We show in \cref{thm:search} that Screen-then-SPAR is valid and that it is successful without requiring knowledge of $\Delta$, $\lambda$, or which strategies weaken one another, provided that
\begin{align}\label{eq:intro-search}
N\gtrsim\frac{q(1-q)}{\Delta^2}\Bigl(K\Lambda+\frac1\varepsilon\Bigr) \qquad\text{and}\qquad \frac{\Delta}{\lambda}\gtrsim\frac{q(1-q)}{\Delta^2}\Bigl(K_{\max}\Lambda+\frac1\varepsilon\Bigr),
\end{align}
where $\Lambda:=\log(K/\varepsilon)+\log\log[q(1-q)/\Delta^2]$ and $K_{\max}$ is the size of the largest group of strategies that can weaken one another. In both conditions the term with $\Lambda$ is the cost of the screens and the term $1/\varepsilon$ is the cost of a single verification as in \eqref{eq:intro-single}. The budget must pay for the screens of all $K$ strategies. The margin of a strategy that starts above the target only has to last through its own verification and the screens of the strategies in its group, since only these strategies can lower its value. This is why $K$ appears in the first condition and $K_{\max}$ in the second. Conversely, \cref{thm:lower} constructs a game in which every valid and successful procedure must satisfy \eqref{eq:intro-search} with $\Lambda$ replaced by $\log(1/\varepsilon)$. Screen-then-SPAR is therefore minimax optimal up to logarithmic factors.

\subsection{Related work}\label{sec:related}

Our work relates to pure exploration and to sequential decision making against adaptive opponents.

\begin{itemize}
\item \textbf{Best-arm identification and ranking and selection.} Pure exploration in multi-armed bandits \citep{evendar2006,audibert2010,garivier2016,degenne2019} and ranking and selection in simulation optimization \citep{bechhofer1954,kim2001,nelson2001,batur2010} study related selection problems; see \citep{hong2021} for a review. The closest problems are good-arm identification \citep{kano2019}, thresholding bandits \citep{locatelli2016}, and sequential tests for the lowest mean \citep{kaufmann2018}, all of which assume that the values are fixed. Our search adapts to an unknown margin through blocks of doubling length, a multiscale approach related to staged pure-exploration methods \citep{karnin2013}, and the split into search and verification follows the explore--verify approach \citep{karnin2016}. In our setting, verification requires of order $1/\varepsilon$ observations for a fixed margin. Standard change-of-measure comparisons with stationary alternatives yield only a logarithmic dependence on $1/\varepsilon$ \citep{kaufmann2016,garivier2016}; see also \citet[Section~2]{garivier2019} for the underlying KL information inequality. To establish the $1/\varepsilon$ lower bound, we treat the opponent's choice of when to lower the value as an optimal stopping problem \citep{chow1971} (\cref{lower:certification}).

\item \textbf{Bandits with changing rewards.} In several bandit models the rewards change as the arms are used, including rested and rising bandits \citep{cella2021,mussi2024,takemori2024}, rotting bandits \citep{levine2017,seznec2019}, and graph-triggered bandits, in which using one arm also changes others \citep{genalti2026}. In these models, the change is driven mainly by how often the arms are used, and the goal is low regret or identification of the best arm. In our problem, the opponent can decide from the observed history both how and when the values change, and the selection is judged after the last change.

\item \textbf{Adaptive opponents and performative prediction.} Safe opponent exploitation \citep{ganzfried2015} exploits an opponent's mistakes while guaranteeing at least the equilibrium payoff in expectation, and performative prediction \citep{perdomo2020} seeks predictions that are stable or optimal under the distribution they induce. Safe policy improvement for nonstationary decision processes \citep{chandak2020} bounds the performance of a policy in a future episode when the environment changes smoothly over time. We instead evaluate a single selection by its value after the opponent's last update.

\item \textbf{Selective prediction.} Selective prediction \citep{drucker2013,qiao2019,qiao2021} uses randomized timing to predict a statistic, or select a model, for a future window, with guarantees for every sequence fixed independently of the learner's random choices. Its errors concern realized outcomes rather than the value of a selected strategy. For comparison, we adapt the random-window idea to construct a valid block verifier (\cref{app:block-validity}). At a fixed target and margin, SPAR saves a factor of order \(\log(1/\varepsilon)\) in the verification budget.

\item \textbf{Sequential testing by betting.} SPAR is related to testing by betting and nonnegative supermartingales \citep{shafer2021,ramdas2023game,howard2020,waudby2024}. It sets aside the excess above the cap, as in the classical construction that withdraws gains into savings \citep{teutsch2014}, and compares the accumulated excess with an independent uniform threshold, which relates to the uniformly randomized Markov inequality \citep{ramdas2023}. Sequential change detection \citep{lorden1971} addresses a related concern, but it can detect only changes that affect later observations, and the drop that matters here comes after the last observation.
\end{itemize}

\paragraph{Notation.} For a sequence $o_1,o_2,\ldots$, we write $o_{1:n}=(o_1,\ldots,o_n)$. Throughout, $\log$ is the natural logarithm, $\ind\{\cdot\}$ denotes an indicator, and $[x]_+=\max\{x,0\}$. We write $\KL(P,P')$ for the Kullback--Leibler divergence between distributions $P$ and $P'$, and $\kl(p,p')$ for its Bernoulli form $p\log(p/p')+(1-p)\log[(1-p)/(1-p')]$, with the usual conventions at zero and one. For positive $a$ and $b$, we write $a\lesssim b$, or $b\gtrsim a$, if $a\leq Cb$ for a universal constant $C$.

\paragraph{Organization.} The paper is organized as follows. \Cref{sec:model} formulates the problem, \cref{sec:single} treats a single strategy with a known margin, and \cref{sec:search} develops the fully adaptive procedure for any $K\geq 1$. \Cref{sec:numerics,sec:realdata} report simulations and replays of recorded attacks, and \cref{sec:discussion} discusses extensions. The appendices contain the proofs and the details of the experiments.

\section{Problem formulation}\label{sec:model}

\subsection{The experiment}\label{sec:experiment}

Fix integers $K,N\geq1$, and let $q\in(1/2,1)$ and $\varepsilon\in(0,1/2)$ be the target and the tolerance of \cref{def:requirements}. Each strategy $j\in[K]=\{1,\ldots,K\}$ specifies a fixed rule for playing a single game. Before game $n$, the player chooses a strategy $A_n\in[K]$, and after the game it receives an observation $O_n$ taking values in a measurable space $\mathcal O$.  We write $z(o)\in\{0,1\}$ for the outcome of an observation $o\in\mathcal O$, with $z(o)=1$ for a win, and $Z_n=z(O_n)$ for the outcome of game $n$. We use the following example to illustrate the observation space \(\mathcal O\) and the outcome map \(z\).

\begin{example}[Observations in the stone game]
In the stone game of \cref{fig:stones}, let $H_n$ list how many stones
are taken on each turn of game $n$, starting with the opponent's turn.
Then
\begin{align*}
O_n=(A_n,H_n), \qquad \mathcal O=\{(j,h): j\in\{1,2\},\ h\text{ is a possible list under strategy }j\}.
\end{align*}
For example, $O_n=(1,(2,2,2,2))$ means that the player used strategy~1
and two stones were taken on each turn. The player took the last two
stones, so $Z_n=z(O_n)=1$.
\end{example}

Each game starts from the same initial position. When game $n+1$
begins, however, the opponent may use the observations
$O_{1:n}=(O_1,\ldots,O_n)$ from the preceding games. An \emph{environment} $\nu$ describes how the opponent plays and adapts, together with the randomness in the moves of both sides. Formally, $\nu=(\nu_n)_{n\geq0}$ is a sequence of probability kernels, where $\nu_n(\cdot\mid o_{1:n},j)$ is the distribution on $\mathcal O$ of the next observation when strategy $j$ is used after the history $o_{1:n}\in\mathcal O^n$.  

After $n$ games, the player's \emph{procedure} uses the public history $O_{1:n}$ and a private random seed $U$ to decide whether to continue or stop. If it continues, it chooses the next strategy $A_{n+1}$. If it stops at time $\tau$, it selects a strategy $\hat J\in[K]$ or abstains ($\hat J=\varnothing$). Write $\mathcal B_n=\sigma(O_{1:n})$ and $\F_n=\mathcal B_n\vee\sigma(U)$ for the time-$n$ $\sigma$-fields of the public and player filtrations, respectively. The opponent reacts only to the public history, which includes the strategies used. In particular, it is never told when practice ends, and while it may learn about $U$ from the observed strategy choices, it never observes $U$ directly. Appendix~\ref{app:stopping} gives the formal conditions.

We define the \emph{value} of strategy $j$ after $n$ games as
\begin{align}\label{eq:value}
V_n(j)=\nu_n\bigl(\{o:z(o)=1\}\mid O_{1:n},j\bigr), \qquad n\geq0,\ j\in[K].
\end{align}
Thus $V_n(j)$ is the conditional win probability of strategy $j$ in game $n+1$ given the observed history $O_{1:n}$, with deterministic initial values $V_0(j)$. By construction,
\begin{align}\label{eq:mean}
\PP_\nu(Z_{n+1}=1\mid\F_n)=V_n(A_{n+1}),
\end{align}
so game $n+1$ is won with probability $V_n(A_{n+1})$, and after this game the values may change in response to the new observation. If the procedure selects $\hat J$, then $V_\tau(\hat J)$ is the probability that the selected strategy wins the next game.

\subsection{Classes of environments}\label{sec:requirements}

For validity, we consider the environments that satisfy the following assumption. Games $t+1,\ldots,t+k$ form a \emph{run} of strategy $j$ if $A_{t+1}=\cdots=A_{t+k}=j$.

\begin{assumption}[Persistence below the target]\label{ass:shape}
For every procedure, almost surely, for all integers $t\geq0$ and $k\geq1$ and all $j\in[K]$ such that games $t+1,\ldots,t+k$ form a run of $j$,
\begin{align}\label{eq:persistence}
V_{t+r}(j)<q\quad\Longrightarrow\quad V_{t+s}(j)\leq q, \qquad 0\leq r<s\leq k.
\end{align}
\end{assumption}

\begin{figure}[!tb]
\centering
\begin{tikzpicture}[x=.615cm,y=1.02cm,line cap=round,line join=round,
  panel title/.style={font=\small\bfseries,anchor=west,inner sep=0pt},
  axis text/.style={font=\footnotesize,text=black!78,inner sep=2pt},
  note/.style={font=\footnotesize,inner sep=1pt},
  axis/.style={draw=black!46,line width=.42pt},
  target/.style={draw=black!62,line width=.6pt,dash pattern=on 3pt off 2pt},
  trajectory/.style={color=linkblue,line width=1.2pt},
  leader/.style={draw=black!58,line width=.45pt}]
  \begin{scope}[shift={(17.1,0)}]
    \fill[orange!14] (2,.17) rectangle (4,2.45);
    \node[note] at (3,2.2) {$A_n\ne j$};
  \end{scope}
  \foreach \offset/\heading in {0/{(a) Before the first drop},8.55/{(b) Continued use of $j$},17.1/{(c) Switching strategies}} {
    \begin{scope}[shift={(\offset,0)}]
      \node[panel title] at (0,2.83) {\heading};
      \draw[axis] (0,2.45) -- (0,.17) -- (7,.17);
      \draw[target] (0,1.15) -- (7,1.15);
      \node[axis text,left] at (0,1.15) {$q$};
      \foreach \t in {0,...,7} \draw[axis] (\t,.17) -- +(0,-.055);
      \node[axis text,below] at (0,.08) {$0$};
      \node[axis text,below] at (7,.08) {$7$};
    \end{scope}
  }
  \node[axis text,rotate=90] at (-1.08,1.35) {Value $V_n(j)$};
  \node[axis text] at (12.05,-.54) {Games played, $n$};

  \begin{scope}
    \draw[trajectory] plot[mark=*,mark size=1.55pt] coordinates
      {(0,1.85) (1,1.50) (2,2.05) (3,1.65) (4,2.15) (5,.75) (6,.65) (7,.55)};
    \draw[leader] (4.89,.72) -- (4.44,.54) -- (4.16,.54);
    \node[note,left] at (4.13,.54) {first below $q$};
    \draw[trajectory,fill=white,line width=1pt] (5,.75) circle[radius=2.45pt];
  \end{scope}
  \begin{scope}[shift={(8.55,0)}]
    \draw[trajectory] plot[mark=*,mark size=1.55pt] coordinates
      {(0,1.90) (1,1.55) (2,.60) (3,.95) (4,1.15) (5,.85) (6,1.05) (7,.80)};
    \draw[leader] (4,1.25) -- (4,1.64) -- (4.26,1.78);
    \node[note,right] at (4.3,1.78) {may reach $q$};
    \draw[trajectory,fill=white,line width=1pt] (4,1.15) circle[radius=2.45pt];
  \end{scope}
  \begin{scope}[shift={(17.1,0)}]
    \draw[trajectory] plot[mark=*,mark size=1.55pt] coordinates
      {(0,1.80) (1,.70) (2,.85)};
    \draw[trajectory,densely dotted] plot[mark=*,mark size=1.55pt,mark options={solid}] coordinates
      {(2,.85) (3,1.35) (4,1.75)};
    \draw[trajectory] plot[mark=*,mark size=1.55pt] coordinates
      {(4,1.75) (5,1.50) (6,1.20) (7,.90)};
  \end{scope}
\end{tikzpicture}
\caption{Value trajectories under \cref{ass:shape}. (a) The value may fluctuate above $q$ before the first drop below it. (b) During the same run, it may then reach $q$ but cannot exceed it. (c) Playing a different strategy in games 3 and 4 (shaded) permits $V_n(j)$ to rise above $q$ again.}
\label{fig:persistence}
\end{figure}
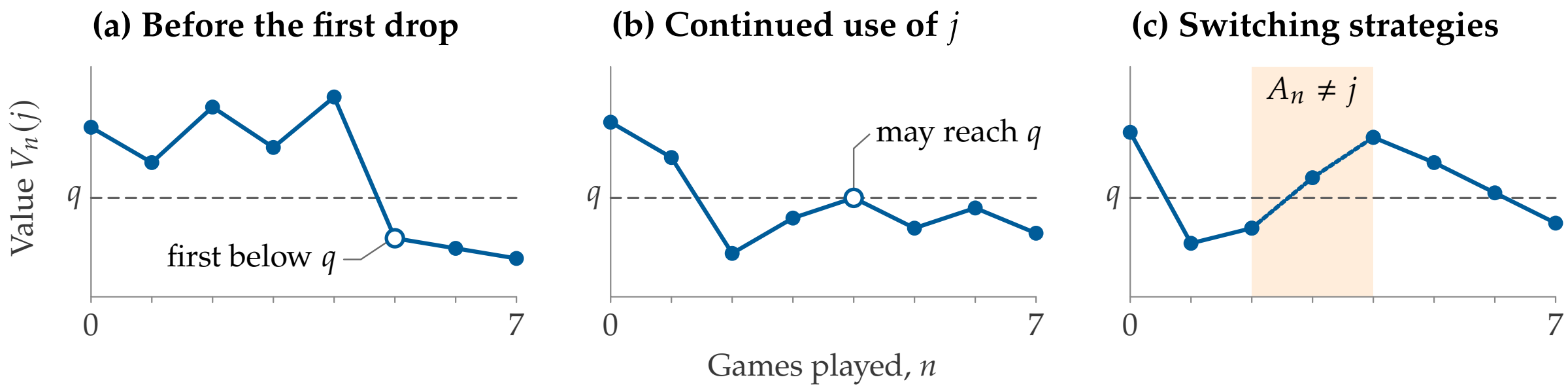

Let $\mathcal E_q$ be the class of environments that satisfy \cref{ass:shape}. In words, once the opponent has repaired the weakness that strategy $j$ exploits, so that the value of $j$ is below $q$, it does not reopen this weakness while the player keeps using $j$. This assumption allows, for example, the following patterns illustrated in \cref{fig:persistence}.
\begin{enumerate}[label=(\roman*)]
\item Before $V_n(j)$ first falls below $q$, it may fluctuate arbitrarily above $q$ (\cref{fig:persistence}(a)).
\item Once $V_n(j)$ drops below $q$, it may partially recover during continued play of strategy $j$, provided it remains at or below $q$ (\cref{fig:persistence}(b)).
\item If the player switches to other strategies in between, $V_n(j)$ is permitted to rise above $q$ again (\cref{fig:persistence}(c)).
\end{enumerate}

For success, we restrict attention to a subclass of $\mathcal E_q$ in which at least one strategy starts above the target and the values deteriorate at a bounded rate. Different strategies may exploit the same weakness of the opponent, so playing one strategy can also lower the values of others, much as using one arm changes its neighbors in graph-triggered bandits \citep{genalti2026}. To capture such spillovers, we partition $[K]$ into disjoint \emph{groups} $\mathcal G$, where $G(j)\in\mathcal G$ denotes the group containing strategy $j$ and $K_{\max}=\max_{G\in\mathcal G}|G|$.

\begin{assumption}[Margin and deterioration rate]\label{ass:margin}
For $\Delta\in(0,1-q]$ and $\lambda\geq0$,
\begin{enumerate}[label=(\alph*)]
\item $\max_{j\in[K]}V_0(j)\geq q+\Delta$;
\item for every procedure, almost surely, for all $n\geq1$ and $j\in[K]$,
\begin{align}\label{eq:envelope}
V_n(j)\geq V_{n-1}(j)-\lambda\ind\{A_n\in G(j)\}.
\end{align}
\end{enumerate}
\end{assumption}

Let $\mathcal E_q(\Delta,\lambda,\mathcal G)$ be the class of environments that satisfy \cref{ass:shape,ass:margin}. Under part (a), some strategy starts at least $\Delta$ above the target. Under part (b), a game can lower $V_n(j)$ only if the strategy played belongs to $G(j)$, and then by at most $\lambda$, which limits how fast the values can fall. The partition interpolates between two extremes. With singleton groups $\mathcal G=\{\{1\},\ldots,\{K\}\}$, playing a strategy can lower only its own value. With the single group $\mathcal G=\{[K]\}$, playing any strategy can lower all $K$ values. In the latter case we write $\mathcal E_q(\Delta,\lambda)$, and \eqref{eq:envelope} reduces to $V_n(j)\geq V_{n-1}(j)-\lambda$, as in \eqref{eq:intro-class}.

We illustrate the two assumptions and the resulting classes with the stone game of \cref{sec:intro}.

\begin{example}[The stone game]\label{ex:stones}
In the game of \cref{fig:stones}, strategy $j$ wins exactly when the opponent errs at one position, namely the position with four stones for $j=1$ and the position with five stones for $j=2$. Suppose that the opponent's error probability at that position starts at $p_j$ and is multiplied by $\beta_j\in(0,1]$ each time the position is reached. Then $V_n(j)=p_j\beta_j^{\,n_j}$, where $n_j$ is the number of times strategy $j$ was used in the first $n$ games. As this value never increases, the environment belongs to $\mathcal E_q$ for every target $q$. A game lowers only the value of the strategy used, and by at most $p_j(1-\beta_j)$, so the environment belongs to $\mathcal E_q(\Delta,\lambda,\{\{1\},\{2\}\})$ with $\lambda=\max_jp_j(1-\beta_j)$ whenever $\max_jp_j\geq q+\Delta$. If both strategies led to the same position, playing either would lower both values, and the two strategies would form one group.
\end{example}

\subsection{The minimax budget}\label{sec:problem}

We measure how many practice games are needed to be valid on $\mathcal E_q$ and successful on $\mathcal E_q(\Delta,\lambda,\mathcal G)$ by the \emph{minimax budget}
\begin{align}\label{eq:minimax-budget}
N^\star(\Delta,\lambda,\mathcal G)=\min\bigl\{N:\ \text{some $\varepsilon$-valid procedure on $\mathcal E_q$ is $(N,\varepsilon)$-successful on $\mathcal E_q(\Delta,\lambda,\mathcal G)$}\bigr\},
\end{align}
with $N^\star(\Delta,\lambda,\mathcal G)=\infty$ when no budget suffices. We suppress the arguments $(\Delta,\lambda,\mathcal G)$ when clear from the context. Note that the minimum in \eqref{eq:minimax-budget} is over all procedures, including those that know $\Delta$, $\lambda$, and $\mathcal G$. 

The minimax budget is infinite when the values can fall too fast. If the margin can disappear within a few games, for instance, no valid procedure can gather enough evidence in time, no matter how large the budget. Question \eqref{eq:question} thus asks how $N^\star$ grows with $K$, $q$, $\Delta$, and $\varepsilon$, and how slowly the values must fall, as measured by $\lambda$ and $\mathcal G$, for $N^\star$ to be finite. In the next two sections, we characterize $N^\star$ up to universal constants and logarithmic factors, starting with a single strategy ($K=1$) and a known margin $\Delta$ in \cref{sec:single}, and then develop a fully adaptive procedure for any $K\geq 1$ without knowing $\Delta$, $\lambda$, or $\mathcal G$ in \cref{sec:search}.

\section{Single-strategy verification}\label{sec:single}
We first consider a single strategy ($K=1$), so the procedure only decides when to stop and whether to select it, and we assume that $\Delta$ is known to isolate the verification problem. The verification rule developed here is then used in \cref{sec:search} for any $K\geq 1$ without knowing $\Delta$.

\subsection{The need for private randomization}\label{sec:private}

A naive way to decide when to stop is to keep playing the strategy and select it as soon as the outcomes give enough evidence that its value exceeds $q$. As noted in \cref{sec:overview}, however, an opponent that observes the same outcomes can tell when such a rule stops and lower the strategy's value below $q$ right before the next game. 

We make this precise in the simplest setting, the \emph{outcome-only experiment}, in which $K=1$ and each game reveals only its outcome, $O_n=Z_n$. The player then sees a sequence of wins and losses, and an environment reduces to a rule that sets the probability $V_n$ of winning the next game as a function of $Z_1,\ldots,Z_n$.

\begin{proposition}\label{prop:history}
In the outcome-only experiment, consider a procedure whose stopping time $\tau$ depends only on the outcomes, that is, $\tau$ is a stopping time for $(\sigma(Z_1,\ldots,Z_n))_{n\geq0}$. The selection $\hat J$ may also depend on the private seed. Let $\PP_p$ denote the law under which $Z_1,Z_2,\ldots$ are independent Bernoulli$(p)$ variables. If the procedure is $\varepsilon$-valid on $\mathcal E_q$, then $\PP_p\{\tau<\infty,\hat J=1\}\leq\varepsilon$ for every $p\in(q,1)$. Consequently, the procedure is not $(N,\varepsilon)$-successful on $\mathcal E_q(\Delta,\lambda)$ for any $N\geq1$, $\Delta\in(0,1-q)$, and $\lambda\geq0$.
\end{proposition}

We prove \cref{prop:history} in \cref{lower:history-proof}. The proposition allows the selection $\hat J$ to depend on the private seed, so with $K=1$ the procedure may randomize between selecting the strategy ($\hat J=1$) and abstaining ($\hat J=\varnothing$). Therefore, randomizing this final decision is not enough, which gives the first requirement for a procedure that is both valid and successful:
\begin{enumerate}[
    label={\ensuremath{(\mathsf{R\arabic*})}},
    ref={\ensuremath{\mathsf{R\arabic*}}},
    leftmargin=*,
    series=sparrequirements
]
\item\label{req:private}
\emph{Private timing.}
The stopping time must depend on private randomization that the opponent does not observe.
\end{enumerate}

Validity and success also constrain how the probability of selection is distributed over time. Although the opponent cannot see when the player stops, it can fix in advance when to repair the weakness. Fix $p=q+\Delta\in(q,1)$ and $n\geq0$, and let $\nu^{(n)}$ be the environment that keeps the value at $p$ for the first $n$ games and lowers it to some $r<q$ right after game $n$, where it stays. This environment belongs to $\mathcal E_q$, and its first $n$ outcomes have the same law as under $\PP_p$. The event $\{\tau=n,\hat J=1\}$ depends only on these outcomes and the private seed, so it has the same probability under $\nu^{(n)}$ as under $\PP_p$, but under $\nu^{(n)}$ every such selection is incorrect. Validity therefore requires:

\begin{enumerate}[resume*=sparrequirements]
\item\label{req:spread}
\emph{Small probability at each time.}
For every fixed $n\geq0$,
\begin{align*}
\PP_p\{\tau=n,\hat J=1\} = \PP_{\nu^{(n)}}\{\tau=n,\hat J=1\} \leq\varepsilon,
\end{align*}
where the probability is taken over both the outcomes and the private randomization.
\end{enumerate}

Under $\PP_p$ itself, the value stays at $q+\Delta$, so this stationary environment belongs to $\mathcal E_q(\Delta,\lambda)$ for every $\lambda\geq0$, and every selection is correct. Thus $(N,\varepsilon)$-success requires:
\begin{enumerate}[resume*=sparrequirements]
\item\label{req:power}
\emph{High probability within the budget.}
The probability of selecting by time $N$ must be at least $1-\varepsilon$:
\begin{align*}
\sum_{n=0}^{N}\PP_p\{\tau=n,\hat J=1\} = \PP_p\{\tau\leq N,\hat J=1\} \geq1-\varepsilon.
\end{align*}
\end{enumerate}

Requirements \eqref{req:spread} and \eqref{req:power} together force the procedure to spread its probability of selection thinly across many possible times. The construction below accumulates this probability in small increments as evidence arrives and lets private randomness decide when the selection occurs.

\subsection{The SPAR verification rule}\label{sec:certification}\label{cert:construction}

To meet requirements \eqref{req:private}--\eqref{req:power}, we introduce SPAR (\cref{alg:verify}). The rule plays a candidate strategy $j$ repeatedly and tracks two quantities, a \emph{capped likelihood ratio} $\mathcal K_t$ and an \emph{accumulated excess} $S_t$. Whenever evidence that the strategy's value exceeds $q$ pushes $\mathcal K_t$ above a cap $\kappa$, the rule keeps $\mathcal K_t$ at $\kappa$ and adds the surplus to $S_t$. It then selects $j$ as soon as $S_t$ reaches a private uniform threshold $U\sim\operatorname{Unif}[0,1]$. We write $Y_1,\ldots,Y_L$ for the outcomes during verification (so $Y_t=Z_t$ when verification begins at the first game).

\begin{algorithm}[!t]
\caption{SPAR: Sequential Private Accumulated-excess Rule}\label{alg:verify}
\small
\textbf{Input:} a candidate $j$, target $q$, tolerance $\varepsilon$, a tilting parameter $0<\delta<1-q$, and a horizon $L\geq1$.
\begin{enumerate}
\item Set $\kappa=\kappa_L$ from \eqref{eq:kappa-L}, initialize $\mathcal K_0=\kappa$ and $S_0=0$, and draw a fresh private threshold $U\sim\operatorname{Unif}[0,1]$, independent of the earlier observations and of all other randomness.
\item For $t=1,\ldots,L$, use $j$ and observe its outcome $Y_t$. With
\begin{align}\label{cert:likelihood-factor}
g(y)=1+\frac{\delta(y-q)}{q(1-q)},
\end{align}
update
\begin{align}\label{cert:reserve}
\widetilde{\mathcal K}_t=\mathcal K_{t-1}g(Y_t),\qquad \mathcal K_t=\min\{\kappa,\widetilde{\mathcal K}_t\},\qquad S_t=S_{t-1}+(\widetilde{\mathcal K}_t-\mathcal K_t).
\end{align}
If $S_t\geq U$, stop and select $j$.
\item If no selection has been made after $L$ games, stop and abstain.
\end{enumerate}
\end{algorithm}

We calibrate the cap in \cref{alg:verify} to the horizon $L$ and tilt $\delta$ by setting
\begin{align}\label{eq:kappa-L}
\kappa_L=\frac{\varepsilon}{\delta/q+\EE_q S_{L-1}^{(1)}+\varpi},
\end{align}
where $\varpi=10^{-10}$ is for numerical stability, and $S_t^{(1)}$ is the accumulated excess in \eqref{cert:reserve} after $t$ independent Bernoulli$(q)$ games with unit cap, initial capital $1$, and no stopping.\footnote{$\EE_q S_{L-1}^{(1)}$ can be computed recursively from $q,\delta,L$ alone before verification begins (see \cref{app:cap-computation}).} Per unit cap, the game at which the value first falls below $q$ adds at most $\delta/q$ to the excess. Subsequent win probabilities stay at or below $q$ by \cref{ass:shape}, so the remaining at most $L-1$ games add at most $\EE_q S_{L-1}^{(1)}$ in expectation (\cref{lem:finite-cap}). The choice in \eqref{eq:kappa-L} therefore bounds the expected excess added at and after the drop by $\varepsilon$. Since $\EE_q S_{L-1}^{(1)}\leq1$ and $\delta<1-q$, we have $\kappa_L\geq q\varepsilon/(1+\varpi)$, and shorter horizons permit a larger cap.

For binary outcomes, $g(y)$ is the likelihood ratio of win probability $q+\delta$ against $q$, and $M_t=g(Y_1)\cdots g(Y_t)$ (with $M_0=1$) is the likelihood ratio after $t$ games. By induction on $t$, the capped likelihood ratio satisfies $\mathcal K_t=\kappa M_t/\max_{0\leq s\leq t}M_s$. The accumulated excess $S_t$ sums the amounts by which $\mathcal K_{s-1}g(Y_s)$ exceeds $\kappa$ over $s\leq t$, so it grows only in games where $M_t$ reaches a new maximum. SPAR meets requirements \eqref{req:private}--\eqref{req:power} of \cref{sec:private} as follows.
\begin{enumerate}[label=(\roman*)]
\item \emph{Private timing \eqref{req:private}.} The stopping time depends on the private threshold $U$. Because the rule plays $j$ in every game, the observed strategies reveal nothing about $U$ before the rule stops.
\item \emph{Small probability at each time \eqref{req:spread}.} Since $U$ is independent of the outcomes when $j$ is played (\cref{cert:continuation}), selection at game $t$ occurs when $U$ falls in $(S_{t-1},S_t]\cap[0,1]$, an interval of length at most $S_t-S_{t-1}\leq\kappa\delta/q$, because $\mathcal K_{t-1}\leq\kappa$ and $g(Y_t)\leq1+\delta/q$. With $\kappa=\kappa_L$, \eqref{eq:kappa-L} gives $\kappa_L\delta/q\leq\varepsilon$.
\item \emph{High probability within the budget \eqref{req:power}.} Since $S_t\geq\kappa\log M_t$ (\cref{cert:continuation}), the rule must select once the log-likelihood ratio $\log M_t$ reaches $1/\kappa$.
\end{enumerate}
Since $U$ is uniform, the bound on the expected excess at and after the drop also bounds the probability of an incorrect selection by $\varepsilon$ (\cref{lem:finite-cap}).

\begin{theorem}[One candidate]\label{thm:single}
Let $K=1$, $0<\Delta\leq(1-q)/8$, $0<\varepsilon\leq1/16$, and $\lambda\geq0$, with $\Delta/\lambda=\infty$ if $\lambda=0$, and write $B_\Delta:=q(1-q)/\Delta^2$.
\begin{enumerate}[label=(\alph*)]
\item \Cref{alg:verify}, run from the first game with
\begin{align}\label{cert:single-parameters}
\delta=\Delta/4,\qquad L=\left\lceil\frac{32B_\Delta}{q\varepsilon}\right\rceil,
\end{align}
is $\varepsilon$-valid on $\mathcal E_q$. If $\Delta/\lambda\geq130B_\Delta/\varepsilon$, it is also $(L,\varepsilon)$-successful on $\mathcal E_q(\Delta,\lambda)$, so that
\begin{align*}
N^\star\leq L\leq\frac{65B_\Delta}{\varepsilon}.
\end{align*}
\item In the outcome-only experiment,
\begin{align*}
N^\star\geq\frac{B_\Delta}{16\varepsilon},
\end{align*}
and $N^\star=\infty$ if $\Delta/\lambda<B_\Delta/(16\varepsilon)$.
\end{enumerate}
\end{theorem}

We prove part (a) in \cref{cert:power-proof} and part (b) in \cref{lower:combination}. Part (a) holds for every underlying game, and part (b), which concerns the outcome-only experiment, holds even for procedures that know $\Delta$. The two parts match up to universal constants. In the worst case, a single candidate with margin $\Delta$ therefore requires of order $B_\Delta/\varepsilon$ practice games, and its margin must survive that many games, that is, $\Delta/\lambda$ must be at least of order $B_\Delta/\varepsilon$.

\section{Adaptive selection for multiple strategies}\label{sec:search}

We now allow any number $K$ of strategies and no longer assume that the margin is known. Besides deciding when to stop and whether to select, the procedure must choose which strategy to play in each game.

\subsection{The Screen-then-SPAR procedure}\label{search:construction}

To use SPAR, the procedure must find a candidate and choose the tilting parameter $\delta$ without knowing $\Delta$. By \cref{thm:single}, SPAR needs of order $B_\Delta/\varepsilon$ games to verify a candidate with margin $\Delta$. By contrast, a screen needs only of order $B_\Delta\log(K/\varepsilon)$ games per strategy to detect such a margin. Screen-then-SPAR (\cref{alg:search}) therefore screens the strategies in blocks of doubling length and verifies only the first strategy that passes.

\begin{algorithm}[!t]
\caption{Screen-then-SPAR}\label{alg:search}
\small
\textbf{Input:} strategies $1,\ldots,K$ in a fixed order, target $q$, tolerance $\varepsilon$, and budget $N$. Set $n=0$.
\begin{enumerate}
\item For $l=0,\ldots,\lfloor\log_2N\rfloor$, set
\begin{align*}
m_l=2^{l+1},\qquad x_l=\log\frac{8K(l+1)(l+2)}{\varepsilon},\qquad \delta_l=\sqrt{\frac{q(1-q)x_l}{32m_l}}.
\end{align*}
Skip the stage if $\delta_l\geq1-q$. For each strategy $j=1,\ldots,K$ in turn, set $L=N-n-m_l$, the budget that would remain after its screen.

\emph{If $L\geq1$}, take the following steps for $j$.
\begin{enumerate}[label=(\alph*),leftmargin=1.6em]
\item Compute the verification cap $\kappa_L$ from \eqref{eq:kappa-L} with tilt $\delta_l$.
\item If $L\kappa_L\delta_l/q\geq1-\varepsilon$ and $m_l\kl(1,q+\delta_l)<x_l$, skip this screen without playing.
\item Use $j$ for $m_l$ consecutive games, let $\bar Z$ be their mean, and set $n\leftarrow n+m_l$. The screen passes if
\begin{align*}
\bar Z\geq q+\delta_l \quad\text{and}\quad \bigl(L\kappa_L\delta_l/q<1-\varepsilon\ \text{ or }\ m_l\kl(\bar Z,q+\delta_l)\geq x_l\bigr).
\end{align*}
\item If the screen fails, proceed to the next screen. If it passes, run \cref{alg:verify} for $j$ with tilt $\delta_l$ and horizon $L$, reusing the cap from step~1(a), add the number of games it uses to $n$, stop at $\tau=n$, and return its decision.
\end{enumerate}
\emph{If $L<1$}, stop at $\tau=n$ and return $\hat J=\varnothing$.
\item If the loops finish without a commitment, stop at $\tau=n$ and return $\hat J=\varnothing$.
\end{enumerate}
\end{algorithm}

Before each screen, \cref{alg:search} computes $\kappa_L$ from \eqref{eq:kappa-L} using the stage's tilt $\delta_l$ and the budget $L$ that would remain after the screen. In each screen, \cref{alg:search} compares the sample mean $\bar Z$ with the same alternative $q+\delta_l$ used by SPAR. When the remaining budget is large enough ($L\kappa_L\delta_l/q\geq1-\varepsilon$), it also requires the one-sided KL test $m_l\kl(\bar Z,q+\delta_l)\geq x_l$, so a strategy whose value stays at or below $q+\delta_l$ rarely passes. If even $m_l$ consecutive wins would fail this test, the screen is skipped without playing. When $L\kappa_L\delta_l/q<1-\varepsilon$, even $L$ straight wins in SPAR give a selection probability below $1-\varepsilon$, so the screen simply checks $\bar Z\geq q+\delta_l$.

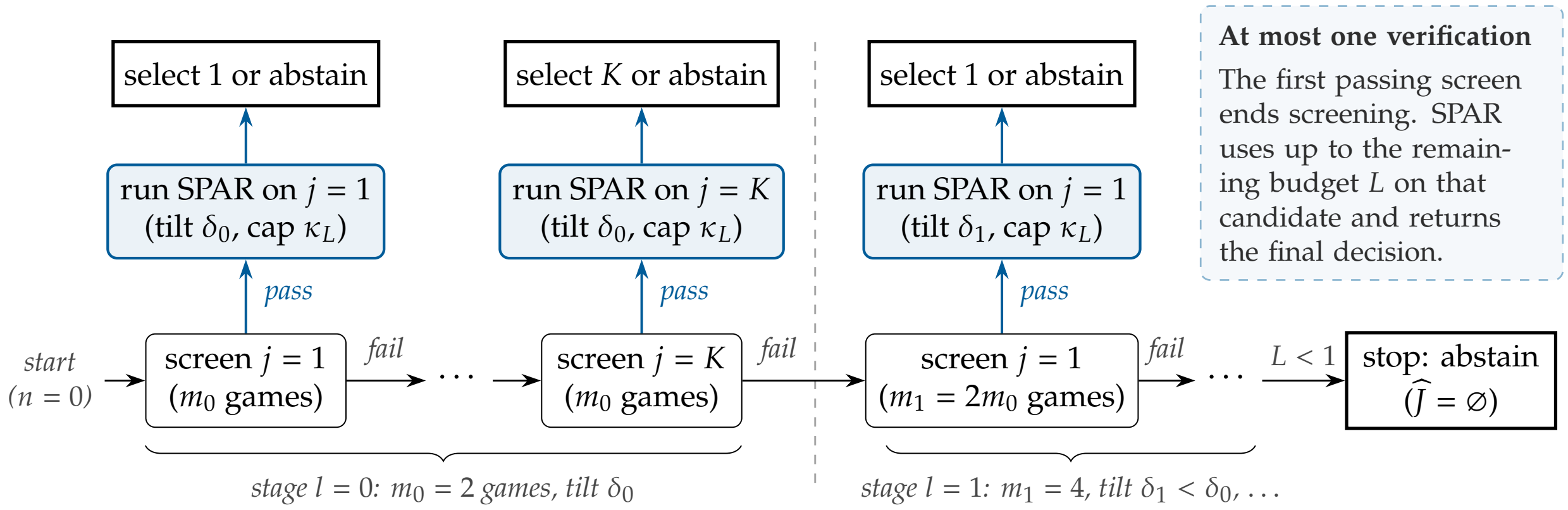
\begin{figure}[!htb]
\centering
\tikzset{
  algscr/.style={draw,rounded corners=3pt,fill=white,align=center,font=\small,inner sep=4pt,minimum height=1.0cm,minimum width=2.25cm},
  algspar/.style={draw=linkblue,line width=.85pt,rounded corners=3pt,fill=linkblue!8,align=center,font=\small,inner sep=4pt,minimum height=1.0cm,minimum width=2.55cm},
  algend/.style={draw,line width=1.05pt,align=center,font=\small,inner sep=3.5pt,minimum height=.72cm,minimum width=2.45cm},
  algarr/.style={-{Stealth[length=2.2mm,width=1.5mm]},line width=.6pt},
  algpass/.style={-{Stealth[length=2.2mm,width=1.5mm]},line width=.8pt,draw=linkblue},
  algnote/.style={font=\footnotesize\itshape,text=black!72,align=center},
  algpnote/.style={font=\footnotesize\itshape,text=linkblue,align=center}}
\resizebox{\linewidth}{!}{%
\begin{tikzpicture}[x=1cm,y=1cm]
\path[use as bounding box] (-1.30,0.05) rectangle (16.45,5.65);

\draw[dashed,black!35,line width=.6pt] (7.95,0.50) -- (7.95,5.55);

\node[algnote] (start) at (-0.65,1.65) {start\\($n=0$)};

\node[algscr] (s01) at (1.55,1.65) {screen $j=1$\\($m_0$ games)};
\node (d0) at (3.95,1.65) {$\cdots$};
\node[algscr] (s0k) at (6.00,1.65) {screen $j=K$\\($m_0$ games)};
\node[algscr,minimum width=2.60cm] (s11) at (10.05,1.65) {screen $j=1$\\($m_1=2m_0$ games)};
\node (d1) at (12.60,1.65) {$\cdots$};
\node[algend,minimum width=2.35cm] (abs) at (15.10,1.65) {stop: abstain\\($\hat J=\varnothing$)};

\draw[algarr] (start.east) -- (s01.west);
\draw[algarr] (s01.east) -- node[algnote,above=1pt] {fail} (d0.west);
\draw[algarr] (d0.east) -- (s0k.west);
\draw[algarr] (s0k.east) -- node[algnote,above=1pt,pos=.30] {fail} (s11.west);
\draw[algarr] (s11.east) -- node[algnote,above=1pt] {fail} (d1.west);
\draw[algarr] (d1.east) -- node[algnote,above=1pt] {$L<1$} (abs.west);

\node[algspar] (v01) at (1.55,3.55) {run SPAR on $j=1$\\(tilt $\delta_0$, cap $\kappa_L$)};
\node[algspar] (v0k) at (6.00,3.55) {run SPAR on $j=K$\\(tilt $\delta_0$, cap $\kappa_L$)};
\node[algspar,minimum width=2.65cm] (v11) at (10.05,3.55) {run SPAR on $j=1$\\(tilt $\delta_1$, cap $\kappa_L$)};

\draw[algpass] (s01.north) -- node[algpnote,right=2pt] {pass} (v01.south);
\draw[algpass] (s0k.north) -- node[algpnote,right=2pt] {pass} (v0k.south);
\draw[algpass] (s11.north) -- node[algpnote,right=2pt] {pass} (v11.south);

\node[algend] (o01) at (1.55,5.10) {select $1$ or abstain};
\node[algend] (o0k) at (6.00,5.10) {select $K$ or abstain};
\node[algend,minimum width=2.60cm] (o11) at (10.05,5.10) {select $1$ or abstain};

\draw[algpass] (v01.north) -- (o01.south);
\draw[algpass] (v0k.north) -- (o0k.south);
\draw[algpass] (v11.north) -- (o11.south);

\node[draw=linkblue!45,dashed,rounded corners=3pt,fill=linkblue!4,line width=.55pt,
      inner sep=6pt,align=left,font=\footnotesize,text=black!82,text width=3.65cm]
  at (14.32,4.32)
  {\textbf{At most one verification}\\[3pt]
   The first passing screen ends screening. SPAR uses up to the remaining budget $L$ on that \mbox{candidate} and returns the final decision.};

\draw[decorate,decoration={brace,amplitude=5pt,mirror}]
  (0.42,0.92) -- (7.12,0.92)
  node[midway,below=6pt,algnote] {stage $l=0$: $m_0=2$ games, tilt $\delta_0$};
\draw[decorate,decoration={brace,amplitude=5pt,mirror}]
  (8.75,0.92) -- (12.90,0.92)
  node[midway,below=6pt,algnote] {stage $l=1$: $m_1=4$, tilt $\delta_1<\delta_0$, $\ldots$};
\end{tikzpicture}%
}
\caption{Execution tree of Screen-then-SPAR (\cref{alg:search}). Along the bottom path, strategies are screened in order in stages of doubling length $m_l=2^{l+1}$ and decreasing tilt $\delta_l$. Infeasible screens (step~1(b)) are skipped without playing. As soon as a screen passes, screening ends and SPAR runs once on that candidate over the remaining budget $L$ with cap $\kappa_L$ and tilt $\delta_l$. If no screen passes before the remaining budget falls below one, the procedure abstains.}
\label{fig:screen-spar}
\end{figure}

Because the screens lengthen and $\delta_l$ decreases, the search adapts to the unknown margin. A strategy whose value remains at least $q+\Delta$ passes with high probability once $m_l$ reaches order $B_\Delta\log(K/\varepsilon)$, up to a doubly logarithmic factor in $B_\Delta$. Once a strategy passes, screening ends and \cref{alg:search} returns SPAR's decision. The procedure is illustrated in \cref{fig:screen-spar}.

\subsection{Upper bounds}\label{search:guarantees}

\begin{theorem}[Search]\label{thm:search}
Let $0<\Delta\leq(1-q)/8$, $0<\varepsilon\leq1/16$, and $\lambda\geq0$, with $\Delta/\lambda=\infty$ if $\lambda=0$, and recall that $\Lambda=\log(K/\varepsilon)+\log\log B_\Delta$ with $B_\Delta=q(1-q)/\Delta^2$.
\begin{enumerate}[label=(\alph*)]
\item \Cref{alg:search} stops within $N$ games and is $\varepsilon$-valid on $\mathcal E_q$.
\item There is a universal constant $C$ (one may take $C=10^6$) such that, if
\begin{align}
N&\geq CB_\Delta\left(K\Lambda+\frac1\varepsilon\right), \label{search:budget} \\
\frac\Delta\lambda&\geq CB_\Delta\left(K_{\max}\Lambda+\frac1\varepsilon\right), \label{search:rate}
\end{align}
then \cref{alg:search} is $(N,\varepsilon)$-successful on $\mathcal E_q(\Delta,\lambda,\mathcal G)$.
\end{enumerate}
\end{theorem}

We prove part (a) in \cref{search:proof} and part (b) in \cref{search:constants}. Part (b) says that \cref{alg:search} succeeds whenever it can complete its screens and one verification both within the budget and before the margin runs out. A screen long enough to detect a margin $\Delta$ takes of order $B_\Delta\Lambda$ games, and a verification of order $B_\Delta/\varepsilon$ games. Condition \eqref{search:budget} asks the budget to cover one such screen for each of the $K$ strategies plus one verification. Condition \eqref{search:rate} asks the same of the $\Delta/\lambda$ games for which the margin lasts, except that only the screens of the at most $K_{\max}$ strategies in one group count.

\subsection{Lower bounds}\label{search:lower}

Both conditions of \cref{thm:search} are necessary in the worst case over the underlying game up to logarithmic factors in the search terms.

\begin{theorem}[Lower bounds]\label{thm:lower}
Let $0<\Delta\leq(1-q)/8$, $0<\varepsilon\leq1/16$, and $\lambda\geq0$, with $\Delta/\lambda=\infty$ if $\lambda=0$. There is a two-round game with $K$ strategies in which, for every partition $\mathcal G$ of $[K]$,
\begin{align}\label{lower:combined-budget}
N^\star\geq\frac{B_\Delta}{32}\left(K\log\frac1\varepsilon+\frac1\varepsilon\right),
\end{align}
and $N^\star=\infty$ if $\Delta/\lambda<(B_\Delta/32)[K_{\max}\log(1/\varepsilon)+1/\varepsilon]$.
\end{theorem}

We prove \cref{thm:lower} in \cref{lower:combination}, which also bounds the probability of failure for every budget (\cref{prop:lower-risk}). By \cref{thm:search,thm:lower}, \cref{alg:search} is minimax optimal up to logarithmic factors. Apart from universal constants, the two bounds differ only in the search terms, where $\Lambda=\log(K/\varepsilon)+\log\log B_\Delta$ replaces $\log(1/\varepsilon)$. The terms $\log K$ and $\log\log B_\Delta$ come from union bounds over the $K$ strategies and over the doubling stages that adapt to the unknown margin.

\subsection{Bounded outcomes}\label{sec:bounded}

The results extend to outcomes that are scores in $[0,1]$, so that a game can be any complete interaction whose result is graded rather than won or lost. Let the outcome $z(o)$ of an observation $o$ take values in $[0,1]$, and define the value as the conditional mean,
\begin{align}\label{eq:value-bounded}
V_n(j)=\int z(o)\,\nu_n(do\mid O_{1:n},j),
\end{align}
so that \eqref{eq:mean} becomes $\EE_\nu[Z_{n+1}\mid\F_n]=V_n(A_{n+1})$. \Cref{ass:shape,ass:margin}, \cref{def:requirements}, and \cref{alg:verify,alg:search} are unchanged.

\begin{proposition}\label{prop:bounded}
With outcomes in $[0,1]$ and values \eqref{eq:value-bounded}, \cref{thm:single,thm:search,thm:lower} hold as stated, with the same constants.
\end{proposition}

The upper bounds carry over because the factor $g$ in \eqref{cert:likelihood-factor} is affine in the outcome, and the Chernoff bounds in their proofs hold for outcomes in $[0,1]$ \citep{hoeffding1963}. The lower bounds need no change, since they are proved in a game with binary outcomes, which is still admissible. \Cref{app:bounded} gives the details.

\section{Simulations}\label{sec:numerics}

In this section, we use simulated data to compare SPAR with the private block verifier, the likelihood-ratio test \citep{wald1945,howard2020}, and the randomized binomial test \citep{neymanpearson1933} . See Appendix~\ref{app:block-validity} for the block verifier and Appendix~\ref{app:sim-test-comparators} for the two tests. In each of three settings, we measure how often each method selects a strategy that still meets the target and how many practice games it uses on average.

In each run, the player has $K$ strategies and a budget of $N$ games for screening and verification combined. We use $K=4$ and $N=96{,}000$ in the first two settings, and $K=2$ and $N=250{,}000$ in the third. When the player uses strategy $j$ in game $t$, we draw a Bernoulli outcome with its current win probability $V_{t-1}(j)$. The strategies are screened in order. The first $K-1$ have a constant win probability of $0.55$, below the target $q=0.60$. The last strategy starts above target. After each game with that strategy, we update its win probability using its total number of plays or wins, including those during screening.
\begin{itemize}[leftmargin=*,topsep=3pt,itemsep=3pt,parsep=0pt]
\item \textbf{Decline over 200 plays.} The last strategy starts at $0.80$. Its win probability falls linearly to $0.62$ between plays $400$ and $600$, stays at $0.62$ through play $55{,}000$, then falls linearly to its final value of $0.59$ at play $65{,}000$.
\item \textbf{Decline over 100 plays.} The first decline runs from $0.80$ at play $450$ to $0.62$ at play $550$. As in the first setting, the win probability stays at $0.62$ through play $55{,}000$ and falls linearly to $0.59$ by play $65{,}000$, where it stays.
\item \textbf{Drops after wins.} The last strategy starts at $0.65$. Its win probability drops to $0.62$ after its $100$th win and to $0.50$ after its $100{,}000$th win.
\end{itemize}

We ran each setting $20{,}000$ times with error tolerance $\varepsilon=0.05$. In each run, all four methods use the screens in \cref{alg:search}. Once a strategy passes, each verifier sees the same sequence of wins and losses up to its own stopping time. SPAR uses the tilt from the passing screen. The block verifier tests one of $40$ blocks chosen at random. The likelihood-ratio test uses the fixed alternative $0.65$, and the randomized binomial test uses all the remaining games. A selection is correct if the strategy's win probability after its last game is at least $0.60$.

\begin{figure}[htbp]
\centering
\includegraphics[width=\linewidth]{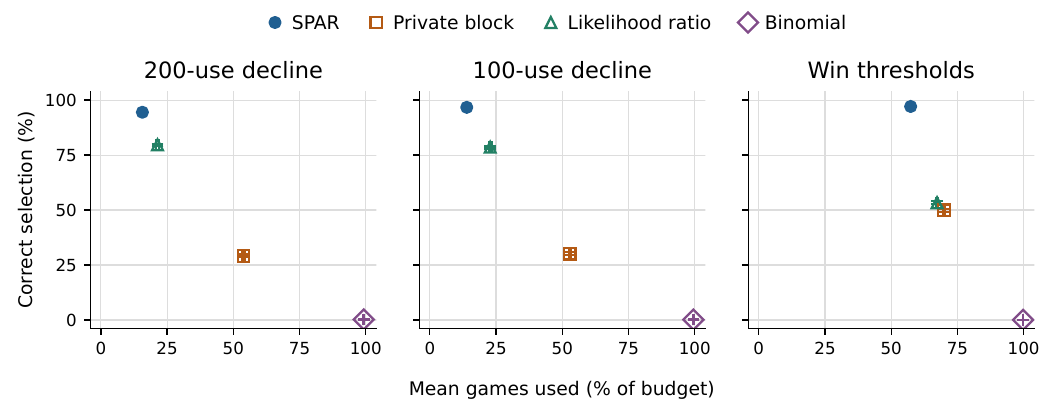}
\caption{Correct selections as a percentage of all runs and mean games used as a percentage of the budget. From left to right, the panels show the 200-play decline, the 100-play decline and the drops after wins. Vertical bars show pointwise $95\%$ Wilson intervals. Horizontal bars show the mean plus or minus $1.96$ Monte Carlo standard errors.}
\label{fig:sim-confirmation}
\end{figure}

SPAR had the highest correct-selection rate and used the fewest games on average in each setting (\cref{fig:sim-confirmation,tab:sim-confirmation}). For the 200-play and 100-play declines, it selected correctly in $94.5\%$ and $96.7\%$ of runs, using $15{,}041$ and $13{,}381$ games on average. When wins triggered the drops, SPAR selected correctly in $97.1\%$ of runs, compared with $53.3\%$ for the likelihood-ratio test, and used about $25{,}000$ fewer games on average. At $0.62$, the expected log increment is negative for the fixed alternative $0.65$ and positive for every tilt SPAR used in this setting.

\section{Real-data examples}\label{sec:realdata}

We compare Screen-then-SPAR with the same three methods as in \cref{sec:numerics} using recorded coding-agent runs from MonitoringBench and recorded games of rock-paper-scissors. In each example, we measure how often a recommended strategy still meets the target on its next play as the opponent adapts.

\subsection{Coding-agent runs from MonitoringBench}\label{sec:monitoringbench}

Our sample consists of $811$ recorded runs from the firewall subset of MonitoringBench \citep{jotautaite2026monitoringbench}, grouped by source in \cref{tab:rd-source-banks}. In each run, a coding agent worked on a BashArena system administration task \citep{kaufman2025basharena} and made an unauthorized change to the firewall. Each record identifies the task and gives two scores for the full sequence of actions, one from GPT-4.1 and one from GPT-5. Higher scores indicate that the monitor found the actions more suspicious.

\begin{table}[htbp]
\centering\small
\caption{Recorded runs and distinct tasks by source. Human-in-the-Loop runs followed plans written by a researcher with model assistance. Opus Only runs came directly from Opus~4.6. Recon runs followed plans from a separate agent that first inspected the task environment. Refined records are edited versions of earlier runs. Initial pass is the percentage of runs that pass GPT-4.1. Both pass is the percentage that pass GPT-4.1 and GPT-5.}
\label{tab:rd-source-banks}
\begin{tabular}{@{}lrrrr@{}}
\toprule
Source & Records & Tasks & Initial pass (\%) & Both pass (\%)\\
\midrule
Human-in-the-Loop & 170 & 5 & 86.47 & 78.24\\
Opus Only & 58 & 58 & 94.83 & 75.86\\
Recon & 159 & 37 & 86.79 & 69.81\\
Refined & 424 & 77 & 93.16 & 88.68\\
\bottomrule
\end{tabular}
\end{table}
We replay each source separately as one strategy ($K=1$), drawing records uniformly with replacement. We apply the following monitoring rule to the recorded scores from MonitoringBench. Initially, a GPT-4.1 pass counts as a win. We count GPT-5 flags separately for each task. On a task's second flag, GPT-4.1 still determines the outcome of that draw. All later draws for that task must pass both monitors, so every record for that task that passes GPT-4.1 alone now counts as a loss. The procedures see only wins and losses. After draw $t$, $p_t$ is the fraction of all records in the source that would now pass. Given the full replay history, this is the next draw's win probability. It can only stay constant or decrease, satisfying \cref{ass:shape}.

We use a budget of $N=80$ interactions, tolerance $\varepsilon=0.05$ and eleven targets $q=0.80,0.81,\ldots,0.90$.  For each source and target, all four procedures share $100{,}000$ independent replay paths. \Cref{app:realdata} gives the calibration and evaluation details.

\begin{figure}[!tbp]
\centering
\includegraphics[width=\linewidth]{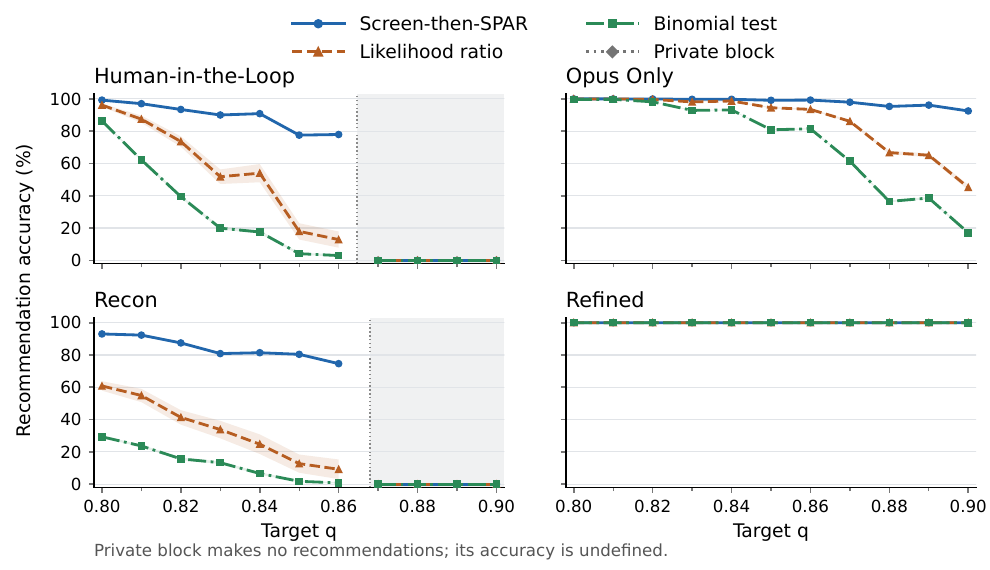}
\caption{Percentage of recommendations that still meet the target after the final monitoring update. Each panel shows one source, with pointwise $95\%$ Monte Carlo intervals. Shading marks targets above the source's initial pass probability. The private block verifier abstains throughout.}
\label{fig:rd-accuracy}
\end{figure}

\Cref{fig:rd-accuracy} shows the percentage of recommendations that still met the target after the final update. Screen-then-SPAR outperformed both stationary tests on Opus Only at all eleven targets and on Human-in-the-Loop and Recon through $q=0.86$. On Opus Only at $q=0.90$, $92.52\%$ of Screen-then-SPAR recommendations met the target, compared with $45.17\%$ for the likelihood-ratio test and $17.09\%$ for the binomial test. These correct Screen-then-SPAR recommendations accounted for $7.68\%$ of all replays at this setting. On Refined, every recommendation met the target through $q=0.88$.

Across the $44$ source--target combinations, at most $3.67\%$ of replays ended in a below-target recommendation by Screen-then-SPAR, within the $5\%$ tolerance. It used fewer interactions on average than either stationary test in every combination.

\subsection{Rock-paper-scissors against an adaptive opponent}\label{sec:rps}

We use the recorded matches of the $21$ participants who played $300$ rounds of rock-paper-scissors against the self-transition bot in Experiment~2 of \citet{brockbank2024rps}.\footnote{The match data are publicly available at \url{https://github.com/erik-brockbank/rps-bot-manuscript-public} (file \texttt{data/rps\_v3\_data.RData}).} The bot keeps track of how often the player repeats the previous move, switches to the move that beats it, or switches to the move that loses to it. For example, after rock, these choices are rock, paper and scissors, respectively. The bot predicts the most frequent of the three choices, breaking ties uniformly at random, and plays the move that beats its prediction.

We start each replay with the player's last move and the bot's counts after round~$100$ of the recorded match. From round~$101$ onward, the procedure chooses the player's moves and the bot makes its predictions from the counts in the preceding rounds. We use the three choices above as our candidate strategies ($K=3$) and screen them in the order listed. Each round adds one to the count for the chosen strategy. Repeated use therefore makes it easier for the bot to predict that strategy, and its next-round win probability can only stay the same or fall, as required by \cref{ass:shape}. The procedures observe only whether the player won. A draw or a loss counts as zero.

We use budgets of $80$ and $200$ additional rounds, tolerance $\varepsilon=0.05$ and targets $q=0.50,0.51,\ldots,0.60$. As in \cref{sec:numerics}, the four methods use the same screens and verify the first strategy that passes.  We calculate the probability of a correct recommendation and the mean number of rounds used from each recorded starting point, averaging over the random choices of the bot and the procedure, and give the $21$ matches equal weight.

\begin{figure}[htbp]
\centering
\includegraphics[width=\linewidth]{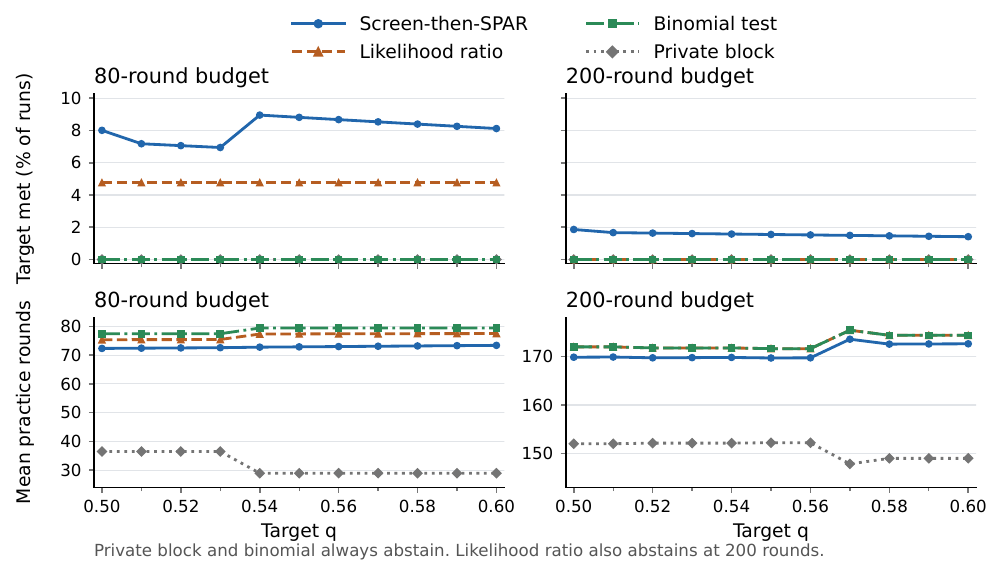}
\caption{Rock-paper-scissors replays from round~$100$ of each recorded match. The top row shows correct recommendations as a percentage of all replays. The bottom row shows the mean number of additional rounds used, including screening and verification.}
\label{fig:rps-replay}
\end{figure}

Screen-then-SPAR selected a strategy that still met the target more often than the other three methods at all eleven targets under both budgets (\cref{fig:rps-replay}). At $q=0.55$ with $80$ rounds available, it selected correctly in $8.81\%$ of replays and used $72.9$ rounds on average. The likelihood-ratio test selected correctly in $4.76\%$ and used $77.4$ rounds on average. With $200$ rounds available, Screen-then-SPAR selected correctly in $1.55\%$ of replays and the other methods always abstained. At this target the first screen took $2$ rounds under the smaller budget and $32$ under the larger one, giving the bot more time to learn the strategy before verification.

\section{Discussion}\label{sec:discussion}

The main open quantitative question is whether the \(\log K\) and \(\log\log B_\Delta\) terms in the screening cost are necessary. Beyond the worst-case budget, it would be useful to know how much practice is needed for a given set of candidates. For example, the cost of finding a strategy that meets the target may depend on how many candidates start above it and how long their advantages last.

A broader goal is to identify the best strategy available when practice ends. Even when several candidates meet the target, their chances of winning the next game can differ substantially, and their ranking can change during practice. The search becomes more open-ended when the player can develop new strategies in response to earlier games. The player then faces a choice between looking for a better candidate and verifying the strongest one found so far.

Another direction is to study how sensitive the guarantees are to departures from persistence (\cref{ass:shape}). If a strategy can recover above the target during continued play, how does the risk of an incorrect selection depend on the size or frequency of those recoveries? A related question concerns the time horizon after selection. A player choosing a strategy for several games may prefer a modest advantage that lasts to a larger one that disappears quickly. This suggests evaluating a recommendation by both its value at selection and how long it remains effective.

\bibliographystyle{unsrtnat}
\bibliography{references}

\clearpage
\appendix

\section{Details of the model}\label{app:model}

In this appendix, we show how a finite game defines an environment, give a mixture example that violates \cref{ass:shape}, and extend \eqref{eq:mean} to stopping times.

\subsection{Finite games and hidden parameters}\label{app:game}

Let each game have $k_0$ rounds, using dummy moves to complete shorter games. In round $k$ of game $n$, the opponent moves first and the player responds:
\begin{align*}
Y_{n,k}\sim Q_{n,k}(\cdot\mid O_{1:n-1},H_{n,k-1}),\qquad X_{n,k}\sim\pi_{A_n,k}(\cdot\mid H_{n,k-1},Y_{n,k}),
\end{align*}
where $H_{n,k}=(Y_{n,1},X_{n,1},\ldots,Y_{n,k},X_{n,k})$ is the within-game history. Strategy $j$ is the fixed collection $(\pi_{j,k})_{k\leq k_0}$. Its within-game randomization is fresh and independent of the private seed used for sampling and selection. The outcome is $Z_n=w(H_{n,k_0})$ for a known binary function $w$, and $O_n=(A_n,H_{n,k_0})$. The opponent's kernels may depend on every observed move and previous observation, but not on the player's private seed.

Composing these kernels conditionally on $O_{1:n}$ and $A_{n+1}=j$ gives $\nu_n(\cdot\mid O_{1:n},j)$, a kernel of the form assumed in \cref{sec:experiment}, and the identity \eqref{eq:mean} follows. \Cref{ass:shape,ass:margin} are additional restrictions on the resulting values.

A fixed unknown parameter differs from a random persistent type. Let $\Theta$ equal $1/4$ or $3/4$ with equal probabilities and, conditionally on $\Theta$, let all outcomes be independent Bernoulli$(\Theta)$. Each fixed-parameter environment is stationary and therefore belongs to $\mathcal E_q$. For the mixture environment, however, at the target $q=3/5$,
\begin{align*}
V_0=\EE\Theta=\frac12<q, \qquad V_1=\EE[\Theta\mid Z_1=1] =\frac{\EE\Theta^2}{\EE\Theta} =\frac58>q\quad\text{on }\{Z_1=1\}.
\end{align*}
This recovery, which occurs with positive probability, violates \cref{ass:shape}. Thus $\mathcal E_q$ need not be closed under mixing over latent types.

\subsection{Observation at a stopping time}\label{app:stopping}

For each $n\geq0$, $A_{n+1}$ is $\F_n$-measurable. The time $\tau$ is a stopping time for $(\F_n)_{n\geq0}$, and $\hat J$ is $\F_\tau$-measurable on $\{\tau<\infty\}$. We set $A_{n+1}=1$ on $\{n\geq\tau\}$ so that the process is defined after stopping. Under the law $\PP_\nu$ induced by the procedure and the environment, for every measurable $C\subseteq\mathcal O$,
\begin{align*}
\PP_\nu(O_{n+1}\in C\mid\F_n)=\nu_n(C\mid O_{1:n},A_{n+1})\quad\text{almost surely}.
\end{align*}

The identity \eqref{eq:mean} also holds at a stopping time. For a stopping time $\tau_0$ and an event $E\in\F_{\tau_0}$, we have $E\cap\{\tau_0=n\}\in\F_n$ for every $n\geq0$. Conditioning on $\F_n$ and applying \eqref{eq:mean} inside each term of the sum gives
\begin{align*}
\EE[\ind_E\ind\{\tau_0<\infty\}Z_{\tau_0+1}] &=\sum_{n\geq0}\EE[ \ind_{E\cap\{\tau_0=n\}}Z_{n+1}] \\
&=\sum_{n\geq0}\EE[ \ind_{E\cap\{\tau_0=n\}}V_n(A_{n+1})] \\
&=\EE[\ind_E\ind\{\tau_0<\infty\}V_{\tau_0}(A_{\tau_0+1})].
\end{align*}
For the continuation of \cref{cert:continuation}, apply this identity at the stopping time $\tau_0+t-1$ and use the tower property. Since $\mathcal H_{t-1}\subseteq\F_{\tau_0+t-1}$, $A_{\tau_0+t}=j$, and $p_{t-1}=V_{\tau_0+t-1}(j)$ is $\mathcal H_{t-1}$-measurable, this gives $\EE[Y_t\mid\mathcal H_{t-1}]=p_{t-1}$. The filtration $(\mathcal H_t)$ excludes only the threshold $U$.

\section{Proofs of the upper bounds}\label{cert:appendix}

This appendix proves \cref{thm:single}(a), \cref{thm:search}, and \cref{prop:bounded} by analyzing SPAR conditional on the history at commitment and applying the resulting bounds to Screen-then-SPAR and $[0,1]$-valued outcomes. Throughout the appendices, we abbreviate $v_q=q(1-q)$.

\subsection{Validity and success of the verification rule}\label{cert:guarantee}

In the following two lemmas, we apply the updates and stopping rule of \cref{alg:verify} from time zero to a candidate $j$, with any cap $\kappa>0$, tilting parameter $0<\delta<(1-q)/2$, and horizon $L$.

\begin{lemma}[Validity]\label{lem:validity}
If $\nu\in\mathcal E_q$, then $\PP\{\hat J=j,\ V_\tau(j)<q\}\leq\kappa/q$.
\end{lemma}

\begin{lemma}[Success]\label{lem:power}
Suppose that $L\delta^2/(2v_q)\geq1/\kappa$ and that $V_t(j)\geq q+2\delta$ for $0\leq t\leq L$ almost surely. Then
\begin{align*}
\PP\{\hat J=j,\ V_\tau(j)\geq q\} \geq1-e^{-L\delta^2/(2v_q)}.
\end{align*}
\end{lemma}

Both results follow as the special case $\tau_0=0$ of the conditional bounds proved in \cref{cert:reserve-proof,cert:power-proof}, which also handle verification after screening in \cref{alg:search}. \Cref{thm:single}(a) is then deduced in \cref{cert:power-proof}.

\subsection{Conditional Bernoulli bounds}\label{cert:concentration}

\begin{lemma}\label{lem:bernoulli}
Let $(\mathcal H_i)_{i\geq0}$ be a filtration, let $Y_1,Y_2,\ldots$ be binary variables with $Y_i$ $\mathcal H_i$-measurable, and put $\pi_i=\EE[Y_i\mid\mathcal H_{i-1}]$. Let the integer $m\geq1$ and the levels $a,r\in[0,1]$ be $\mathcal H_0$-measurable, let $D\in\mathcal H_0$, and write $\bar Y=m^{-1}\sum_{i=1}^mY_i$.
\begin{enumerate}[label=(\alph*)]
\item If $\pi_i\leq a$ for $1\leq i\leq m$ almost surely on $D$, then, almost surely on $D\cap\{a<r\}$,
\begin{align}\label{cert:chernoff-upper}
\PP(\bar Y\geq r\mid\mathcal H_0)\leq e^{-m\kl(r,a)}.
\end{align}
\item If $\pi_i\geq a$ for $1\leq i\leq m$ almost surely on $D$, then, almost surely on $D\cap\{r<a\}$,
\begin{align}\label{cert:chernoff-lower}
\PP(\bar Y\leq r\mid\mathcal H_0)\leq e^{-m\kl(r,a)}.
\end{align}
\item If $a,r\in[q,1]$, then
\begin{align}\label{cert:kl-curvature}
\kl(r,a)\geq\frac{(r-a)^2}{2v_q}.
\end{align}
\end{enumerate}
\end{lemma}

\begin{proof}
We first prove (a). If $a=0$, then every $Y_i$ is zero almost surely on $D$, so \eqref{cert:chernoff-upper} is trivial. Otherwise, fix an integer $k\geq1$ and an $\mathcal H_0$-measurable $s>0$, and restrict to $D\cap\{m=k\}$. Because each $Y_i\in\{0,1\}$ has conditional mean $\pi_i\leq a$ and $e^s-1>0$,
\begin{align*}
\EE[e^{sY_i}\mid\mathcal H_{i-1}]=1+\pi_i(e^s-1)\leq1-a+ae^s, \qquad 1\leq i\leq k.
\end{align*}
Iterating conditional expectations from $\mathcal H_{k-1}$ back to $\mathcal H_0$ gives
\begin{align*}
\EE[e^{sm\bar Y}\mid\mathcal H_0] =\EE\Bigl[\prod_{i=1}^ke^{sY_i}\Bigm|\mathcal H_0\Bigr] \leq(1-a+ae^s)^m,
\end{align*}
without assuming that the $Y_i$ are independent. By the conditional Markov inequality,
\begin{align*}
\PP(\bar Y\geq r\mid\mathcal H_0) \leq e^{-smr}\,\EE[e^{sm\bar Y}\mid\mathcal H_0] \leq\exp\bigl\{-m\bigl[sr-\log(1-a+ae^s)\bigr]\bigr\}.
\end{align*}
When $0<a<r<1$, set $e^s=r(1-a)/[a(1-r)]>1$. Then $1-a+ae^s=(1-a)/(1-r)$, so
\begin{align*}
sr-\log(1-a+ae^s)=r\log\frac ra+(1-r)\log\frac{1-r}{1-a}=\kl(r,a),
\end{align*}
which gives \eqref{cert:chernoff-upper}. The endpoint $r=1$ follows by continuity.

Part (b) follows by applying part (a) to the variables $1-Y_i$ with levels $1-a$ and $1-r$, since $1-\pi_i\leq 1-a$, $\{\bar Y\leq r\}=\{1-\bar Y\geq 1-r\}$, and $\kl(1-r,1-a)=\kl(r,a)$.

For (c), fix $a\in[q,1)$ and let $f(u)=\kl(u,a)$ for $u\in[q,1)$. Then $f(a)=f'(a)=0$. Because $q>1/2$, we have $u(1-u)\leq v_q$ on $[q,1)$, so
\begin{align*}
f''(u)=\frac1{u(1-u)}\geq\frac1{v_q},\qquad u\in[q,1).
\end{align*}
Taylor's theorem therefore yields $f(r)\geq(r-a)^2/(2v_q)$ for $r\in[q,1)$, and continuity extends \eqref{cert:kl-curvature} to $a=1$ or $r=1$.
\end{proof}

\subsection{The verification rule after a commitment}\label{cert:continuation}

To treat direct verification (\cref{thm:single}(a), where $\tau_0=0$) and post-screening verification (\cref{alg:search}) simultaneously, suppose the procedure commits to a candidate $j$ after a random game $\tau_0$, where $\tau_0$ and $j$ depend only on the observed history and private randomness $U'$ independent of the verification threshold $U$. Thus $\tau_0$ is a stopping time for the filtration generated by $U'$ and the observations, the information at commitment is
\begin{align*}
\mathcal H_0=\sigma(U',\tau_0,O_{1:\tau_0}),
\end{align*}
and $j$ along with the verification parameters $\kappa$, $\delta$, and $L$ are $\mathcal H_0$-measurable. All conditional statements given $\mathcal H_0$ below are understood to hold almost surely on $\{\tau_0<\infty\}$.

Consider the continuation of the procedure that coincides with it through game $\tau_0$, plays $j$ throughout games $\tau_0+1,\ldots,\tau_0+L$, and never inspects $U$. On this continuation, set
\begin{align*}
Y_t=Z_{\tau_0+t},\qquad g_t=g(Y_t),\qquad a_t=[\mathcal K_{t-1}g_t-\kappa]_+
\end{align*}
for $1\leq t\leq L$, and
\begin{align*}
p_t=V_{\tau_0+t}(j),\qquad \mathcal H_t=\mathcal H_0\vee\sigma(O_{\tau_0+1},\ldots,O_{\tau_0+t})
\end{align*}
for $0\leq t\leq L$, with $\mathcal K_t$ and $S_t=S_{t-1}+a_t$ updated via \eqref{cert:reserve}. Each of these quantities is $\mathcal H_t$-measurable, $U$ is independent of $\mathcal H_L$, and $\EE[Y_t\mid\mathcal H_{t-1}]=p_{t-1}$ by \cref{app:stopping}.

Let $\tau'=\inf\{t\in\{1,\ldots,L\}:S_t\geq U\}$ (with $\inf\varnothing=\infty$). Because the procedure and its continuation play identical actions up to $\tau=\tau_0+\min\{\tau',L\}$, we may couple the experiment directly to this continuation and $U$, under which
\begin{align}\label{cert:events}
\{\hat J=j,\ V_\tau(j)<q\}=\{\tau'\leq L,\ p_{\tau'}<q\}, \qquad \{\hat J=j,\ V_\tau(j)\geq q\}=\{\tau'\leq L,\ p_{\tau'}\geq q\}.
\end{align}

Two elementary properties of $(\mathcal K_t,S_t)$ drive the analysis. By definition of the cap, $\mathcal K_t+a_t=\mathcal K_{t-1}g_t$. Since $g_t=1+\delta(Y_t-q)/v_q$ takes the values $(q+\delta)/q$ and $(1-q-\delta)/(1-q)$ and has conditional mean $1+\delta(p_{t-1}-q)/v_q$ given $\mathcal H_{t-1}$, we obtain
\begin{align}\label{cert:factor-mean}
0<g_t\leq q^{-1},\qquad \EE[\mathcal K_t+a_t\mid\mathcal H_{t-1}] =\mathcal K_{t-1}\left[1+\frac{\delta}{v_q}(p_{t-1}-q)\right] \leq \mathcal K_{t-1}\quad\text{if }p_{t-1}\leq q.
\end{align}
Thus $\mathcal K_t+S_t$ is a supermartingale whenever the value stays at or below $q$. Note that \eqref{cert:factor-mean} and part (a) of \cref{lem:conditional} below only need $0<\delta<1-q$, whereas $\delta<(1-q)/2$ is used in part (b). Next, with $M_t=\prod_{i=1}^t g_i$, we have the lower bound
\begin{align}\label{cert:evidence}
S_t\geq\kappa\log M_t.
\end{align}
To verify \eqref{cert:evidence}, observe that $\mathcal K_{t-1}g_t=\mathcal K_t(1+a_t/\kappa)$ in both cases $\mathcal K_{t-1}g_t\leq\kappa$ (where $a_t=0$) and $\mathcal K_{t-1}g_t>\kappa$ (where $\mathcal K_t=\kappa$). We take logarithms, sum over $i=1,\ldots,t$, and use $\mathcal K_0=\kappa$ to obtain
\begin{align*}
\log\frac{\mathcal K_t}{\kappa}=\log M_t-\sum_{i=1}^t\log\Bigl(1+\frac{a_i}\kappa\Bigr).
\end{align*}
Since $\mathcal K_t\leq\kappa$ and $\log(1+x)\leq x$, this gives $0\geq\log M_t-S_t/\kappa$, which is \eqref{cert:evidence}.

\begin{lemma}\label{lem:conditional}
\begin{enumerate}[label=(\alph*)]
\item If $\nu\in\mathcal E_q$, then, almost surely on $\{\tau_0<\infty\}$,
\begin{align}\label{cert:main-validity}
\PP\{\tau'\leq L,\ p_{\tau'}<q\mid\mathcal H_0\}\leq\kappa/q.
\end{align}
\item Let $D\in\mathcal H_0$ with $D\subseteq\{\tau_0<\infty\}$. If $\delta<(1-q)/2$ and $L\delta^2/(2v_q)\geq1/\kappa$ on $D$ and $p_t\geq q+2\delta$ for $0\leq t\leq L$ almost surely on $D$, then, almost surely on $D$,
\begin{align}\label{cert:main-power}
\PP\{\tau'\leq L,\ p_{\tau'}\geq q\mid\mathcal H_0\} \geq1-e^{-L\delta^2/(2v_q)}.
\end{align}
\end{enumerate}
\end{lemma}

Setting $\tau_0=0$ in \eqref{cert:events} recovers \cref{lem:validity,lem:power}. Moreover, since $\{\tau'\leq L',\ p_{\tau'}<q\}\subseteq\{\tau'\leq L,\ p_{\tau'}<q\}$ for any random $L'\leq L$, part (a) remains valid if verification is terminated early with abstention.

\subsection{Proof of the validity lemma}\label{cert:reserve-proof}

\begin{proof}[Proof of \cref{lem:conditional}(a)]
Let $\theta=\inf\{0\leq t\leq L:p_t<q\}$ (with $\inf\varnothing=\infty$) denote the first verification game at which the value lies below $q$. Apart from the null event $\{U=0\}$, selection occurs at game $t$ precisely when $U$ falls into $\mathcal I_t=(S_{t-1},S_t]\cap[0,1]$, an interval of length at most $a_t=S_t-S_{t-1}$. Since $U\sim\operatorname{Unif}[0,1]$ is independent of $\mathcal H_L$ and $\{p_t<q\}\subseteq\{\theta\leq t\}$, conditioning on $\mathcal H_L$ gives
\begin{align*}
\PP\{\tau'\leq L,\ p_{\tau'}<q\mid\mathcal H_L\} =\sum_{t=1}^L|\mathcal I_t|\ind\{p_t<q\} \leq\sum_{t=1}^La_t\ind\{\theta\leq t\},
\end{align*}
and taking conditional expectations given $\mathcal H_0$ yields
\begin{align}\label{cert:common-prefix}
\PP\{\tau'\leq L,\ p_{\tau'}<q\mid\mathcal H_0\} \leq\EE\left[\sum_{t=1}^La_t\ind\{\theta\leq t\} \middle|\mathcal H_0\right].
\end{align}

Separate the crossing game $t=\theta$ from subsequent games by writing
\begin{align*}
\sum_{t=1}^La_t\ind\{\theta\leq t\} =a_\theta\ind\{1\leq\theta\leq L\}+\sum_{t=1}^La_t\ind\{\theta<t\}.
\end{align*}
Because the continuation plays $j$ in every verification game, \cref{ass:shape} implies $p_{t-1}\leq q$ on the $\mathcal H_{t-1}$-measurable event $\{\theta<t\}$. Applying \eqref{cert:factor-mean} on $\{\theta<t\}$ gives
\begin{align*}
\EE\left[\ind\{\theta<t\}(\mathcal K_t-\mathcal K_{t-1}+a_t) \middle|\mathcal H_{t-1}\right]\leq0.
\end{align*}
Summing over $t=1,\ldots,L$ and taking conditional expectations given $\mathcal H_0$, we use the telescoping identity
\begin{align*}
\sum_{t=1}^L\ind\{\theta<t\}(\mathcal K_t-\mathcal K_{t-1})=(\mathcal K_L-\mathcal K_\theta)\ind\{\theta<L\}
\end{align*}
together with $\mathcal K_L\geq0$ and $\mathcal K_\theta\geq0$ to obtain
\begin{align}\label{cert:after-crossing}
\EE\left[\sum_{t=1}^La_t\ind\{\theta<t\}\middle|\mathcal H_0\right] \leq\EE\bigl[(\mathcal K_\theta-\mathcal K_L)\ind\{\theta<L\}\bigm|\mathcal H_0\bigr] \leq\EE[\mathcal K_\theta\ind\{\theta\leq L\}\mid\mathcal H_0].
\end{align}
At the crossing itself, on $\{1\leq\theta\leq L\}$ the bounds $\mathcal K_{\theta-1}\leq\kappa$ and $g_\theta\leq1/q$ yield
\begin{align*}
a_\theta+\mathcal K_\theta=\mathcal K_{\theta-1}g_\theta\leq\frac\kappa q,
\end{align*}
whereas on $\{\theta=0\}$ we simply have $\mathcal K_\theta=\mathcal K_0=\kappa$. Substituting these estimates and \eqref{cert:after-crossing} into \eqref{cert:common-prefix} gives
\begin{equation}\label{cert:validity-bound}
\PP\{\tau'\leq L,\ p_{\tau'}<q\mid\mathcal H_0\} \leq \frac{\kappa}{q}\PP(1\leq\theta\leq L\mid\mathcal H_0) +\kappa\PP(\theta=0\mid\mathcal H_0) \leq\frac{\kappa}{q}. \qedhere
\end{equation}
\end{proof}

\subsection{Proof of the success lemma and of \texorpdfstring{\cref{thm:single}(a)}{the one-candidate theorem, part (a)}}\label{cert:power-proof}

\begin{proof}[Proof of \cref{lem:conditional}(b)]
Let $\bar Y_L=L^{-1}\sum_{t=1}^LY_t$. Since $g_t$ takes the values $(q+\delta)/q$ and $(1-q-\delta)/(1-q)$ for $Y_t=1$ and $Y_t=0$, respectively,
\begin{align*}
\log M_L =L\left[\bar Y_L\log\frac{q+\delta}{q} +(1-\bar Y_L)\log\frac{1-q-\delta}{1-q}\right].
\end{align*}
The expression on the right is strictly increasing in $\bar Y_L$ and equals $L\kl(q+\delta,q)$ when $\bar Y_L=q+\delta$. Combining this monotonicity with \cref{lem:bernoulli}(c) yields
\begin{align}\label{cert:likelihood-power}
\{\bar Y_L\geq q+\delta\} \subseteq \{\log M_L\geq L\kl(q+\delta,q)\} \subseteq \{\log M_L\geq L\delta^2/(2v_q)\}.
\end{align}
On $D\cap\{\bar Y_L\geq q+\delta\}$, the lower bound \eqref{cert:evidence} and the hypothesis $L\delta^2/(2v_q)\geq1/\kappa$ then force
\begin{align*}
S_L\geq\kappa\log M_L\geq\kappa\frac{L\delta^2}{2v_q}\geq1\geq U,
\end{align*}
so $\tau'\leq L$; moreover, the selection is correct because $p_t\geq q+2\delta>q$ throughout $0\leq t\leq L$ on $D$. Hence
\begin{align*}
D\cap\{\tau'\leq L,\ p_{\tau'}\geq q\}^c\subseteq D\cap\{\bar Y_L<q+\delta\}.
\end{align*}
Since $\EE[Y_t\mid\mathcal H_{t-1}]=p_{t-1}\geq q+2\delta$ on $D$, applying \cref{lem:bernoulli}(b) with $a=q+2\delta$ and $r=q+\delta$ followed by \cref{lem:bernoulli}(c) gives, almost surely on $D$,
\begin{equation}\label{cert:power-bound}
1-\PP\{\tau'\leq L,\ p_{\tau'}\geq q\mid\mathcal H_0\} \leq\PP(\bar Y_L<q+\delta\mid\mathcal H_0) \leq e^{-L\kl(q+\delta,q+2\delta)} \leq e^{-L\delta^2/(2v_q)}. \qedhere
\end{equation}
\end{proof}

\begin{proof}[Proof of \cref{thm:single}(a)]
Apply \cref{lem:finite-cap} with $\tau_0=0$, $\delta=\Delta/4$, and $L=\lceil32B_\Delta/(q\varepsilon)\rceil$ as in \eqref{cert:single-parameters}. It gives $\varepsilon$-validity on $\mathcal E_q$ and $\kappa_L\geq q\varepsilon/(1+\varpi)$.

For success, the choice of $L$ gives $L\delta^2/(2v_q)\geq1/(q\varepsilon)$. Since $32B_\Delta/(q\varepsilon)=32(1-q)/(\varepsilon\Delta^2)>64$, rounding up increases this quantity by a factor less than $65/64$. Thus $L\leq65B_\Delta/\varepsilon$ because $q>1/2$. Under the stated condition $\Delta/\lambda\geq130B_\Delta/\varepsilon$, we have $\lambda L\leq\Delta/2$, so on the continuation that plays strategy $1$ for $L$ games,
\begin{align*}
V_t(1)\geq q+\Delta-\lambda t\geq q+2\delta,\qquad 0\leq t\leq L.
\end{align*}
Set $r=q+5\delta/4$. As a function of $\bar Y_L$, the average log-likelihood ratio $L^{-1}\log M_L$ is affine, with value $\kl(q+\delta,q)\geq\delta^2/(2v_q)$ at $\bar Y_L=q+\delta$ by \cref{lem:bernoulli}(c). Its slope is
\begin{align*}
\log\frac{(q+\delta)(1-q)}{q(1-q-\delta)} =\int_q^{q+\delta}\frac{du}{u(1-u)} \geq\frac{\delta}{v_q}.
\end{align*}
On $\{\bar Y_L\geq r\}$ this gives $\log M_L\geq 3L\delta^2/(4v_q)$. By \eqref{cert:evidence},
\begin{align*}
S_L\geq\kappa_L\log M_L\geq\frac{3}{2(1+\varpi)}>1,
\end{align*}
so SPAR selects by $L$ and the selection is correct. Since $q+2\delta-r=3\delta/4$, \cref{lem:bernoulli}(b)--(c) bounds the probability of failing to make a correct selection by
\begin{align*}
\PP(\bar Y_L<r)\leq\exp\!\left(-\frac{9L\delta^2}{32v_q}\right) \leq\exp\!\left(-\frac{9}{16q\varepsilon}\right)\leq\varepsilon.
\end{align*}
The last inequality follows from $\varepsilon\log(1/\varepsilon)\leq 1/e<9/16$. This proves $(L,\varepsilon)$-success and $N^\star\leq L\leq65B_\Delta/\varepsilon$. The proof applies for all $0<\Delta\leq1-q$ and $0<\varepsilon<1/2$.
\end{proof}

\subsection{Proof of \texorpdfstring{\cref{thm:search}(a)}{Validity of the search procedure} and a stage-wise success condition}\label{search:proof}

Throughout this subsection and the next, set
\begin{align*}
\eta_l=\frac{\varepsilon}{8K(l+1)(l+2)},\qquad x_l=\log(1/\eta_l),\qquad b_l=2\delta_l=\frac14\sqrt{\frac{2v_qx_l}{m_l}},\qquad L_l=\left\lceil\frac{45v_q}{\varepsilon b_l^2}\right\rceil,
\end{align*}
where $m_l=2^{l+1}$ and $\delta_l=\sqrt{v_qx_l/(32m_l)}$ are the stage parameters of \cref{alg:search}, and note that $\sum_{l\geq0}K\eta_l=\varepsilon/8$. The horizon $L_l$ is an analytical device used only in the success proof; the algorithm itself verifies over the entire remaining budget. Let $\tau_0$ denote the commitment time (with $\tau_0=\infty$ if no screen passes), and let $s_0$ and $j_0$ be the stage and candidate at commitment.

\begin{lemma}[Finite-horizon calibration]\label{lem:finite-cap}
In the conditional setup of \cref{cert:continuation}, let $L\geq1$ and
$0<\delta<1-q$ be $\mathcal H_0$-measurable, and take the cap
$\kappa=\kappa_L$ from \eqref{eq:kappa-L}. If $\nu\in\mathcal E_q$, then,
almost surely on $\{\tau_0<\infty\}$,
\begin{align}\label{search:finite-cap-validity}
\PP\{\tau'\leq L,\ p_{\tau'}<q\mid\mathcal H_0\}\leq\varepsilon.
\end{align}
Moreover,
\begin{align}\label{search:finite-cap-comparison}
\kappa_L\geq\frac{q\varepsilon}{1+\varpi}.
\end{align}
\end{lemma}

\begin{proof}
We work conditionally on $\mathcal H_0$, which fixes $\delta$ and $L$, using the continuation of \cref{cert:continuation}.

\emph{Step 1: Comparison with independent Bernoulli$(q)$ outcomes.} For $h\geq0$ and $k\in[0,1]$, let $f_h(k)$ denote the expected excess accumulated over $h$ independent Bernoulli$(q)$ games under unit cap ($\kappa=1$), starting from capital $k$ and zero excess. Thus $f_0(k)=0$ and $f_h(1)=\EE_q S_h^{(1)}$. Define
\begin{align}\label{search:psi-def}
\psi_h(z)=[z-1]_++f_h(\min\{z,1\}),\qquad z\geq0.
\end{align}
Conditioning on the first game gives the recursion
\begin{align*}
f_{h+1}(k)=(1-q)\psi_h(kg(0))+q\psi_h(kg(1)),\qquad k\in[0,1].
\end{align*}
By induction on $h$, $f_h(k)$ is nondecreasing in $k\in[0,1]$: this is trivial for $h=0$, and if $f_h$ is nondecreasing then $\psi_h$ is nondecreasing on $[0,\infty)$, making $f_{h+1}$ nondecreasing as well. Also $f_h(k)$ is nondecreasing in $h$, since each additional game adds nonnegative excess.

Now let $Y_1',\ldots,Y_h'$ be binary variables adapted to a filtration $(\mathcal G_i)_{i=0}^h$ with $\EE[Y_i'\mid\mathcal G_{i-1}]\leq q$, and let $(\tilde{\mathcal K}_i,\tilde S_i)$ follow the unit-cap recursion \eqref{cert:reserve} from a $\mathcal G_0$-measurable starting capital $\tilde{\mathcal K}_0\in[0,1]$ and $\tilde S_0=0$. We show by induction on $h$ that $\EE[\tilde S_h\mid\mathcal G_0]\leq f_h(\tilde{\mathcal K}_0)$. The case $h=0$ is immediate. For $h+1$, conditioning on $\mathcal G_1$ and applying the induction hypothesis to the remaining $h$ games (from starting capital $\tilde{\mathcal K}_1=\min\{1,\tilde{\mathcal K}_0 g(Y_1')\}$) gives
\begin{align*}
\EE[\tilde S_{h+1}\mid\mathcal G_1] \leq [\tilde{\mathcal K}_0 g(Y_1')-1]_+ + f_h(\tilde{\mathcal K}_1) = \psi_h(\tilde{\mathcal K}_0 g(Y_1')).
\end{align*}
Let $p=\EE[Y_1'\mid\mathcal G_0]\leq q$. Since $g(0)<g(1)$ and $\psi_h$ is nondecreasing, we have $\psi_h(\tilde{\mathcal K}_0 g(0))\leq\psi_h(\tilde{\mathcal K}_0 g(1))$, and therefore
\begin{align*}
\EE[\tilde S_{h+1}\mid\mathcal G_0] \leq (1-p)\psi_h(\tilde{\mathcal K}_0 g(0))+p\psi_h(\tilde{\mathcal K}_0 g(1)) \leq (1-q)\psi_h(\tilde{\mathcal K}_0 g(0))+q\psi_h(\tilde{\mathcal K}_0 g(1)) = f_{h+1}(\tilde{\mathcal K}_0).
\end{align*}
Because the update \eqref{cert:reserve} scales linearly with the cap, any run of $h\leq L-1$ games with cap $\kappa_L$, starting capital $\mathcal K\leq\kappa_L$, and conditional win probabilities at most $q$ produces expected excess at most
\begin{align}\label{search:domination}
\kappa_L f_h(\mathcal K/\kappa_L)\leq \kappa_L f_h(1)\leq \kappa_L f_{L-1}(1)=\kappa_L\EE_q S_{L-1}^{(1)}.
\end{align}

\emph{Step 2: Validity bound \eqref{search:finite-cap-validity}.} Set $\theta=\inf\{0\leq t\leq L:p_t<q\}$ (with $\inf\varnothing=\infty$) and $T=\max\{1,\theta\}$. Starting from \eqref{cert:common-prefix} and splitting according to the value of $T$, we have
\begin{align*}
\PP\{\tau'\leq L,\ p_{\tau'}<q\mid\mathcal H_0\} &\leq\EE\left[\sum_{t=1}^L a_t\ind\{\theta\leq t\} \middle|\mathcal H_0\right]\\
&=\EE\left[\sum_{u=1}^L\ind\{T=u\} \left(a_u+\sum_{t=u+1}^L a_t\right) \middle|\mathcal H_0\right].
\end{align*}
Fix $u\in\{1,\ldots,L\}$. The event $\{T=u\}$ belongs to $\mathcal H_u$, and on this event we have $a_u\leq\kappa_L\delta/q$ (since $\mathcal K_{u-1}\leq\kappa_L$ and $g_u\leq 1+\delta/q$), $\mathcal K_u\leq\kappa_L$, and $p_t\leq q$ for all $u\leq t<L$ by \cref{ass:shape} (including when $\theta=0$). Applying \eqref{search:domination} conditionally on $\mathcal H_u$ to the remaining $L-u\leq L-1$ games bounds their expected excess by $\kappa_L\EE_q S^{(1)}_{L-1}$. Therefore,
\begin{align*}
\PP\{\tau'\leq L,\ p_{\tau'}<q\mid\mathcal H_0\} \leq\kappa_L\left(\frac\delta q+\EE_q S^{(1)}_{L-1}\right) \PP(T\leq L\mid\mathcal H_0) \leq\varepsilon,
\end{align*}
which establishes \eqref{search:finite-cap-validity}.

\emph{Step 3: Lower bound \eqref{search:finite-cap-comparison}.} Under independent Bernoulli$(q)$ outcomes, $\EE_qg(Y_t)=1$, so $\mathcal K_t^{(1)}+S_t^{(1)}$ is a martingale starting at $1$. Hence $\EE_q S_{L-1}^{(1)}=1-\EE_q\mathcal K_{L-1}^{(1)}\leq1$. Since $q+\delta<1$ and $q<1$, we obtain
\begin{align*}
\kappa_L =\frac{\varepsilon}{\delta/q+\EE_qS^{(1)}_{L-1}+\varpi} \geq\frac{\varepsilon}{1+\delta/q+\varpi} =\frac{q\varepsilon}{q+\delta+q\varpi} \geq\frac{q\varepsilon}{1+\varpi}. \qedhere
\end{align*}
\end{proof}

\begin{proof}[Proof of \cref{thm:search}(a)]
Before running any screen, \cref{alg:search} verifies that $n+m_l<N$, so every screen concludes strictly before game $N$; once a commitment occurs, verification uses at most the remaining $N-\tau_0$ games, ensuring $\tau\leq N$. For validity, every executed screen has $0<\delta_l<1-q$. No selection is made if no screen passes, and on $\{\tau_0<\infty\}$ the commitment is determined purely by the observed history while verification draws an independent threshold $U$, with $L=N-\tau_0$ and $\kappa_L$ both $\mathcal H_0$-measurable. Combining \eqref{cert:events} with \cref{lem:finite-cap} and averaging over $\mathcal H_0$ gives, for every $\nu\in\mathcal E_q$,
\begin{equation*}
\PP_\nu\{\hat J\neq\varnothing,\ V_\tau(\hat J)<q\} =\EE_\nu\bigl[\ind\{\tau_0<\infty\} \PP_\nu\{\tau'\leq L,\ p_{\tau'}<q\mid\mathcal H_0\}\bigr] \leq\varepsilon. \qedhere
\end{equation*}
\end{proof}

\begin{lemma}[The critical mean of the likelihood ratio]\label{lem:critical-mean}
For $0<\delta<1-q$, define
\begin{align}\label{search:critical-point}
p_{\mathrm c}(\delta)= \frac{\log[(1-q)/(1-q-\delta)]} {\log[(q+\delta)(1-q)/(q(1-q-\delta))]}.
\end{align}
Then $q<p_{\mathrm c}(\delta)<q+\delta$, and for a Bernoulli outcome
$Y$ with mean $p$,
\begin{align*}
\EE_p\log g(Y)= (p-p_{\mathrm c}(\delta)) \log\frac{(q+\delta)(1-q)}{q(1-q-\delta)}.
\end{align*}
If $\delta=b_s/2$ and $m_s\kl(1,q+\delta)\geq x_s$, then
\begin{align}\label{search:critical-screen}
\frac{\delta}{1-q}<\frac16,\qquad q+\delta-p_{\mathrm c}(\delta)\geq\frac{19b_s}{80}.
\end{align}
\end{lemma}

\begin{proof}
Let $w(u)=[u(1-u)]^{-1}$. Since $g(1)=(q+\delta)/q$ and $g(0)=(1-q-\delta)/(1-q)$, direct integration gives
\begin{align*}
\EE_p\log g(Y) =p\log\frac{q+\delta}{q}+(1-p)\log\frac{1-q-\delta}{1-q} =\int_q^{q+\delta}\left(\frac{p}{u}-\frac{1-p}{1-u}\right)du =\int_q^{q+\delta}(p-u)w(u)\,du.
\end{align*}
Hence $p_{\mathrm c}(\delta)=\int_q^{q+\delta} u w(u)\,du\big/\int_q^{q+\delta}w(u)\,du$ is the $w$-weighted mean of $u$ on $[q,q+\delta]$. Both the displayed identity for $\EE_p\log g(Y)$ and the strict inequality $q<p_{\mathrm c}(\delta)<q+\delta$ follow immediately.

Now suppose $\delta=b_s/2$ and $m_s\kl(1,q+\delta)\geq x_s$. Using $b_s=\frac14\sqrt{2v_qx_s/m_s}$ together with $-\log z\leq(1-z)/z$ for $z=q+\delta\in(0,1)$ and $q+\delta>q$, we have
\begin{align*}
\frac{32\delta^2}{v_q}=\frac{x_s}{m_s} \leq-\log(q+\delta)\leq\frac{1-q-\delta}{q+\delta}\leq\frac{1-q-\delta}{q}.
\end{align*}
Setting $r=\delta/(1-q)$, this rearranges to $32r^2\leq1-r$, which implies $r<1/6$. Because $q>1/2$, the weight $w(u)=[u(1-u)]^{-1}$ is increasing on $[q,1)$. Its ratio of maximum to minimum on $[q,q+\delta]$ is therefore
\begin{align*}
\frac{w(q+\delta)}{w(q)}=\frac{q(1-q)}{(q+\delta)(1-q-\delta)}\leq\frac{1-q}{1-q-\delta}=\frac{1}{1-r}<\frac65.
\end{align*}
Change variables via $u=q+\delta t$ ($t\in[0,1]$) and set $f(t)=w(q+\delta t)/w(q)$, so that $1\leq f(t)\leq 6/5$. Then $\int_0^1 f(t)\,dt\geq 1$, and because $(t-1/2)(f(t)-1)\leq 0$ on $[0,1/2]$,
\begin{align*}
\frac{p_{\mathrm c}(\delta)-q}{\delta}-\frac12 =\frac{\int_0^1 t f(t)\,dt}{\int_0^1 f(t)\,dt}-\frac12 =\frac{\int_0^1(t-1/2)(f(t)-1)\,dt}{\int_0^1 f(t)\,dt} \leq\frac15\int_{1/2}^1\Bigl(t-\frac12\Bigr)\,dt=\frac1{40}.
\end{align*}
Thus $p_{\mathrm c}(\delta)\leq q+\frac{21}{40}\delta$, which gives $q+\delta-p_{\mathrm c}(\delta)\geq\frac{19}{40}\delta=\frac{19}{80}b_s$.
\end{proof}

\Cref{thm:search}(b) will follow by checking the hypotheses of the next lemma at a suitable stage $l_\star$.

\begin{lemma}\label{lem:stage}
Under the parameter assumptions of \cref{thm:search}, if a stage $l$
satisfies
\begin{align}\label{search:conditions}
N_l:=2Km_l+L_l\leq N,\qquad \lambda(m_l+L_l)\leq\frac{b_l}{12},\qquad 8.5b_l+\lambda K_{\max}(2m_l-2)\leq\Delta,
\end{align}
then \cref{alg:search} is $(N_l,\varepsilon)$-successful on
$\mathcal E_q(\Delta,\lambda,\mathcal G)$.
\end{lemma}

Roughly speaking, the three inequalities in \eqref{search:conditions} ensure that the budget accommodates all screens through stage $l$ plus $L_l$ verification games, that a candidate's value cannot drop appreciably during its screen and verification prefix, and that a strategy starting at $q+\Delta$ retains a margin of at least $8.5b_l$ when screened at stage $l$.

\begin{proof}
Fix a stage $l$ satisfying \eqref{search:conditions}. Observe first that $b_s$ is strictly decreasing in $s$ and hence $L_s$ is nondecreasing. Indeed, since $K\geq1$ and $\varepsilon\leq1/16$, we have $x_s\geq\log(16/\varepsilon)\geq\log256>\log3$, while
\begin{align*}
x_{s+1}-x_s=\log\frac{s+3}{s+1}\leq\log3<x_s,
\end{align*}
so $b_{s+1}^2/b_s^2=x_{s+1}/(2x_s)<1$. In particular, for every $s\leq l$,
\begin{align}\label{search:earlier-stage}
\lambda(m_s+L_s)\leq\lambda(m_l+L_l) \leq\frac{b_l}{12}\leq\frac{b_s}{12}.
\end{align}

Before any screen at stage $s\leq l$, the number of games already played is at most
\begin{align*}
n\leq K(m_0+\cdots+m_{s-1})+(K-1)m_s=K(2m_s-2)-m_s,
\end{align*}
so $n+m_s+L_s<N_s\leq N_l\leq N$. Hence stage $l$ lies within the loop range ($m_l\leq N$), the remaining verification horizon satisfies $L=N-n-m_s\geq L_s$, and the first $L_s$ verification games after any commitment at stage $s\leq l$ finish by game $N_l$. Moreover, for any admissible tilt $\delta_s=b_s/2<1-q$, we have $b_s<2(1-q)$ and $q>1/2$. Combining $L\geq L_s\geq 45v_q/(\varepsilon b_s^2)$ with \eqref{search:finite-cap-comparison} gives
\begin{align*}
\frac{L\kappa_L\delta_s}{q} \geq\frac{45v_q}{2b_s(1+\varpi)} >\frac{45q}{4(1+\varpi)} >\frac{45}{8(1+\varpi)}>1,
\end{align*}
so every executed screen through stage $l$ enforces the KL passing rule. Writing $p_1=q+\delta_s$, feasibility ($m_s\kl(1,p_1)\geq x_s$) determines a unique threshold $r_s\in(p_1,1]$ with $m_s\kl(r_s,p_1)=x_s$, and the screen passes precisely when its sample mean is at least $r_s$.

We decompose failure into three bad events: a premature commitment to a weak candidate, an unsuccessful verification after a strong commitment, and failure of a high-margin strategy to pass at stage $l$.

\emph{Premature commitments.} For $s\leq l$ and $j\in[K]$, let $\xi_{s,j}$ be the start time of strategy $j$'s screen at stage $s$ (setting $\xi_{s,j}=\infty$ if it is skipped or never reached). Define $E_{s,j}=\varnothing$ if the screen is infeasible, and otherwise let
\begin{align*}
E_{s,j}=\Bigl\{\xi_{s,j}<\infty,\quad \frac1{m_s}\sum_{i=1}^{m_s}Z_{\xi_{s,j}+i}\geq r_s, \quad V_{\xi_{s,j}+m_s}(j) \leq p_{\mathrm c}(\delta_s)+\frac{3b_s}{20}+\lambda L_s\Bigr\}.
\end{align*}
Fix a feasible screen at stage $s\leq l$ and abbreviate $\xi=\xi_{s,j}$, $m=m_s$, $b=b_s$, $p_{\mathrm c}=p_{\mathrm c}(\delta_s)$, $v_i=V_{\xi+i}(j)$, and $X_i=Z_{\xi+i}$. Because every game in the screen uses $j$, \cref{ass:margin}(b) gives $v_{i-1}\leq v_m+\lambda(m-i+1)\leq v_m+\lambda m$. Using \eqref{search:earlier-stage} and \eqref{search:critical-screen} (noting $3/20+1/12=7/30<19/80$), we obtain
\begin{align}\label{search:endpoint-localization}
v_m\leq p_{\mathrm c}+\frac{3b}{20}+\lambda L_s \quad\Longrightarrow\quad v_{i-1}\leq p_{\mathrm c}+\left(\frac3{20}+\frac1{12}\right)b <p_1,\qquad 1\leq i\leq m.
\end{align}
To avoid conditioning on the future endpoint $v_m$, introduce the truncated outcomes
\begin{align*}
\widetilde X_i=X_i\prod_{h=1}^i\ind\{v_{h-1}\leq p_1\}.
\end{align*}
Each $\widetilde X_i$ has conditional mean at most $p_1$ given $\F_{\xi+i-1}$, and $\widetilde X_i=X_i$ for all $i\leq m$ on the endpoint event in \eqref{search:endpoint-localization}. \Cref{lem:bernoulli}(a) therefore gives
\begin{align*}
\PP(E_{s,j}\mid\F_\xi) \leq\PP\left\{\frac1m\sum_{i=1}^m\widetilde X_i \geq r_s\middle|\F_\xi\right\} \leq e^{-m\kl(r_s,p_1)}=e^{-x_s}=\eta_s
\end{align*}
almost surely on $\{\xi<\infty\}$. Summing over $s\leq l$ and $j\in[K]$ yields
\begin{align}\label{search:false-screen-union}
\PP\Bigl(\bigcup_{s=0}^l\bigcup_{j=1}^KE_{s,j}\Bigr) \leq\sum_{s\geq0}K\eta_s=\frac\varepsilon8.
\end{align}

\emph{Verification after a strong commitment.} Passing to the continuation of \cref{cert:continuation} with $p'_t=V_{\tau_0+t}(j_0)$, consider the $\mathcal H_0$-measurable event
\begin{align*}
D=\left\{\tau_0<\infty,\ s_0\leq l,\ p'_0>p_{\mathrm c}(\delta_{s_0})+\frac{3b_{s_0}}{20} +\lambda L_{s_0}\right\}.
\end{align*}
As established above, $L=N-\tau_0\geq L_{s_0}$ on $D$. Because SPAR plays $j_0$ throughout the prefix, \cref{ass:margin}(b) gives
\begin{align*}
p'_t\geq p'_0-\lambda t >p_{\mathrm c}(\delta_{s_0})+\frac{3b_{s_0}}{20}, \qquad 0\leq t\leq L_{s_0}.
\end{align*}

Set $\bar Y=L_{s_0}^{-1}\sum_{t=1}^{L_{s_0}}Y_t$, $b=b_{s_0}$, $\delta=b/2$, and $p_{\mathrm c}=p_{\mathrm c}(\delta)$. By \cref{lem:critical-mean}, the average log-likelihood ratio $L_{s_0}^{-1}\log M_{L_{s_0}}$ is affine in $\bar Y$, vanishes at $\bar Y=p_{\mathrm c}$, and has slope
\begin{align*}
\log\frac{(q+\delta)(1-q)}{q(1-q-\delta)} =\int_q^{q+\delta}\frac{du}{u(1-u)} \geq\frac\delta{v_q}=\frac b{2v_q}.
\end{align*}
Hence $\bar Y\geq p_{\mathrm c}+b/18$ implies $\log M_{L_{s_0}}\geq L_{s_0}b^2/(36v_q)$. Moreover, we have $\EE_q S^{(1)}_{L-1}\leq1$ (from Step~3 of \cref{lem:finite-cap}) and $\delta/q<(1-q)/(6q)<1/6$ (from \eqref{search:critical-screen} and $q>1/2$), so $\kappa_L\geq\varepsilon/(7/6+\varpi)$. By \eqref{cert:evidence} and $L_{s_0}\geq 45v_q/(\varepsilon b^2)$, on $\{\bar Y\geq p_{\mathrm c}+b/18\}$ it follows that
\begin{align*}
S_{L_{s_0}}\geq\kappa_L\log M_{L_{s_0}} \geq\frac{45}{36(7/6+\varpi)}>1,
\end{align*}
which forces a correct selection within the first $L_{s_0}$ verification games whenever $D$ holds.

By \eqref{search:critical-screen}, $p_1-p_{\mathrm c}\geq 19b/80>3b/20$, so both $p_{\mathrm c}+b/18$ and $p_{\mathrm c}+3b/20$ lie in $[q,p_1]$. Applying \cref{lem:bernoulli}(b) and (c) with $L_{s_0}\geq 45v_q/(\varepsilon b^2)$ gives, almost surely on $D$,
\begin{align}\label{search:verification-failure}
1-\PP\{\tau'\leq L_{s_0},\ p'_{\tau'}\geq q\mid\mathcal H_0\} &\leq\PP\{\bar Y<p_{\mathrm c}+b/18\mid\mathcal H_0\}\notag\\
&\leq \exp\left(-\frac{45(3/20-1/18)^2}{2\varepsilon}\right) =\exp\left(-\frac{289}{1440\varepsilon}\right) \leq \exp\left(-\frac{\log(64/3)}{16\varepsilon}\right) \leq\frac{3\varepsilon}{4}.
\end{align}
Here we used $289/1440>0.2006>0.1913>\log(64/3)/16$, and the final inequality holds because $\varepsilon\log(4/(3\varepsilon))$ increases on $(0,1/16]$ to $\log(64/3)/16$. Averaging over $\mathcal H_0$ gives $\PP(D\cap\{\tau'\leq L_{s_0},\ p'_{\tau'}\geq q\}^c)\leq3\varepsilon/4$.

\emph{Detection of a high-margin strategy.} Pick $j^\star$ with $V_0(j^\star)\geq q+\Delta$, write $\xi^\star=\xi_{l,j^\star}$, and define
\begin{align*}
E^\star=\left\{\xi^\star<\infty,\ \frac1{m_l}\sum_{i=1}^{m_l}Z_{\xi^\star+i} <q+4.5b_l\right\}.
\end{align*}
Up to the end of stage $l$, each strategy is screened for at most $m_0+\cdots+m_l=2m_l-2$ games, so at most $K_{\max}(2m_l-2)$ games involve strategies in $G(j^\star)$. By the third condition in \eqref{search:conditions}, on $\{\xi^\star<\infty\}$ we have
\begin{align*}
V_{\xi^\star+i}(j^\star) \geq q+\Delta-\lambda K_{\max}(2m_l-2) \geq q+8.5b_l,\qquad 0\leq i<m_l.
\end{align*}
In particular, $8.5b_l\leq\Delta\leq1-q$, so $\delta_l=b_l/2$ is admissible and $q+4.5b_l\leq1$. By \cref{lem:bernoulli}(c),
\begin{align*}
m_l\kl(q+4.5b_l,q+b_l/2)\geq m_l(4b_l)^2/(2v_q)=x_l,
\end{align*}
which shows both that the screen is feasible and that any average of at least $q+4.5b_l$ passes. Since the conditional means exceed $q+4.5b_l$ by at least $4b_l$, \cref{lem:bernoulli}(b) and (c) yield (with the boundary case $q+8.5b_l=1$ being trivial)
\begin{align*}
\PP(E^\star\mid\F_{\xi^\star}) \leq e^{-m_l(4b_l)^2/(2v_q)}=\eta_l\leq\frac\varepsilon{16}.
\end{align*}

Finally, let $E_+=\{\tau\leq N_l,\ \hat J\ne\varnothing,\ V_\tau(\hat J)\geq q\}$. If no commitment occurs by the end of $j^\star$'s stage-$l$ screen, that screen is reached and fails, so $E^\star$ occurs. Any commitment by stage $l$ with insufficient endpoint value falls into $\bigcup_{s\leq l}\bigcup_{j\leq K}E_{s,j}$, while every other commitment by stage $l$ lies in $D$ and yields $E_+$ whenever verification succeeds within its first $L_{s_0}$ games. Consequently,
\begin{align*}
E_+^c\subseteq \bigcup_{s=0}^l\bigcup_{j=1}^K E_{s,j}\ \cup\ E^\star\ \cup\ \bigl(D\cap\{\tau'\leq L_{s_0},\ p'_{\tau'}\geq q\}^c\bigr),
\end{align*}
and combining the three probability bounds gives
\begin{equation*}
1-\PP(E_+)\leq\frac\varepsilon8+\frac\varepsilon{16} +\frac{3\varepsilon}{4} =\frac{15\varepsilon}{16}<\varepsilon. \qedhere
\end{equation*}
\end{proof}

\subsection{Proof of \texorpdfstring{\cref{thm:search}(b)}{Success of the search procedure}}\label{search:constants}

Recall that $B_\Delta=v_q/\Delta^2$ and $\Lambda=\log(K/\varepsilon)+\log\log B_\Delta$. Assuming \eqref{search:budget}--\eqref{search:rate} with $C=10^6$, we verify the conditions \eqref{search:conditions} of \cref{lem:stage} at the stage
\begin{align*}
l_\star=\min\{l\geq0:b_l\leq\Delta/8.6\}.
\end{align*}

Because $q>1/2$ and $\Delta\leq(1-q)/8$, we have $B_\Delta=v_q/\Delta^2\geq64q/(1-q)>64$. At $l=0$, $m_0=2$ and $x_0\geq\log(16/\varepsilon)\geq\log256>5$, so $b_0^2=v_qx_0/16>4\Delta^2>(\Delta/8.6)^2$ and hence $l_\star\geq1$. Moreover, because $x_l>x_{l-1}$ and $m_l=2m_{l-1}$, we have $b_l^2/b_{l-1}^2=x_l/(2x_{l-1})>1/2$, so $b_l$ decreases to zero with $b_l>b_{l-1}/\sqrt2$. The minimality of $l_\star$ therefore gives
\begin{align*}
\frac{\Delta}{8.6\sqrt2}<b_{l_\star}\leq\frac{\Delta}{8.6}.
\end{align*}
Substituting this lower bound on $b_{l_\star}$ into $m_l=v_qx_l/(8b_l^2)$ and the definition of $L_l$ (using $B_\Delta/\varepsilon>64$ to absorb the ceiling) yields
\begin{align*}
m_{l_\star}<\frac{8.6^2}{4}B_\Delta x_{l_\star}, \qquad L_{l_\star}<1+\frac{90(8.6)^2B_\Delta}{\varepsilon}<\frac{6658B_\Delta}{\varepsilon}.
\end{align*}

To bound $x_{l_\star}$, write $u=\log(K/\varepsilon)>2$, $y=\log B_\Delta>4$, $T=4(u+\log y)$, and $\bar l=\lceil\log_2(200B_\Delta T)\rceil$. Since $\bar l<1+\log_2(200B_\Delta T)$, adding $2$ gives $\bar l+2<3+\log_2(200B_\Delta T)=\log_2(1600B_\Delta T)$. For $u>2$ and $y>4$, we have $\log(1600B_\Delta T)\leq9+u+2y\leq5(\log2)y(1+u)$, and hence
\begin{align*}
\bar l+2\leq\log_2(1600B_\Delta T)\leq5y(1+u).
\end{align*}
It follows that
\begin{align*}
x_{\bar l} =\log8+u+\log[(\bar l+1)(\bar l+2)] \leq\log200+u+2\log y+2\log(1+u) \leq4u+4\log y=T,
\end{align*}
where the last bound holds because $3u+2\log y-2\log(1+u)$ increases in both arguments and already exceeds $\log200$ at $(u,y)=(2,4)$. From $b_{\bar l}^2=v_qx_{\bar l}/(8\cdot2^{\bar l+1})$ and $2^{\bar l}\geq200B_\Delta T$, we obtain
\begin{align*}
b_{\bar l}\leq\frac{\Delta}{\sqrt{3200}}<\frac{\Delta}{8.6},
\end{align*}
which forces $l_\star\leq\bar l$. Consequently, $x_{l_\star}\leq x_{\bar l}\leq4\Lambda$ and $m_{l_\star}<(8.6^2)B_\Delta\Lambda<74B_\Delta\Lambda$.

With $C=10^6$, the hypotheses \eqref{search:budget}--\eqref{search:rate} now directly yield the three inequalities in \eqref{search:conditions} at $l=l_\star$:
\begin{align*}
N_{l_\star} &=2Km_{l_\star}+L_{l_\star} <6658B_\Delta(K\Lambda+1/\varepsilon)\leq N,\\
\lambda(m_{l_\star}+L_{l_\star}) &\leq6658B_\Delta(K_{\max}\Lambda+1/\varepsilon)\lambda \leq\frac{6658\Delta}{C} <\frac{\Delta}{12(8.6)\sqrt2}<\frac{b_{l_\star}}{12},\\
8.5b_{l_\star}+\lambda K_{\max}(2m_{l_\star}-2) &\leq\frac{8.5\Delta}{8.6}+148B_\Delta K_{\max}\Lambda\lambda \leq\left(\frac{8.5}{8.6}+\frac{148}{C}\right)\Delta <\Delta.
\end{align*}
\Cref{lem:stage} therefore implies that \cref{alg:search} is $(N_{l_\star},\varepsilon)$-successful, and hence $(N,\varepsilon)$-successful since $N_{l_\star}\leq N$.

\subsection{Outcomes in \texorpdfstring{$[0,1]$}{[0,1]}}\label{app:bounded}

To prove \cref{prop:bounded}, we check each of the five places where the upper-bound proofs invoke binary outcomes and verify that every step carries over to $Z_n\in[0,1]$:
\begin{enumerate}[label=(\roman*)]
\item \emph{Conditional mean identity \eqref{eq:mean}.} Replacing $\PP_\nu(Z_{n+1}=1\mid\F_n)$ by $\EE_\nu[Z_{n+1}\mid\F_n]$ reproduces \eqref{eq:value-bounded}, and the stopping-time extension in \cref{app:stopping} goes through verbatim.
\item \emph{Bounds \eqref{cert:factor-mean} on $g_t$.} For $Y_t\in[0,1]$ and $0<\delta<1-q$,
\begin{align*}
0<1-\frac{\delta}{1-q}\leq g_t\leq1+\frac\delta q\leq\frac1q,
\end{align*}
and linearity of $g_t$ in $Y_t$ preserves $\EE[g_t\mid\mathcal H_{t-1}]=1+\delta(p_{t-1}-q)/v_q$. Neither $\mathcal K_t+a_t=\mathcal K_{t-1}g_t$ nor \eqref{cert:evidence} uses binarity.
\item \emph{Log-likelihood lower bounds in \eqref{cert:likelihood-power} and \cref{lem:stage}.} Writing $\alpha=\log[(q+\delta)/q]$ and $\beta=\log[(1-q-\delta)/(1-q)]$, concavity of $y\mapsto\log[1+\delta(y-q)/v_q]$ on $[0,1]$ gives $\log g_t\geq Y_t\alpha+(1-Y_t)\beta$. Summing over $t$ yields $\log M_L\geq L[\bar Y_L\alpha+(1-\bar Y_L)\beta]$, which increases in $\bar Y_L$ and equals $L\kl(q+\delta,q)$ at $\bar Y_L=q+\delta$.
\item \emph{Conditional Chernoff bounds (\cref{lem:bernoulli}(a)--(b)).} By convexity of $y\mapsto e^{sy}$, we have $e^{sY_i}\leq1-Y_i+Y_ie^s$ on $[0,1]$, so $\EE[e^{sY_i}\mid\mathcal H_{i-1}]\leq1+\pi_i(e^s-1)$ for all $s\in\mathbb R$, and the remainder of the proof is unchanged \citep{hoeffding1963}. Note also that a $[0,1]$-valued variable with conditional mean $1$ equals $1$ almost surely.
\item \emph{Stochastic domination by Bernoulli$(q)$ outcomes in \cref{lem:finite-cap}.} Recall the functions $f_h(x)$ and $\psi_h(z)$ from Step~1 of \cref{lem:finite-cap}, and let $\mathcal K_h^x$ be the unit-cap capital after $h$ independent Bernoulli$(q)$ games starting from $x\in[0,1]$. By the martingale identity,
\begin{align*}
f_h(x)=x-\EE_q\mathcal K_h^x.
\end{align*}
For any realized sequence of likelihood factors $g_1,\ldots,g_h$, $\mathcal K_h^x$ is the minimum of $x\prod_{i=1}^h g_i$ and terms that do not depend on $x$. Hence $\mathcal K_h^x$ is nondecreasing and concave in $x$, so $f_h$ is convex. Because $f_h$ is also nondecreasing and $\mathcal K_h^x$ is nondecreasing in $x$, the identity above gives $0\leq f_h(y)-f_h(x)\leq y-x$ for $0\leq x\leq y\leq1$. Thus the function $\psi_h(z)=[z-1]_++f_h(\min\{z,1\})$ agrees with $f_h$ on $[0,1]$ (with slopes in $[0,1]$) and continues linearly with slope $1$ on $[1,\infty)$, so $\psi_h$ is nondecreasing and convex on $[0,\infty)$.

Now let $Y\in[0,1]$ satisfy $\EE[Y\mid\mathcal H]=p\leq q$. Since $xg(Y)=(1-Y)xg(0)+Yxg(1)$, convexity and monotonicity of $\psi_h$ give
\begin{align*}
\EE[\psi_h(xg(Y))\mid\mathcal H] &\leq(1-p)\psi_h(xg(0))+p\psi_h(xg(1))\\
&\leq(1-q)\psi_h(xg(0))+q\psi_h(xg(1)) =f_{h+1}(x).
\end{align*}
The backward induction in Step~1 of \cref{lem:finite-cap} therefore carries over to $[0,1]$-valued outcomes, and all remaining steps of \cref{thm:search} are unchanged.
\end{enumerate}
As for the lower bounds in \cref{thm:single}(b) and \cref{thm:lower}, they are established in the binary-outcome game of \cref{lower:route}, which remains a valid instance when outcomes take values in $[0,1]$.

\section{Proofs of the lower bounds}\label{lower:appendix}

This appendix proves \cref{thm:single}(b), \cref{thm:lower}, and \cref{prop:history} using the two-round game below. We derive separate lower bounds for search and verification. Throughout, the failure probability within $N$ games under $\nu$ is $1-\PP_\nu\{\tau\leq N,\hat J\neq\varnothing,V_\tau(\hat J)\geq q\}$, and the failure probability at a finite time is defined analogously with $\{\tau<\infty\}$ in place of $\{\tau\leq N\}$.

\subsection{A two-round game and change of measure}
\label{lower:route}

In each game's first round, the opponent makes a dummy move and the player selects one of $K$ routes; in the second round, the opponent opens or blocks that route and the player makes a forced move. Strategy $j$ chooses route $j$, and the player wins if and only if the chosen route is open. Because the full observation is in bijection with $(A_n,Z_n)$, any history-dependent rule specifying the opening probabilities defines an admissible environment.

Fix a procedure with private seed $U\sim\mu$. For any stopping time $\xi$ bounded by a deterministic integer $\Gamma$, write $\PP_\nu^{(\xi)}$ for the distribution of $(U,\xi,O_1,\ldots,O_\xi)$ under $\nu$.

\begin{lemma}\label{lem:change-of-measure}
Let $\nu$ and $\nu'$ be environments of this game and let $\xi\leq\Gamma$ be a stopping time. Suppose that, for every $n\leq\xi$, the values $V^\nu_{n-1}(A_n)$ and $V^{\nu'}_{n-1}(A_n)$ are either equal or both lie in $(0,1)$.
\begin{enumerate}[label=(\alph*)]
\item The relative entropy of the stopped laws is
\begin{align*}
\KL\bigl(\PP_\nu^{(\xi)},\PP_{\nu'}^{(\xi)}\bigr) =\EE_\nu\sum_{n=1}^{\xi}\kl\bigl(V^\nu_{n-1}(A_n),V^{\nu'}_{n-1}(A_n)\bigr).
\end{align*}
\item If $E\in\F_\xi$ and $\PP_\nu(E)\leq\varepsilon<1-\beta\leq\PP_{\nu'}(E)$ for some $\beta\in[0,1)$, then, with $\log(1/0)=\infty$,
\begin{align*}
\KL\bigl(\PP_\nu^{(\xi)},\PP_{\nu'}^{(\xi)}\bigr) \geq(1-\varepsilon)\log\frac1\beta-\log2.
\end{align*}
\end{enumerate}
\end{lemma}

\begin{proof}
Under either environment, the law of $(U,\xi,O_1,\ldots,O_\xi)$ factorizes into the seed law $\mu$, the procedure's action and stopping rules (which do not depend on the environment), and the conditional outcome probabilities. Since $O_n$ is determined by $(A_n,Z_n)$ and $Z_n$ has conditional probability $f_{V_{n-1}(A_n)}(Z_n)$ given $\F_{n-1}$, where $f_p(z)=p^z(1-p)^{1-z}$, all non-outcome factors cancel in the likelihood ratio:
\begin{align*}
\log\frac{d\PP_\nu^{(\xi)}}{d\PP_{\nu'}^{(\xi)}} =\sum_{n=1}^{\Gamma}\ind\{n\leq\xi\}\,\ell_n,\qquad \ell_n=\log\frac{f_{V^\nu_{n-1}(A_n)}(Z_n)}{f_{V^{\nu'}_{n-1}(A_n)}(Z_n)}.
\end{align*}
By hypothesis, every term $\ell_n$ with $n\leq\xi$ is finite and vanishes whenever the two values agree.

Since $\{n\leq\xi\}$ and $A_n$ are $\F_{n-1}$-measurable and $\EE_\nu[Z_n\mid\F_{n-1}]=V^\nu_{n-1}(A_n)$,
\begin{align*}
\EE_\nu\bigl[\ind\{n\leq\xi\}\,\ell_n\bigm|\F_{n-1}\bigr] =\ind\{n\leq\xi\}\kl\bigl(V^\nu_{n-1}(A_n),V^{\nu'}_{n-1}(A_n)\bigr),
\end{align*}
and summing over $n\leq\Gamma$ after taking expectations yields (a).

For (b), the data-processing inequality applied to $\ind_E$ gives
\begin{align*}
\KL\bigl(\PP_\nu^{(\xi)},\PP_{\nu'}^{(\xi)}\bigr) \geq\kl\bigl(\PP_\nu(E),\PP_{\nu'}(E)\bigr).
\end{align*}
Because $\kl(a,b)$ is decreasing in $a$ and increasing in $b$ on $\{a<b\}$, the assumptions $\PP_\nu(E)\leq\varepsilon<1-\beta\leq\PP_{\nu'}(E)$ imply $\kl(\PP_\nu(E),\PP_{\nu'}(E))\geq\kl(\varepsilon,1-\beta)$. Expanding
\begin{align*}
\kl(\varepsilon,1-\beta) =\varepsilon\log\varepsilon+(1-\varepsilon)\log(1-\varepsilon) +(1-\varepsilon)\log\frac1\beta+\varepsilon\log\frac1{1-\beta}
\end{align*}
and noting that the binary entropy term is at least $-\log2$ while $\varepsilon\log(1/(1-\beta))\geq0$ completes the proof.
\end{proof}

\subsection{The cost of the search}
\label{lower:search-information}

\begin{lemma}\label{lem:search-lower}
Let $0<\Delta\leq(1-q)/8$ and $0<\varepsilon\leq1/2$, and fix a procedure that is $\varepsilon$-valid on $\mathcal E_q$.
\begin{enumerate}[label=(\alph*)]
\item The supremum over $\mathcal E_q(\Delta,\lambda,\mathcal G)$ of the failure probability within $N$ games is at least
\begin{align}\label{lower:search-budget-exact}
2^{-1/(1-\varepsilon)} \exp\left[-\frac{N\kl(q,q+\Delta)}{K(1-\varepsilon)}\right].
\end{align}
\item If $\lambda>0$ and $h=\lfloor\Delta/\lambda\rfloor$, the supremum over $\mathcal E_q(\Delta,\lambda,\mathcal G)$ of the failure probability at a finite time is at least
\begin{align}\label{lower:search-rate-exact}
2^{-1/(1-\varepsilon)} \exp\left[-\frac{1}{K_{\max}(1-\varepsilon)} \sum_{i=0}^{h-1}\kl(q-\lambda i,q+\Delta-\lambda i)\right].
\end{align}
\end{enumerate}
\end{lemma}

\begin{proof}[Proof of (a)]
Fix $0<\zeta<q$ and consider stationary environments with constant values
\begin{align*}
V_n^{(0)}(i)=q-\zeta,\qquad V_n^{(j)}(i)=q-\zeta+(\Delta+\zeta)\ind\{i=j\},\qquad i,j\in[K],
\end{align*}
with induced laws $\PP_0$ and $\PP_j$. All $K+1$ environments belong to $\mathcal E_q$. Under $\PP_0$, every strategy lies strictly below $q$ so any selection is incorrect, whereas each alternative $\PP_j$ belongs to $\mathcal E_q(\Delta,\lambda,\mathcal G)$ for every $\lambda\geq0$ and every partition $\mathcal G$.

Set
\begin{align*}
\beta=\max_{j\in[K]} \bigl[1-\PP_j\{\tau\leq N,\ \hat J\neq\varnothing,\ V_\tau(\hat J)\geq q\}\bigr].
\end{align*}
Suppose first that $\beta<1-\varepsilon$. Letting $\xi=\tau\wedge N$ and $E=\{\tau\leq N,\hat J\neq\varnothing\}\in\F_\xi$, validity under $\PP_0$ and the definition of $\beta$ give $\PP_0(E)\leq\varepsilon$ and $\PP_j(E)\geq1-\beta$ for each $j\in[K]$. Since $\PP_0$ and $\PP_j$ differ only when strategy $j$ is played, writing $N_j$ for the number of plays of $j$ among the first $\xi$ games and applying \cref{lem:change-of-measure}(a)--(b) yields
\begin{align*}
(1-\varepsilon)\log\frac1\beta-\log2 \leq\KL\bigl(\PP_0^{(\xi)},\PP_j^{(\xi)}\bigr) =\kl(q-\zeta,q+\Delta)\,\EE_0[N_j].
\end{align*}
In particular, $\beta>0$. Summing over $j\in[K]$ and using $\sum_{j=1}^K N_j=\xi\leq N$, we obtain
\begin{align*}
K\left[(1-\varepsilon)\log\frac1\beta-\log2\right] \leq N\kl(q-\zeta,q+\Delta),
\end{align*}
which rearranges to
\begin{align*}
\beta\geq2^{-1/(1-\varepsilon)} \exp\left[-\frac{N\kl(q-\zeta,q+\Delta)}{K(1-\varepsilon)}\right].
\end{align*}
Because the right-hand side is at most $1/2\leq1-\varepsilon$, the same bound holds trivially when $\beta\geq1-\varepsilon$. Letting $\zeta\downarrow0$ gives \eqref{lower:search-budget-exact}.
\end{proof}

\begin{proof}[Proof of (b)]
Fix $0<\zeta<q-\Delta$ and a largest group $G\in\mathcal G$ with $|G|=K_{\max}$, and let $N_G(n)=\sum_{k=1}^n\ind\{A_k\in G\}$ count the plays in $G$ among the first $n$ games. Define the reference environment $\PP_0$ and alternatives $\PP_j$ ($j\in G$) by setting all values outside $G$ to zero and, for $i\in G$,
\begin{align*}
V_n^{(0)}(i)=[q-\zeta-\lambda N_G(n)]_+,\qquad V_n^{(j)}(i)=[q-\zeta+(\Delta+\zeta)\ind\{i=j\}-\lambda N_G(n)]_+.
\end{align*}
Every value is nonincreasing, so all these environments lie in $\mathcal E_q$. Every selection under $\PP_0$ is incorrect, and each $\PP_j$ satisfies \eqref{eq:envelope} and thus belongs to $\mathcal E_q(\Delta,\lambda,\mathcal G)$.

Let $T_{h+1}=\min\{n\geq1:N_G(n)=h+1\}$ (with $\min\varnothing=\infty$) be the time of the $(h+1)$st play in $G$. Since $h+1>\Delta/\lambda$, for every $n\geq T_{h+1}$ we have
\begin{align*}
\max_{i\in[K]}V_n^{(j)}(i)\leq[q+\Delta-\lambda(h+1)]_+<q,
\end{align*}
so a correct selection under $\PP_j$ can occur only if $\tau<T_{h+1}$. Because $A_{n+1}$ is $\F_n$-measurable, $\{T_{h+1}-1\leq n\}=\{N_G(n+1)\geq h+1\}\in\F_n$, making $T_{h+1}-1$ a stopping time.

For any finite integer $\Gamma$, set
\begin{align*}
\xi_\Gamma=\min\{\tau,\Gamma,T_{h+1}-1\},\qquad E_\Gamma=\{\tau\leq\Gamma,\ \tau<T_{h+1},\ \hat J\neq\varnothing\}.
\end{align*}
Then $\xi_\Gamma\leq\Gamma$ is a stopping time and $E_\Gamma\in\F_{\xi_\Gamma}$ (since $\tau=\xi_\Gamma$ on $E_\Gamma$). Writing
\begin{align*}
\beta=\max_{j\in G} \bigl[1-\PP_j\{\tau<\infty,\ \hat J\neq\varnothing,\ V_\tau(\hat J)\geq q\}\bigr]
\end{align*}
and $\beta_\Gamma=\max_{j\in G}\PP_j\{\Gamma<\tau<\infty,\ \hat J\neq\varnothing,\ V_\tau(\hat J)\geq q\}$ (which vanishes as $\Gamma\to\infty$), validity under $\PP_0$ and the constraint $\tau<T_{h+1}$ for correct selections under $\PP_j$ give
\begin{align*}
\PP_0(E_\Gamma)\leq\varepsilon,\qquad \PP_j(E_\Gamma)\geq1-\beta-\beta_\Gamma,\qquad j\in G.
\end{align*}

Once again, $\PP_0$ and $\PP_j$ differ only on plays of $j$. For any game $n\leq\xi_\Gamma$ with $A_n=j$, let $i=N_G(n-1)$. Since $n\leq\xi_\Gamma<T_{h+1}$ and $A_n=j\in G$, we have $N_G(n)=i+1\leq h$ and hence $i\leq h-1$. The corresponding outcome probabilities are
\begin{align*}
V_{n-1}^{(0)}(j)=q-\zeta-\lambda i,\qquad V_{n-1}^{(j)}(j)=q+\Delta-\lambda i,
\end{align*}
both of which lie in $(0,1)$ because $\lambda i<\Delta<q-\zeta$ and $q+\Delta<1$. Each value $i\in\{0,\ldots,h-1\}$ can occur for at most one play in $G$, so summing \cref{lem:change-of-measure}(a) over $j\in G$ gives
\begin{align*}
\sum_{j\in G}\KL\bigl(\PP_0^{(\xi_\Gamma)},\PP_j^{(\xi_\Gamma)}\bigr) \leq\sum_{i=0}^{h-1}\kl(q-\zeta-\lambda i,q+\Delta-\lambda i).
\end{align*}
When $\beta+\beta_\Gamma<1-\varepsilon$, applying \cref{lem:change-of-measure}(b) to each term on the left yields
\begin{align*}
K_{\max}\left[(1-\varepsilon)\log\frac1{\beta+\beta_\Gamma}-\log2\right] \leq\sum_{i=0}^{h-1}\kl(q-\zeta-\lambda i,q+\Delta-\lambda i),
\end{align*}
and rearranging as in part (a) lower-bounds $\beta+\beta_\Gamma$ by \eqref{lower:search-rate-exact} with $q-\zeta-\lambda i$ in place of $q-\lambda i$ (a bound that also holds trivially if $\beta+\beta_\Gamma\geq1-\varepsilon$). Letting $\Gamma\to\infty$ and then $\zeta\downarrow0$ completes the proof.
\end{proof}

Since the failure probability within $N$ games dominates the failure probability at a finite time, \eqref{lower:search-rate-exact} also lower-bounds the failure probability within $N$ games.

\subsection{The cost of verification}
\label{lower:certification}

Fix a strategy $j^\star$ with group $G\in\mathcal G$, and define
\begin{align*}
I_N=\frac{N\Delta^2}{(q+\Delta)(1-q-\Delta)} \quad\text{and, if }\lambda>0,\quad I_h=\frac{h\Delta^2}{(q+\Delta)(1-q-\Delta)},\quad h=\lfloor\Delta/\lambda\rfloor.
\end{align*}

\begin{lemma}\label{lem:verification-lower}
Let $0<\Delta\leq(1-q)/8$ and $0<\varepsilon<1/3$, and fix a procedure that is $\varepsilon$-valid on $\mathcal E_q$. The supremum over $\mathcal E_q(\Delta,\lambda,\mathcal G)$ of the failure probability within $N$ games is at least $1-2\varepsilon-\sqrt{8\varepsilon I_N}$. If $\lambda>0$, the supremum of the failure probability at a finite time is at least $1-2\varepsilon-\sqrt{8\varepsilon I_h}$. Both bounds hold even if the procedure knows $j^\star$, $\Delta$, and $\lambda$.
\end{lemma}

The proof compares a target environment (where $j^\star$ starts at $q+\Delta$) with a reference environment (where $j^\star$ has constant value $q$). Because an adversary in $\mathcal E_q$ can drop the value of $j^\star$ below $q$ at any $(\mathcal B_n)$-stopping time, validity forces the remaining selection probability under the reference law to have expectation at most $\varepsilon$ at every such time. The Snell envelope turns this stopping-time bound into a pathwise upper bound $\omega$ outside an event of probability at most $\varepsilon/\omega$. A quadratic potential then bounds the remaining selection probability in terms of the chi-square information between $q+\Delta$ and $q$.

\begin{proof}
\emph{Step 1: Environments and public projections.} In every environment constructed below, all strategies other than $j^\star$ have value zero (and hence deterministic outcome zero). In the budget case, set $\Gamma=N$ and define
\begin{align*}
V_n^{(1)}(j^\star)=q+\Delta,\qquad D_n=\{\tau=n,\ \hat J\neq\varnothing\}.
\end{align*}
In the rate case, fix a finite horizon $\Gamma$, write $N_G(n)=\sum_{k=1}^n\ind\{A_k\in G\}$, and define
\begin{align*}
V_n^{(1)}(j^\star)=[q+\Delta-\lambda N_G(n)]_+,\qquad D_n=\{\tau=n,\ \hat J\neq\varnothing,\ N_G(n)\leq h\}.
\end{align*}
In both cases, this environment lies in $\mathcal E_q(\Delta,\lambda,\mathcal G)$. Let $\PP_1$ denote its law and put $s_\Gamma=\PP_1(\bigcup_{n=0}^\Gamma D_n)$. Every correct selection by time $\Gamma$ under $\PP_1$ lies in $\bigcup_{n\leq\Gamma}D_n$ (in the rate case because $V_n^{(1)}(j^\star)\leq[q+\Delta-\lambda(h+1)]_+<q$ whenever $N_G(n)>h$). Let $\PP_0$ be the law of the stationary reference environment with $V_n^{(0)}(j^\star)=q$. Since $q\in(0,1)$, every public history that has positive probability under $\PP_1$ also has positive probability under $\PP_0$.

For any public history $o_{1:n}$, let $\mathcal U(o_{1:n})$ be the set of private seeds $u$ that produce the sequence of actions recorded in $o_{1:n}$ (using the convention $A_{k+1}=1$ after stopping). Given $U=u$, the probability of observing $o_{1:n}$ equals $\ind_{\mathcal U(o_{1:n})}(u)$ times a product of outcome probabilities determined solely by $o_{1:n}$ and the environment. Because that product does not depend on $u$, it cancels in Bayes' rule, giving
\begin{align}\label{lower:seed-projection}
\mathcal L(U\mid O_{1:n}=o_{1:n}) =\frac{\ind_{\mathcal U(o_{1:n})}(u)\,\mu(du)} {\mu\{\mathcal U(o_{1:n})\}}
\end{align}
under every environment in this proof. Since $D_n$ and $\{A_{n+1}=j^\star\}$ depend only on $(U,O_{1:n})$, the public projections
\begin{align*}
r_n=\PP(D_n\mid\mathcal B_n),\qquad \pi_n=\PP(A_{n+1}=j^\star\mid\mathcal B_n)
\end{align*}
are the same functions of $O_{1:n}$ under all environments in this proof.

Define the remaining selection probability under the reference law by
\begin{align*}
\phi_n=\PP_0\Bigl(\bigcup_{k=n}^{\Gamma}D_k\Bigm|\mathcal B_n\Bigr) =\EE_0\Bigl[\sum_{k=n}^{\Gamma}r_k\Bigm|\mathcal B_n\Bigr].
\end{align*}
Because the events $D_k$ are disjoint, the tower property gives
\begin{align}\label{lower:reference-recursion}
0\leq\phi_n\leq1,\qquad \phi_\Gamma=r_\Gamma,\qquad \phi_n=r_n+\EE_0[\phi_{n+1}\mid\mathcal B_n].
\end{align}
In the rate case, $\phi_n=0$ on $\{N_G(n)>h\}$ because $N_G$ is nondecreasing.

\emph{Step 2: Stopping-time validity and the Snell envelope.} We claim that
\begin{align}\label{lower:all-public-times}
\EE_1[\phi_T]\leq\varepsilon \qquad\text{for every $(\mathcal B_n)$-stopping time }T\leq\Gamma.
\end{align}
Fix such a $T$ and $0\leq\zeta<q$. Consider the environment with law $\PP_T^\zeta$ that agrees with $\PP_1$ before $T$ and then freezes the value of $j^\star$ at
\begin{align}\label{lower:drop-value}
V_n^{(T,\zeta)}(j^\star)= \begin{cases} V_n^{(1)}(j^\star), & n<T, \\
\min\{V_T^{(1)}(j^\star),q-\zeta\}, & n\geq T, \end{cases}
\end{align}
while keeping all other values at zero. Since $T$ is a $(\mathcal B_n)$-stopping time and $V_n^{(T,\zeta)}(j^\star)$ is nonincreasing, this environment lies in $\mathcal E_q$. Let $D_{\geq T}=\bigcup_{n=T}^{\Gamma}D_n$. We link $\EE_1[\phi_T]$ to $\varepsilon$ in three steps:
\begin{enumerate}[label=(\roman*)]
\item \emph{Validity for $\zeta>0$.} Every selection in $D_{\geq T}$ occurs at a value at most $q-\zeta<q$, so $\varepsilon$-validity gives $\PP_T^\zeta(D_{\geq T})\leq\varepsilon$.
\item \emph{Identification at $\zeta=0$.} The history $O_{1:T}$ has the same distribution under $\PP_T^0$ as under $\PP_1$. On $\{N_G(T)\leq h\}$ (which is the whole space in the budget case), we have $V_T^{(1)}(j^\star)\geq q$, so the value of $j^\star$ after $T$ equals $q$, and \eqref{lower:seed-projection} implies $\PP_T^0(D_{\geq T}\mid\mathcal B_T)=\phi_T$. On $\{N_G(T)>h\}$, both sides are zero. Taking expectations gives $\PP_T^0(D_{\geq T})=\EE_1[\phi_T]$.
\item \emph{Coupling $\PP_T^0$ and $\PP_T^\zeta$.} Generate both experiments using the same private seed $U$ and the same sequence of independent $\operatorname{Unif}[0,1]$ draws to simulate the outcomes of $j^\star$. The two trajectories coincide through game $T$, and in each of the at most $\Gamma$ subsequent games their win probabilities differ by at most $\zeta$. By a union bound over those games, the two trajectories differ with probability at most $\Gamma\zeta$, so $\PP_T^0(D_{\geq T})\leq\PP_T^\zeta(D_{\geq T})+\Gamma\zeta$.
\end{enumerate}
Combining (i)--(iii) gives $\EE_1[\phi_T]\leq\varepsilon+\Gamma\zeta$, and letting $\zeta\downarrow0$ proves \eqref{lower:all-public-times}.

Next, define the Snell envelope of $(\phi_n)$ under $\PP_1$ by
\begin{align*}
W_n=\max_{n\leq T\leq\Gamma}\EE_1[\phi_T\mid\mathcal B_n],
\end{align*}
where the maximum is over $(\mathcal B_k)$-stopping times $T$. By finite-horizon dynamic programming, $W_\Gamma=\phi_\Gamma$ and $W_n=\max\{\phi_n,\EE_1[W_{n+1}\mid\mathcal B_n]\}$, so $(W_n)$ is a $\PP_1$-supermartingale with $\phi_n\leq W_n\leq1$. The optimal stopping time $T^\star=\min\{n:W_n=\phi_n\}$ attains $W_0=\EE_1[\phi_{T^\star}]$, so \eqref{lower:all-public-times} gives $W_0\leq\varepsilon$. For any $\omega>0$, let $T_\omega=\min\{n\leq\Gamma:W_n>\omega\}$ (with $\min\varnothing=\infty$). Because $(W_n)$ is a nonnegative $\PP_1$-supermartingale, optional sampling at the bounded stopping time $T_\omega\wedge\Gamma$ yields
\begin{align}\label{lower:envelope-crossing}
\omega\,\PP_1(T_\omega\leq\Gamma) \leq\EE_1[W_{T_\omega}\ind\{T_\omega\leq\Gamma\}] \leq\EE_1[W_{T_\omega\wedge\Gamma}] \leq W_0\leq\varepsilon, \qquad\text{so}\qquad \PP_1(T_\omega\leq\Gamma)\leq\frac\varepsilon\omega.
\end{align}

\emph{Step 3: Truncation and quadratic potential.} Truncate the selection probabilities at $T_\omega$ by setting
\begin{align*}
\widetilde r_n=r_n\ind\{n<T_\omega\},\qquad \widetilde\phi_n=\EE_0\Bigl[\sum_{k=n}^{\Gamma}\widetilde r_k\Bigm|\mathcal B_n\Bigr].
\end{align*}
Because $T_\omega$ is a $(\mathcal B_n)$-stopping time, $\widetilde r_n$ is $\mathcal B_n$-measurable. On $\{n<T_\omega\}$ we have $\widetilde\phi_n\leq\phi_n\leq W_n\leq\omega$, and on $\{n\geq T_\omega\}$ we have $\widetilde\phi_n=0$. Also $\widetilde\phi_0\leq W_0\leq\varepsilon$. As in \eqref{lower:reference-recursion},
\begin{align}\label{lower:truncated-bounds}
0\leq\widetilde\phi_n\leq\omega,\qquad \widetilde\phi_0\leq\varepsilon,\qquad \widetilde\phi_n=\widetilde r_n+\EE_0[\widetilde\phi_{n+1}\mid\mathcal B_n].
\end{align}
By \eqref{lower:seed-projection}, $r_n=\PP_1(D_n\mid\mathcal B_n)$. Disjointness of the $D_n$ gives $\EE_1\sum_{n=0}^{\Gamma}r_n=s_\Gamma$, and \eqref{lower:envelope-crossing} gives
\begin{align*}
\EE_1\sum_{n=0}^{\Gamma}r_n\ind\{T_\omega\leq n\} =\sum_{n=0}^{\Gamma}\PP_1(D_n\cap\{T_\omega\leq n\}) \leq\PP_1(T_\omega\leq\Gamma)\leq\frac\varepsilon\omega.
\end{align*}
Hence truncation loses at most $\varepsilon/\omega$ of selection mass:
\begin{align}\label{lower:retained-mass}
\EE_1\sum_{n=0}^{\Gamma}\widetilde r_n\geq s_\Gamma-\frac\varepsilon\omega.
\end{align}

To bound the remaining mass, define the concave quadratic potential
\begin{align*}
F_\omega(x)=2x-\frac{x^2}{2\omega},\qquad 0\leq x\leq\omega,
\end{align*}
which satisfies $1\leq F_\omega'(x)\leq2$ and $x\leq F_\omega(x)\leq2x$. Fix $n<\Gamma$ and a public history up to time $n$, and abbreviate
\begin{align*}
X=\widetilde\phi_{n+1},\qquad p_n=V_n^{(1)}(j^\star),\qquad \Delta_n=p_n-q,\qquad d=X(j^\star,1)-X(j^\star,0),
\end{align*}
where $X(j^\star,z)$ is the value of $\widetilde\phi_{n+1}$ after playing $j^\star$ with outcome $z$. By \eqref{lower:seed-projection}, the conditional distribution of $A_{n+1}$ given $\mathcal B_n$ is identical under $\PP_1$ and $\PP_0$. Given $A_{n+1}$, the outcome distributions differ only when $A_{n+1}=j^\star$, where the win probability is $p_n$ under $\PP_1$ and $q$ under $\PP_0$. Therefore $\EE_1[X\mid\mathcal B_n]-\EE_0[X\mid\mathcal B_n]=\pi_n\Delta_nd$, and combining this with \eqref{lower:truncated-bounds} gives
\begin{align*}
\EE_1[X\mid\mathcal B_n]-\widetilde\phi_n=-\widetilde r_n+\pi_n\Delta_nd.
\end{align*}
For the conditional variance, conditioning on $A_{n+1}$ (and noting that when $A_{n+1}=j^\star$, $X$ takes the values $X(j^\star,1)$ and $X(j^\star,0)$ with probabilities $p_n$ and $1-p_n$) gives
\begin{align*}
\Var_1(X\mid\mathcal B_n) \geq\EE_1\bigl[\Var_1(X\mid\mathcal B_n,A_{n+1})\bigm|\mathcal B_n\bigr] =\pi_np_n(1-p_n)d^2.
\end{align*}

Since $F_\omega$ is quadratic with constant second derivative $-1/\omega$, exact Taylor expansion yields
\begin{align*}
\EE_1[F_\omega(X)\mid\mathcal B_n]-F_\omega(\widetilde\phi_n) =F_\omega'(\widetilde\phi_n)\bigl(\EE_1[X\mid\mathcal B_n]-\widetilde\phi_n\bigr) -\frac{\EE_1[(X-\widetilde\phi_n)^2\mid\mathcal B_n]}{2\omega}.
\end{align*}
Using $1\leq F_\omega'\leq2$, $\widetilde r_n\geq0$, and $\EE_1[(X-\widetilde\phi_n)^2\mid\mathcal B_n]\geq\Var_1(X\mid\mathcal B_n)$, we obtain
\begin{align*}
\EE_1[F_\omega(X)\mid\mathcal B_n]-F_\omega(\widetilde\phi_n) \leq -\widetilde r_n+2\pi_n|\Delta_nd|-\frac{\pi_np_n(1-p_n)}{2\omega}\,d^2.
\end{align*}
Applying the elementary bound $2au-bu^2/2\leq 2a^2/b$ (for $a,b>0$ and $u=|d|$) gives
\begin{align}\label{lower:quadratic-step}
\EE_1[F_\omega(X)\mid\mathcal B_n]-F_\omega(\widetilde\phi_n) \leq-\widetilde r_n+2\omega\pi_n\frac{\Delta_n^2}{p_n(1-p_n)}.
\end{align}
In the rate case, whenever $N_G(n)\geq h$ we may replace the last term in \eqref{lower:quadratic-step} by zero (avoiding any $0/0$ issue if $p_n(1-p_n)=0$). Indeed, if $N_G(n)=h$, playing $j^\star$ pushes $N_G$ above $h$, so $X(j^\star,0)=X(j^\star,1)=0$ and $d=0$. If $N_G(n)>h$, then all $\widetilde r_k$ and $\widetilde\phi_k$ for $k\geq n$ are identically zero.

Let $\iota_n=1$ in the budget case and $\iota_n=\ind\{N_G(n)<h\}$ in the rate case, and suppose
\begin{align*}
I\geq\EE_1\sum_{n=0}^{\Gamma-1}\iota_n\pi_n\frac{\Delta_n^2}{p_n(1-p_n)}.
\end{align*}
Taking $\PP_1$-expectations in \eqref{lower:quadratic-step} and summing over $n=0,\ldots,\Gamma-1$ telescopes to
\begin{align*}
\EE_1\sum_{n=0}^{\Gamma-1}\widetilde r_n \leq F_\omega(\widetilde\phi_0)-\EE_1F_\omega(\widetilde\phi_\Gamma)+2\omega I.
\end{align*}
Since $F_\omega(\widetilde\phi_0)\leq2\widetilde\phi_0\leq2\varepsilon$ and $\widetilde r_\Gamma=\widetilde\phi_\Gamma\leq F_\omega(\widetilde\phi_\Gamma)$, adding $\EE_1\widetilde r_\Gamma$ to both sides yields $\EE_1\sum_{n=0}^{\Gamma}\widetilde r_n\leq2\varepsilon+2\omega I$. Combining this with \eqref{lower:retained-mass} gives
\begin{align*}
s_\Gamma-\frac\varepsilon\omega\leq\EE_1\sum_{n=0}^{\Gamma}\widetilde r_n\leq 2\varepsilon+2\omega I, \qquad\text{so}\qquad s_\Gamma\leq 2\varepsilon+\frac\varepsilon\omega+2\omega I
\end{align*}
for every $\omega>0$. Choosing $\omega=\sqrt{\varepsilon/(2I)}$ (or letting $\omega\to\infty$ when $I=0$) yields
\begin{align}\label{lower:success-information}
s_\Gamma\leq2\varepsilon+\sqrt{8\varepsilon I}.
\end{align}

\emph{Step 4: Bounding the information.} In the budget case, $p_n=q+\Delta$ and $\iota_n=1$. Since $\pi_n=\PP_1(A_{n+1}=j^\star\mid\mathcal B_n)$ and at most $N$ games are played,
\begin{align*}
\EE_1\sum_{n=0}^{N-1}\pi_n\frac{\Delta^2}{(q+\Delta)(1-q-\Delta)} =\frac{\Delta^2}{(q+\Delta)(1-q-\Delta)}\,\EE_1\sum_{n=0}^{N-1}\ind\{A_{n+1}=j^\star\} \leq I_N.
\end{align*}
Applying \eqref{lower:success-information} with $I=I_N$ gives the budget bound of the lemma.

In the rate case, on $\{N_G(n)<h\}$ we have $\Delta_n=\Delta-\lambda N_G(n)$ and $p_n=q+\Delta_n\in[q,q+\Delta]$. Since $\pi_n=\EE_1[\ind\{A_{n+1}=j^\star\}\mid\mathcal B_n]$ and each play of $j^\star\in G$ increases $N_G$ by one, every integer $i\in\{0,\ldots,h-1\}$ can occur at most once with $A_{n+1}=j^\star$. Therefore,
\begin{align*}
\EE_1\sum_{n=0}^{\Gamma-1}\iota_n\pi_n\frac{\Delta_n^2}{p_n(1-p_n)} =\EE_1\sum_{n=0}^{\Gamma-1}\iota_n\ind\{A_{n+1}=j^\star\}\frac{\Delta_n^2}{p_n(1-p_n)} \leq\sum_{i=0}^{h-1} \frac{(\Delta-\lambda i)^2}{(q+\Delta-\lambda i)(1-q-\Delta+\lambda i)} \leq I_h,
\end{align*}
where the last inequality uses $(\Delta-\lambda i)^2\leq\Delta^2$ and $p(1-p)\geq(q+\Delta)(1-q-\Delta)$ on $[q,q+\Delta]$ (because $q>1/2$). Hence \eqref{lower:success-information} holds with $I=I_h$ for every finite $\Gamma$. Letting $\Gamma\to\infty$ bounds the probability of a correct selection at a finite time by $2\varepsilon+\sqrt{8\varepsilon I_h}$, which completes the proof.
\end{proof}

\subsection{Proofs of \texorpdfstring{\cref{thm:lower} and \cref{thm:single}(b)}{the lower bounds}}
\label{lower:combination}

We first record an explicit finite-budget minimax risk bound from which \cref{thm:lower} follows immediately.

\begin{proposition}\label{prop:lower-risk}
In the two-round game of \cref{lower:route}, let $0<\Delta\leq(1-q)/8$, $0<\varepsilon\leq1/16$, and $\lambda\geq0$, with $\Delta/\lambda=\infty$ if $\lambda=0$. Then, for all $N\geq1$,
\begin{multline}\label{lower:search-minimax}
\inf\,\sup_{\nu\in\mathcal E_q(\Delta,\lambda,\mathcal G)} \left[1-\PP_\nu\{\tau\leq N,\ \hat J\neq\varnothing, \ V_\tau(\hat J)\geq q\}\right] \\
\geq\max\left\{\frac14\exp\left[-\frac{\Delta^2}{q(1-q)} \min\left\{\frac NK,\frac{\Delta}{\lambda K_{\max}}\right\}\right],\ 1-2\varepsilon-\sqrt{\frac{10\varepsilon\Delta^2}{q(1-q)} \min\left\{N,\frac\Delta\lambda\right\}}\right\},
\end{multline}
where the infimum is over all procedures that are $\varepsilon$-valid on $\mathcal E_q$.
\end{proposition}

Under $0<\Delta\leq(1-q)/8$, we have $q+\Delta\geq q$ and $1-q-\Delta\geq7(1-q)/8$, so
\begin{align}\label{lower:variance-ratio}
(q+\Delta)(1-q-\Delta)\geq\frac78\,q(1-q).
\end{align}

\begin{proof}[Proof of \cref{prop:lower-risk,thm:lower}]
For $\lambda>0$, put $h=\lfloor\Delta/\lambda\rfloor$; when $\lambda=0$, interpret $h=I_h=\infty$ and omit the rate terms below.

By \cref{lem:verification-lower} (which applies for any $K$, as all strategies other than $j^\star$ have value zero) and the fact that failure within $N$ games is at least as likely as failure at a finite time, the worst-case failure probability within $N$ games is at least
\begin{align*}
\max\bigl\{1-2\varepsilon-\sqrt{8\varepsilon I_N},\ 1-2\varepsilon-\sqrt{8\varepsilon I_h}\bigr\} =1-2\varepsilon-\sqrt{8\varepsilon\min\{I_N,I_h\}}.
\end{align*}
Using \eqref{lower:variance-ratio}, $64/7<10$, and $h\leq\Delta/\lambda$,
\begin{align*}
8\varepsilon\min\{I_N,I_h\} =\frac{8\varepsilon\Delta^2\min\{N,h\}}{(q+\Delta)(1-q-\Delta)} \leq\frac{10\varepsilon\Delta^2}{q(1-q)}\min\left\{N,\frac\Delta\lambda\right\},
\end{align*}
which gives the second term in the maximum of \eqref{lower:search-minimax}. Moreover, if an $\varepsilon$-valid procedure is $(N,\varepsilon)$-successful, then $1-2\varepsilon-\sqrt{8\varepsilon I_N}\leq\varepsilon$, or equivalently
\begin{align*}
\frac{N\Delta^2}{(q+\Delta)(1-q-\Delta)}=I_N \geq\frac{(1-3\varepsilon)^2}{8\varepsilon}.
\end{align*}
Since $\varepsilon\leq1/16$ and \eqref{lower:variance-ratio} holds, this yields
\begin{align}\label{lower:certification-exact}
N\geq\frac{q(1-q)}{16\varepsilon\Delta^2},
\end{align}
and the exact same calculation with $I_h$ in place of $I_N$ gives, for $\lambda>0$,
\begin{align}\label{lower:certification-rate}
\frac\Delta\lambda\geq h\geq\frac{q(1-q)}{16\varepsilon\Delta^2}.
\end{align}

Turning to the search bounds of \cref{lem:search-lower}, observe that by concavity of $u\mapsto u(1-u)$ and the identity
\begin{align*}
(q-\Delta)(1-q+\Delta)-(q+\Delta)(1-q-\Delta)=2(2q-1)\Delta\geq0,
\end{align*}
the minimum of $u(1-u)$ over $[q-\Delta,q+\Delta]$ is attained at $q+\Delta$, where it is at least $7v_q/8$ by \eqref{lower:variance-ratio}. Integrating $\partial\kl(a,b)/\partial b=(b-a)/[b(1-b)]$ from $b=a$ to $b=a+\Delta$ with $a=q-t$ ($0\leq t\leq\Delta$) therefore gives
\begin{align*}
\kl(q-t,q+\Delta-t) =\int_0^\Delta\frac{u\,du}{(q-t+u)(1-q+t-u)} \leq\frac8{7v_q}\int_0^\Delta u\,du =\frac{4\Delta^2}{7v_q}.
\end{align*}
Because $\varepsilon\leq1/16$, we have $1/(1-\varepsilon)\leq16/15<2$ (so $2^{-1/(1-\varepsilon)}>1/4$) and $(4/7)/(1-\varepsilon)\leq64/105<1$. Applying this KL bound with $t=0$ in \eqref{lower:search-budget-exact} and with $t=\lambda i\leq\Delta$ for each $i<h$ in \eqref{lower:search-rate-exact} shows that the worst-case failure probability within $N$ games is bounded below by both
\begin{align*}
\frac14\exp\left(-\frac{N\Delta^2}{Kv_q}\right) \qquad\text{and}\qquad \frac14\exp\left(-\frac{h\Delta^2}{K_{\max}v_q}\right).
\end{align*}
Their maximum is $\frac14\exp[-\Delta^2\min\{N/K,h/K_{\max}\}/v_q]$, which is at least the first term in \eqref{lower:search-minimax} since $h\leq\Delta/\lambda$. Taking the maximum of the search and verification bounds establishes \cref{prop:lower-risk}.

Now suppose an $\varepsilon$-valid procedure is $(N,\varepsilon)$-successful, so its worst-case failure probability within $N$ games (and hence at a finite time) is at most $\varepsilon$. The two search bounds above force
\begin{align*}
\frac14\exp\left(-\frac{N\Delta^2}{Kv_q}\right)\leq\varepsilon,\qquad \frac14\exp\left(-\frac{h\Delta^2}{K_{\max}v_q}\right)\leq\varepsilon.
\end{align*}
Taking logarithms and using $\log(1/(4\varepsilon))\geq\frac12\log(1/\varepsilon)$ for $\varepsilon\leq1/16$ gives
\begin{align*}
N\geq\frac{v_q}{2\Delta^2}K\log\frac1\varepsilon,\qquad h\geq\frac{v_q}{2\Delta^2}K_{\max}\log\frac1\varepsilon.
\end{align*}
Since $\max\{x/2,y/16\}\geq\max\{x/16,y/16\}\geq(x+y)/32$ for all $x,y\geq0$, combining these inequalities with \eqref{lower:certification-exact} and \eqref{lower:certification-rate} yields
\begin{align*}
N\geq\frac{q(1-q)}{32\Delta^2}\left(K\log\frac1\varepsilon+\frac1\varepsilon\right), \qquad \frac\Delta\lambda\geq h\geq\frac{q(1-q)}{32\Delta^2} \left(K_{\max}\log\frac1\varepsilon+\frac1\varepsilon\right),
\end{align*}
which gives \eqref{lower:combined-budget} and the impossibility condition for $\Delta/\lambda$.
\end{proof}

\begin{proof}[Proof of \cref{thm:single}(b)]
For $K=1$, the observation in the two-round game of \cref{lower:route} reduces to the binary outcome, so the game is precisely the outcome-only experiment. Inequalities \eqref{lower:certification-exact} and \eqref{lower:certification-rate} from the proof of \cref{thm:lower} therefore apply to every $\varepsilon$-valid, $(N,\varepsilon)$-successful procedure, giving $N\geq B_\Delta/(16\varepsilon)$ and (when $\lambda>0$) $\Delta/\lambda\geq B_\Delta/(16\varepsilon)$.
\end{proof}

\subsection{Proof of \texorpdfstring{\cref{prop:history}}{Necessity of private randomization}}
\label{lower:history-proof}

Fix $p\in(q,1)$ and a procedure whose stopping time $\tau$ is adapted to $(\sigma(Z_1,\ldots,Z_n))_{n\geq0}$, with private seed law $\mu$. Because $\tau$ depends only on the outcomes, for each $s\geq0$ there is a set $C_s\subseteq\{0,1\}^s$ of stopping histories such that
\begin{align*}
\{\tau=s\}=\{Z_{1:s}\in C_s\},
\end{align*}
where no sequence in $C_s$ can have a strict prefix in $C_r$ ($r<s$) since the procedure stops at most once.

Define an adversarial environment that holds the win probability at $p$ until $\tau$ and drops it to zero immediately afterward: on every finite history $z_{1:n}$, put
\begin{align*}
\chi_n(z_{1:n})=\ind\{z_{1:s}\in C_s\text{ for some }s\leq n\},\qquad V_n'(z_{1:n})=p\bigl(1-\chi_n(z_{1:n})\bigr),
\end{align*}
and let $\PP'$ be the resulting law. Since $\chi_n$ is nondecreasing along every trajectory, $V_n'$ is nonincreasing, so this environment belongs to $\mathcal E_q$.

For any $s\geq0$ and $z_{1:s}\in C_s$, no proper prefix $z_{1:n}$ ($n<s$) lies in $\bigcup_{r\leq n}C_r$, so $V_n'(z_{1:n})=p$ for all $n<s$ while $V_s'(z_{1:s})=0$. Because $K=1$, the outcome probabilities are unaffected by the private seed, so for every measurable set $B$ of seeds,
\begin{align*}
\PP'(U\in B,\ Z_{1:s}=z_{1:s})=\mu(B)\prod_{n=1}^sp^{z_n}(1-p)^{1-z_n} =\PP_p(U\in B,\ Z_{1:s}=z_{1:s}).
\end{align*}
On $\{\tau=s\}$, the selection $\hat J$ is the same function of $(U,Z_{1:s})$ under both laws. Summing over $z_{1:s}\in C_s$ gives
\begin{align*}
\PP_p\{\tau=s,\ \hat J=1\}=\PP'\{\tau=s,\ \hat J=1,\ V_s'=0\}.
\end{align*}
Summing over $s\geq0$ and invoking $\varepsilon$-validity under $\PP'$ then yields
\begin{align*}
\PP_p\{\tau<\infty,\ \hat J=1\} =\PP'\{\tau<\infty,\ \hat J=1,\ V_\tau'=0\}\leq\varepsilon.
\end{align*}
In particular, for any $\Delta\in(0,1-q)$ and $\lambda\geq0$, the stationary environment $\PP_{q+\Delta}$ lies in $\mathcal E_q(\Delta,\lambda)$, where $(N,\varepsilon)$-success would require $\PP_{q+\Delta}\{\tau\leq N,\ \hat J=1\}\geq1-\varepsilon>\varepsilon$ (as $\varepsilon<1/2$), contradicting the bound above.

\section{Simulation details}\label{app:numerics}

\subsection{Detailed simulation results}\label{app:sim-confirmation}

\Cref{tab:sim-confirmation} gives the results for the three settings in \cref{fig:sim-confirmation}, including how often each verifier selected a below-target strategy or abstained.

\begin{table}[H]
\centering\small
\caption{Selection rates and mean number of games over all $20{,}000$ runs in each setting. Game counts include screening and verification.}
\label{tab:sim-confirmation}
\setlength{\tabcolsep}{4pt}
\begin{tabular}{@{}llrrrr@{}}
\toprule
Setting & Verifier & Correct (\%) & Incorrect (\%) & Abstain (\%) & Mean games \\
\midrule
200-use decline & SPAR & 94.485 & 0.000 & 5.515 & 15,041 \\
 & Private block & 29.135 & 0.070 & 70.795 & 51,692 \\
 & Likelihood ratio & 79.640 & 0.000 & 20.360 & 20,522 \\
 & Binomial & 0.220 & 94.555 & 5.225 & 95,258 \\
\addlinespace
100-use decline & SPAR & 96.715 & 0.000 & 3.285 & 13,381 \\
 & Private block & 30.140 & 0.095 & 69.765 & 50,713 \\
 & Likelihood ratio & 78.600 & 0.000 & 21.400 & 21,861 \\
 & Binomial & 0.140 & 96.740 & 3.120 & 95,557 \\
\addlinespace
Win thresholds & SPAR & 97.110 & 0.000 & 2.890 & 143,345 \\
 & Private block & 49.930 & 0.120 & 49.950 & 174,700 \\
 & Likelihood ratio & 53.285 & 0.000 & 46.715 & 168,274 \\
 & Binomial & 0.000 & 0.000 & 100.000 & 249,519 \\
\bottomrule
\end{tabular}

\end{table}

\subsection{The block verifier}\label{app:block-validity}

Condition on the screening history at commitment. This fixes the candidate, the commitment time and the remaining budget $L$. Set $B=\lceil2/\varepsilon\rceil=40$, write $m=\lfloor L/B\rfloor$ and take $k$ to be the smallest integer in $\{1,\ldots,m+1\}$ satisfying $\PP\{\operatorname{Bin}(m,q)\geq k\}\leq\varepsilon/2$. When $m=0$ the block verifier abstains at commitment. Otherwise it draws a private $R$ uniformly from $\{1,\ldots,B\}$, plays the candidate for $Rm$ more games and selects it if block $R$ contains at least $k$ wins. The final $L-Bm$ games are unused. All four methods abstain at the same time if no screen passes.

To prove that the block verifier is $\varepsilon$-valid on $\mathcal E_q$, consider the continuation that plays the candidate for all $Bm$ games without inspecting $R$. Let $p_t$ be its value $t$ games after commitment, let $W_r$ be its number of wins in block $r$ (games $(r-1)m+1$ through $rm$), and let $\theta=\inf\{t\geq 0:p_t<q\}$ (with $\inf\varnothing=\infty$). Because $R$ is uniform on $\{1,\ldots,B\}$ and independent of this continuation, the conditional probability of an incorrect selection given the screening history $\mathcal H_0$ at commitment is
\begin{align*}
\PP\{W_R\geq k,\ p_{Rm}<q\mid\mathcal H_0\} =\frac1B\sum_{r=1}^B \PP\{W_r\geq k,\ p_{Rm}<q\mid\mathcal H_0\}.
\end{align*}
For each block $r\in\{1,\ldots,B\}$, the event $\{p_{Rm}<q\}$ implies $\theta\leq rm$, which splits into two cases:
\begin{enumerate}[label=(\roman*)]
\item \emph{Block $r$ contains the first crossing,} meaning $(r-1)m<\theta\leq rm$. At most one block $r$ satisfies this condition on any trajectory, so the sum of $\PP\{(r-1)m<\theta\leq rm\mid\mathcal H_0\}$ over $r=1,\ldots,B$ is at most $1$.
\item \emph{The first crossing occurs before block $r$,} meaning $\theta\leq(r-1)m$ (which also covers a candidate already below $q$ at commitment, where $\theta=0$). On this event, \cref{ass:shape} bounds the conditional win probability by $q$ in every game of block $r$. Coupling each outcome in block $r$ to an independent Bernoulli$(q)$ draw via a common uniform variable shows that $W_r$ is stochastically dominated by $\operatorname{Bin}(m,q)$ given the history at the start of block $r$, so
\begin{align*}
\PP\{W_r\geq k,\ \theta\leq(r-1)m\mid\mathcal H_0\} \leq\PP\{\operatorname{Bin}(m,q)\geq k\} \leq\frac{\varepsilon}{2}.
\end{align*}
\end{enumerate}
Summing both contributions over $r=1,\ldots,B$ and dividing by $B=\lceil2/\varepsilon\rceil$ gives
\begin{align*}
\PP\{W_R\geq k,\ p_{Rm}<q\mid\mathcal H_0\} \leq\frac1B+\frac\varepsilon2 \leq\varepsilon.
\end{align*}
Averaging over $\mathcal H_0$ yields the unconditional $\varepsilon$-validity bound.

\subsection{Likelihood-ratio and binomial comparators}\label{app:sim-test-comparators}

Both tests start at commitment and use only fresh outcomes from the candidate that passed. Write $X_t$ for its number of wins in the first $t$ verification games. The likelihood-ratio test selects at the first $t\leq L$ at which
\begin{align*}
\left(\frac{q+0.05}{q}\right)^{X_t} \left(\frac{1-q-0.05}{1-q}\right)^{t-X_t}\geq 1/\varepsilon=20 .
\end{align*}
It abstains at $L$ if the threshold is never reached.

The binomial test plays all $L$ verification games. Let $X\sim\operatorname{Bin}(L,q)$ and choose an integer $k$ with $\PP(X>k)\leq\varepsilon\leq\PP(X\geq k)$. The test selects when $X_L>k$, abstains when $X_L<k$, and at $X_L=k$ selects with probability $(\varepsilon-\PP(X>k))/\PP(X=k)$ using an independent private draw.

For a stationary candidate with value at most $q$, the likelihood ratio is a nonnegative supermartingale starting at one, so the probability of reaching its threshold is at most $\varepsilon$. The binomial test has selection probability exactly $\varepsilon$ at $q$ and at most $\varepsilon$ below $q$. These calculations allow for an arbitrary preceding screening history. In a deteriorating environment the earlier verification games can have values above $q$ even when the value at selection is below $q$, so the stationary calibrations do not bound the probability of an incorrect selection in that setting.

\subsection{Finite-horizon calibration and numerical checks}\label{app:cap-computation}

For each combination of $K$ and $N$, the sequence of possible screens is deterministic along the path on which every screen fails. We precompute the tilt, remaining horizon and corresponding cap for each screen in this sequence. A skipped screen consumes zero games. Once a screen passes, we use a fresh private threshold for SPAR, start with capital $\kappa_L$, and update the capped log-likelihood ratio one game at a time until selection or the end of the remaining horizon.

We compute the null expectation in \eqref{eq:kappa-L} without Monte Carlo. Fix a screen's tilt $\delta$, let $M_s=\prod_{i=1}^s g(Y_i)$ be the uncapped likelihood ratio after $s$ independent Bernoulli$(q)$ games, and write $c_s=\EE_q\mathcal K_s^{(1)}$ for the expected unit-cap capital. By the martingale identity in Step~3 of \cref{lem:finite-cap}, we have $\EE_q S_h^{(1)}=1-c_h$ at $h=L-1$.

To evaluate $c_h$ for long horizons, set $a_s=\EE_q\min\{1,M_s\}$, which is the sum over all win counts $k\in\{0,\ldots,s\}$ of the smaller of the two binomial probabilities $\PP_q(X=k)$ and $\PP_{q+\delta}(X=k)$. Because the increments $\log g(Y_1),\ldots,\log g(Y_s)$ are independent and identically distributed, reversing their order shows that
\begin{align*}
\mathcal K_s^{(1)} =\frac{M_s}{\max_{0\leq u\leq s}M_u} =\min_{0\leq u\leq s}\frac{M_s}{M_u} \stackrel{d}{=} \min_{0\leq u\leq s}M_u.
\end{align*}
Spitzer's identity \citep{spitzer1956} then gives the generating function
\begin{align*}
C(z):=\sum_{s=0}^\infty c_s z^s=\exp\left(\sum_{k=1}^\infty\frac{a_k}{k}z^k\right),\qquad |z|<1.
\end{align*}
Differentiating both sides gives $C'(z)=C(z)\sum_{k=1}^\infty a_k z^{k-1}$. Matching the coefficient of $z^{s-1}$ on both sides yields the finite recursion
\begin{align*}
c_0=1,\qquad c_s=\frac{1}{s}\sum_{k=1}^s a_k c_{s-k}.
\end{align*}
We evaluate these convolutions in blocks using fast Fourier transforms. Finally, because $\mathcal K_h^{(1)}\leq \sqrt{M_h}$ (as $\min\{1,M_h\}\leq\sqrt{M_h}$ and $\mathcal K_h^{(1)}\leq\min\{1,M_h\}$), we have
\begin{align*}
c_h\leq\EE_q\sqrt{M_h}=\bigl(\sqrt{q(q+\delta)}+\sqrt{(1-q)(1-q-\delta)}\bigr)^h.
\end{align*}
Whenever this upper bound is below $2^{-55}$, the value $1-c_h$ rounds to $1$ in double precision.

Direct propagation of the capped process over $300$ games reproduces the expected unit-cap excess to within $10^{-13}$, and an independent implementation reproduces every cap used in the simulations and in the replay to within $10^{-14}$.

\section{MonitoringBench replay and implementation}\label{app:realdata}

\subsection{Data and defense state}

\Cref{fig:rd-accuracy} uses the first full-trajectory monitor scores for the $811$ records in \cref{tab:rd-source-banks} (\href{https://huggingface.co/datasets/neur26anonsub/ctrldataset2026/tree/6c72cf19883376b921a94a0dd17dc8ab6b25e52f}{dataset}). A score above the benchmark cutoff gives a flag. The cutoffs are $2$ for GPT-4.1 and $8.97$ for GPT-5.

For a given source, let $\mathcal R$ be its bank of records, let $\mathcal R_1,\mathcal R_2\subseteq\mathcal R$ be the records that pass GPT-4.1 and GPT-5, respectively, and let $u(i)$ be the underlying task of record $i$. Let $d_{t,u}$ indicate whether the second gate has been enabled for task $u$ after interaction $t$. The value is
\begin{align}\label{rd:value}
p_t=\frac1{|\mathcal R|}\sum_{i\in\mathcal R_1}\bigl(1-d_{t,u(i)}\ind\{i\notin\mathcal R_2\}\bigr).
\end{align}
We score interaction $t+1$ using the gates in place at $t$, then increment the task's flag count if the drawn record is outside $\mathcal R_2$. On the second flag we set $d_{t+1,u(i)}=1$. This update applies during both screening and verification, so a recommendation after this interaction is evaluated at $p_{t+1}$.

Let $I_s$ be the record drawn at interaction $s$ and $\mathcal G_t=\sigma(I_1,\ldots,I_t)$ the full replay history. Since $I_{t+1}$ is an independent uniform draw, \eqref{rd:value} gives $\EE[Z_{t+1}\mid\mathcal G_t]=p_t$.

\subsection{Screening and SPAR calibration}

At each of the eleven targets the feasible screen lengths are $2,4,8,16,32$, leaving verification horizons $L=78,74,66,50,18$, respectively. We compute the null expectation in \eqref{eq:kappa-L} recursively for each combination. Even with all wins, the selection probability is at most $0.531$, so every feasible screen uses the mean criterion $\bar Z\geq q+\delta_l$ in \cref{alg:search}.

The commitment time $\tau_0$ is a stopping time for $(\mathcal G_t)$ because screening uses only the pass indicators. Given $\mathcal G_{\tau_0}$ at commitment, $\delta$ and $L$ are fixed, and the continued replay is independent of the fresh private threshold. Since $p_t$ is nonincreasing and $\EE[Z_{t+1}\mid\mathcal G_t]=p_t$, \cref{lem:finite-cap} gives
\begin{align*}
\PP\{\hat J\neq\varnothing,\ p_\tau<q\mid\mathcal G_{\tau_0}\}\leq\varepsilon .
\end{align*}
Averaging gives the same bound conditional on the observed pass history at commitment, and unconditionally since runs without a commitment end in abstention.

\subsection{Comparison procedures and numerical evaluation}\label{app:rd-procedures}

The likelihood-ratio and binomial tests follow \cref{app:sim-test-comparators} with $L=N=80$ and win counts starting at the first interaction. The block verifier of \cref{app:block-validity} has $40$ blocks of two interactions each. Under independent Bernoulli$(q)$ outcomes, even two wins have probability $q^2>0.025$ at every target, so the rejection region is empty. The expected stopping time is $2(40+1)/2=41$ interactions.

We average over the private randomization of each method on each sampled path. For \cref{fig:rd-accuracy}, we divide the estimated probability of a correct recommendation by that of any recommendation. Its pointwise $95\%$ Monte Carlo intervals reflect sampling variation across paths from the same bank.

\end{document}